\documentclass[preprint]{elsarticle}
\usepackage{CJKutf8} 
\usepackage{times}  
\usepackage{helvet} 
\usepackage{courier}  
\usepackage{graphicx} 
\usepackage{caption} 

\usepackage{subfigure}
\usepackage{hyperref}
\usepackage[utf8]{inputenc}
\usepackage[english]{babel}

\newtheorem{theorem}{Theorem}

\usepackage{epsfig}
\usepackage{epstopdf}
\usepackage{amssymb}
\usepackage{amsmath}
\usepackage{multirow}
\usepackage{makecell}
\usepackage{tikz}
\usepackage{algorithm}
\usepackage{booktabs}
\usepackage{algorithmic}
\usepackage{array}
\usepackage{longtable}
\usepackage{stfloats}
\usepackage{supertabular}
\usepackage{lineno} 

\usepackage{color}

\newcommand{\nvars}{\mathsf{V}}

\newcommand{\ubounds}{\mathsf{U}}
\newcommand{\lbounds}{\mathsf{L}}
\newcommand{\props}{\mathcal{P}}
\newcommand{\intervals}{\mathcal{I}}

\newcommand{\mxplanner}{\textsf{mx}}
\newcommand{\metricff}{\textsf{MFF}}
\newcommand{\ScottyActivity}{\textsf{SA}}
\newcommand{\popcorn}{\textsf{POP}}

\newcommand{\Revised}[1]{\color{black} #1 \color{black}}
\newcommand{\RevisedByHankz}[1]{\color{black} #1 \color{black}}

\newcommand{\MajorRevision}[1]{\color{black} #1 \color{black}}

\def\ours{\tt mxPlanner}
\newcommand{\ignore}[1]{{}}
\newcommand{\Actions}{\mathcal{A}}

\newcommand{\maximum}{\mathsf{max}}
\newcommand{\ReLu}{\mathrm{ReLU}}

\newcommand{\plan}{\sigma}

\newcommand{\bound}{\delta}

\begin{document}

\title{\MajorRevision{ Gradient-Based Mixed Planning with Symbolic and Numeric Action Parameters
}}
\author[mymainaddress]{Kebing Jin}
\ead{jinkb@mail2.sysu.edu.cn}

\author[mymainaddress]{Hankz Hankui Zhuo\corref{mycorrespondingauthor}}
\cortext[mycorrespondingauthor]{Corresponding author}
\ead{zhuohank@mail.sysu.edu.cn}

\author[mymainaddress]{Zhanhao Xiao}
\ead{xiaozhh9@mail.sysu.edu.cn}

\author[mymainaddress]{Hai Wan}
\ead{wanhai@mail.sysu.edu.cn}

\author[mysecondaryaddress]{Subbarao Kambhampati}
\ead{rao@asu.edu}

\address[mymainaddress]{School of Computer Science and Engineering, Sun Yat-sen University, Guangzhou, China}
\address[mysecondaryaddress]{Department of Computer Science and Engineering, Arizona State University, US}

\begin{frontmatter}
\begin{abstract}
Dealing with planning problems with \MajorRevision{both logical relations and numeric changes} in real-world dynamic environments is challenging.  \Revised{Existing numeric planning systems for the problem} often discretize numeric variables or impose convex constraints on numeric variables, which harms the performance when solving problems. In this paper, we propose a novel algorithm framework to solve numeric planning problems mixed with \MajorRevision{logical relations and numeric changes}
based on \Revised{gradient descent.} We cast the numeric planning with
\MajorRevision{logical relations and numeric changes} as an optimization problem. Specifically, \MajorRevision{we extend syntax to allow parameters of action models to be either objects or real-valued numbers, which enhances the ability to model real-world numeric effects. Based on the extended modelling language,}\Revised{we propose a gradient-based framework to simultaneously optimize numeric parameters and compute appropriate actions to form candidate plans. \MajorRevision{The gradient-based framework is composed of an algorithmic heuristic module based on propositional operations to select actions and generate constraints for gradient descent, an algorithmic transition module to update states to next ones, and a loss module to compute loss. We repeatedly minimize loss by updating numeric parameters and compute candidate plans until it converges into a valid plan for the planning problem.
}
In the empirical study, \MajorRevision{we exhibit that our algorithm framework is both effective and efficient in solving planning problems mixed with logical relations and numeric changes, especially when the problems contain obstacles and non-linear numeric effects.}
}
\end{abstract}
\end{frontmatter}

\section{Introduction}
Autonomous robots have become commonplace in commercial and industrial settings. For example, hospitals use autonomous mobile robots to move materials. Warehouses exploit mobile robotic systems to efficiently move materials from stocking shelves to order fulfillment zones.
\Revised{In scientific missions, Woods Hole Oceanographic Institution (WHOI) uses autonomous underwater vehicles (AUVs) to collect data of scientific interest.}\ignore{In these real-world applications, it is desirable to have autonomous robots be capable of autonomously planning with optimization of numeric objectives related to metrics such as resources, time, and navigation distance, besides achieving desirable goals (e.g., in the form of propositions).}\MajorRevision{In those real-world applications, it is desirable to have autonomous robots be capable of autonomously planning with optimization of numeric objectives, such as minimizing resources and navigation distance, during planning towards desirable goals (e.g., in the form of propositions).
}


\Revised{

To handle numeric planning problems, there have been approaches proposed to discretize numeric space and then use heuristic searching to approximate the optimization result, such as Metric-FF \cite{DBLP:journals/jair/Hoffmann03} and LPRPG \cite{DBLP:conf/aips/ColesFLS08a}.
Those numeric planners, however, do not address planning missions over long-term reasoning for autonomous robots since the size of discretization needs to be fixed in advance manually.
It is hard to determine a proper size of discretization beforehand for various planning problems with respect to different environments.}\emph{For example, in the ocean mission scenario as shown in Figure \ref{figure:motivating_example}, a ship is equipped with an AUV (Autonomous Underwater Vehicle), i.e., the submarine in the figure, and an ROV (Remotely Operated Vehicle), i.e., the robot in the figure. The AUV aims to take images in region A,
and the ROV aims to take samples in regions B and C.
The planning mission is to make the three vehicles, i.e., the ship, AUV and ROV, reach the destination region (the blue area denoted by ``destination region''), with avoiding obstacles (the black areas).
\MajorRevision{Each action has multiple parameters, where a movement is determined by three numeric ones, i.e., x-velocity $v_x$, y-velocity $v_y$ and duration $d$.}
In particular, if the ship deploys the ROV,
\begin{figure}[!ht]
\setlength{\abovedisplayskip}{1pt}
\setlength{\belowdisplayskip}{1pt}
\centering
\subfigure[Metric-FF]{
    \begin{minipage}[b]{1\textwidth}
    \includegraphics[width=1.1\textwidth]{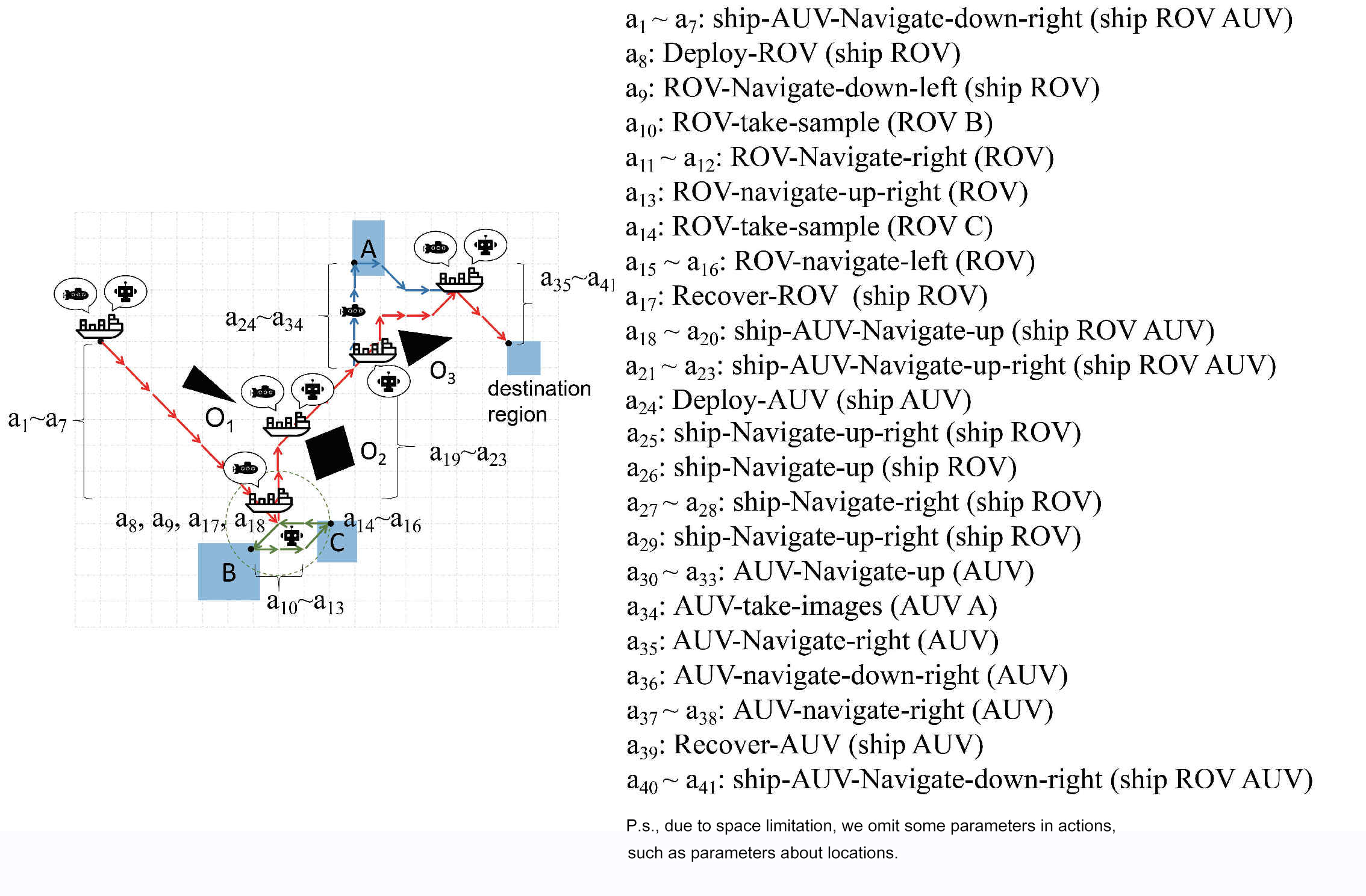}
    \end{minipage}
}
\subfigure[{\ours}]{
    \begin{minipage}[b]{1\textwidth}
    \includegraphics[width=1\textwidth]{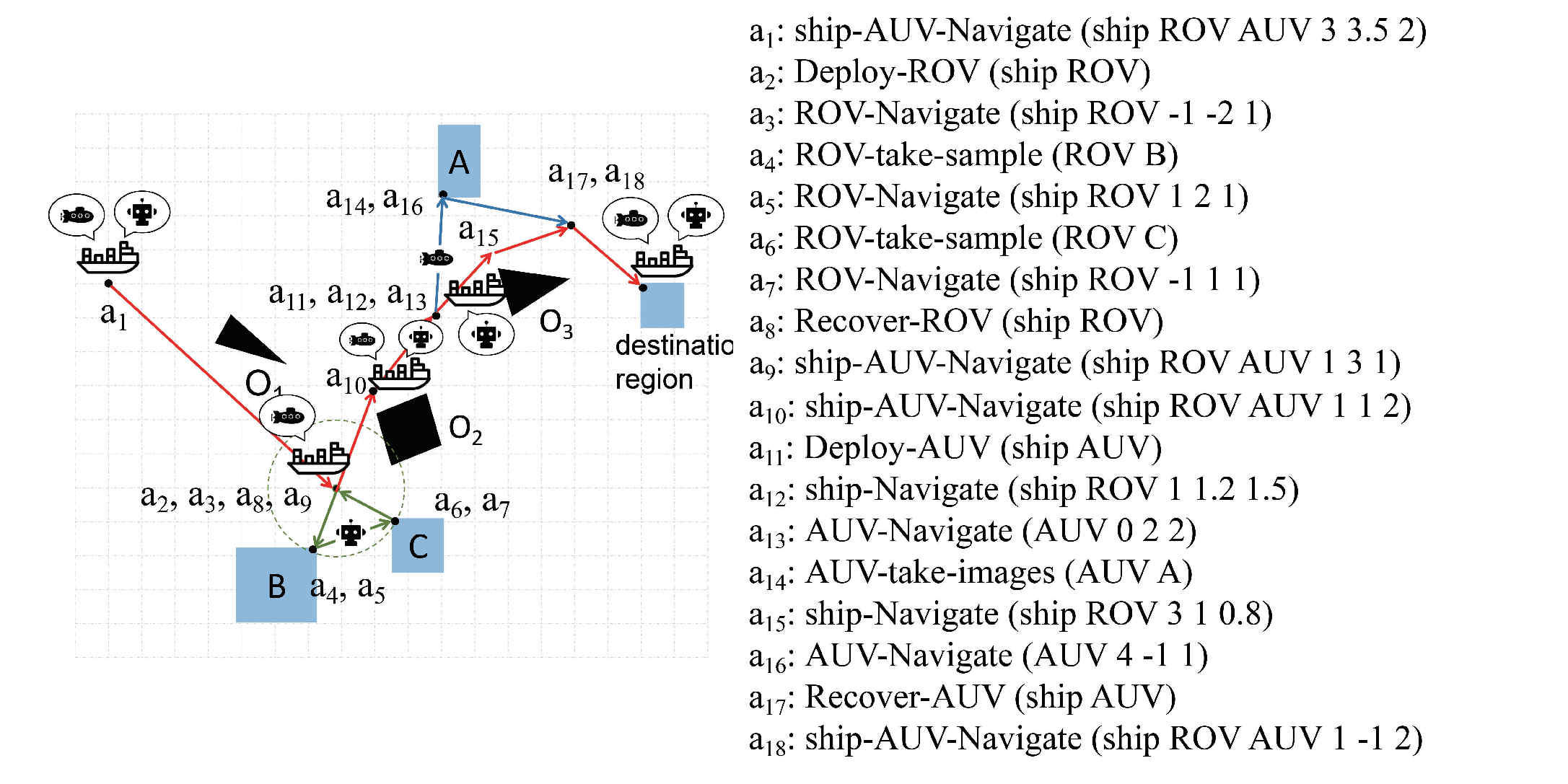}
    \end{minipage}
}
\caption{\MajorRevision{Valid plans based on Metric-FF and {\ours} in ocean mission scenario.}}
\label{figure:motivating_example}
\end{figure}
which moves within a circle centered over the ship with radius $R$, the ship needs to stay at the deployment location until the ROV comes back again. We use red arrows to indicate the trajectory of the ship, green arrows to indicate the trajectory of the ROV, and blue arrows to indicate the trajectory of the AUV. An example plan generated by Metric-FF can be found from the red curve shown in Figure \ref{figure:motivating_example}(a) (the corresponding action sequence is shown on the right).
\MajorRevision{Note that all the movements of Metric-FF are fixed, owing to the requirements of fixed numeric effects of Metric-FF. Action models, such as ``ROV-Navigate-up-right'', can be found from Figure \ref{problem_definition:comparison_with_PDDL}(d). The transition of locations are done by dividing space beforehand and using propositions to indicate them.}}As shown in Figure \ref{figure:motivating_example}(a), each movement \MajorRevision{computed by} Metric-FF is fixed, which indicates they cannot make up a flexible plan. The desired solution is to dynamically adapt step lengths according to the positions and shapes of obstacles when generating the plan, as the one
shown in Figure  \ref{figure:motivating_example}(b).
There are also approaches that introduce control parameters into action models to handle numeric planning problems. For example, Kongming \cite{DBLP:conf/aips/LiW08} was proposed to merge activity planning and trajectory optimization by introducing control parameters into effects of actions. It, however, requires fixed time discretization, which restricts its ability to scale to more complicated planning missions where short and long lived activities coexist. \Revised{POPCORN \cite{DBLP:conf/ecai/SavasFLM16} was built, using real number control parameters to allow action models with infinite domain parameters. It can, however, only be used in discrete numeric effects with constant rates of change, and only supports linear constraints.} To relax the limitations of linear constraints, ScottyActivity \cite{DBLP:journals/jair/Fernandez-Gonzalez18}, a hybrid activity and trajectory planner, was developed to effectively generate plans for autonomous robots over long-term reasoning. ScottyActivity supports convex state constraints and control parameters that are often used to model controllable change rates and robot velocities. With a continuous time formulation, ScottyActivity scales effectively to planning missions with long durations without discretizing control parameters, even in the presence of both short and long span activities. ScottyActivity utilizes convex optimization to choose continuous states, control parameters and times. \Revised{Despite the success of ScottyActivity, it still requires the continuous search space to be convex, which in consequence does not allow ``obstacles'' in a navigation domain, which makes the search space be non-convex.} In many real-world applications, however, the continuous search space is often non-convex. \Revised{The problem will become more complicated when the non-convex space involves different variable types.} It is undoubtedly challenging to solve planning missions that require both discrete action planning and non-convex continuous state searching.

\Revised{
In this paper, we extend the scope of the previous numeric planning problem by simultaneously allowing three properties: non-convex continuous numeric space, propositional and numeric action preconditions and action effects, and more complicated action effects.
We propose a novel approach, {\ours}, which stands for gradient-based} \underline{\textbf{m}}i\underline{\textbf{x}}ed \underline{\textbf{Planner}},\ignore{with discrete and continuous actions} to deal with \MajorRevision{problems with logical relations and numeric changes in non-convex numeric space, determined by multiple symbolic and numeric action parameters.} Specifically, \MajorRevision{we build a gradient-based framework, which is composed of an algorithmic heuristic module, an algorithmic transition module and a loss function. The heuristic module estimates appropriate actions to generate a plan $\plan$ to transit initial state $s_0$ to the goal by relaxation. The transition module updates states according to plan $\sigma$.}After that, we explore \MajorRevision{gradient descent to optimize numeric parameters to minimize accumulated loss computed by loss functions.}
We repeat \ignore{these two phases}\MajorRevision{the above-mentioned procedures of computing plans and optimizing numeric parameters} until our approach converges to valid plans of our planning problems. Note that we do not only update the continuous numeric parameters of actions, but also modify the actions in $\sigma$ simultaneously.
Our {\ours} is capable of handling flexible numeric changes determined by multiple numeric parameters of actions, instead of fixing numeric effects as done by Metric-FF.
\Revised{\emph{For example, in the AUV domain shown in Figure \ref{figure:motivating_example}, another plan calculated by our {\ours} can be found \MajorRevision{from Figure \ref{figure:motivating_example}(b). The navigation distance computed by {\ours} is much shorter than the one calculated by Metric-FF, since {\ours} is more flexible than Metric-FF with fixed step length. }
}}


\Revised{
We summarize the contributions of the paper as follows:
\begin{itemize}
    \item We extend previous numeric planning problems to simultaneously allow continuous numeric space to be non-convex, preconditions and effects of actions to be both propositional and numerical, and numerical effects to be non-linear.
    \item To handle \emph{mixed} planning problems, we propose a novel approach, {\ours}, which borrows the framework of recurrent neural networks with integration of algorithmic heuristic searching in the framework. \ignore{The combination keeps {\ours} from designating parameter bounds.}
    \item We empirically show that our {\ours} outperforms existing planners in solving \emph{mixed} planning problems with non-convex numeric space \MajorRevision{(i.e., including obstacles)} and complex preconditions and effects (involving propositions and numeric expressions determined by multiple action parameters). Besides, the experimental results show {\ours} is also competitive when handling \emph{mixed} planning problems with convex continuous numeric space.
\end{itemize}

}

In the remainder of the paper, we first introduce related works and a formal definition of our \Revised{\emph{mixed} planning} problem. After that, we present our approach in detail and evaluate our approach by comparing with previous approaches to exhibit its superiority. Finally, we conclude the paper with future work.

\section{Related Work}
\MajorRevision{Our work is related to the following four aspects: (1) numeric planning, (2) hybrid planning, (3) path planning, and (4) gradient-based planning. We will describe those aspects in detail below.}
\subsection{Numeric Planning}
Automated planning aims to find a sequence of actions to complete a given task. There have been considerable advancements in the development to solve state-dependent goals planning tasks. A significant part of this progress comes from successful studies on classical planning composed of discrete actions, \Revised{which are mostly based on heuristic forward search \cite{DBLP:conf/aips/EyerichMR09,DBLP:conf/aips/GereviniS02,DBLP:journals/jair/Helmert06,DBLP:journals/aim/Hoffmann01}. }

However, real-world planning problems are often along with numeric variables, such as time, fuel, money and materials. Researchers extended the classical formulation to support numeric variables and metric minimizing, in order to solve real-world missions.
For example, Metric-FF \cite{DBLP:journals/jair/Hoffmann03} was built based on discretization and ignoring all effects that decrease the value of the affected variables. LPRPG \cite{DBLP:conf/aips/ColesFLS08a} uses linear programming to calculate interval bounds from the constraints in the action preconditions and effects, as an adjunct to a relaxed planning graph. \citet{DBLP:conf/aips/IvankovicHTSN14} introduced an algorithm to find an optimal plan under the classical, discrete action models with systems of constraints. \citet{DBLP:journals/jair/KeyderHH14} proposed an approach to improve delete relaxations to compute upper and lower bounds by \Revised{finding} an informative set of conjunctions. Besides, in order to handle objective optimization, linear programming is an important technique, which can minimize or maximize a linear function when subjected to various constraints. For example, \citet{do2003improving} proposed a model using MILP to deal with optimization under an objective function with a variety of temporal flexibility criteria, such as makespan. ILP-PLAN \cite{ILP-PLAN} represents planning problems with resources or complex objective functions as Integer Linear Programming models. BBOP-LP \cite{DBLP:conf/aips/BentonBK07} uses linear programming to encode a relaxed version of the partial satisfaction planning problem to obtain search heuristics.

\MajorRevision{On the other hand, in order to model complex numeric changes in realistic problems,}\Revised{researchers introduced control parameters. For example, \citet{DBLP:conf/mates/PantkeEH14} proposed a PDDL-based multi-agent planning system reasoning about the control parameters in the production control domain applications. Kongming \cite{DBLP:conf/aips/LiW08} was proposed to capture the control parameters with hybrid flow graphs which are capable of representing continuous trajectories in discrete planning frameworks. However, Kongming is limited by its fixed time discretization. POPCORN \cite{DBLP:conf/ecai/SavasFLM16} introduces continuous control parameters into the domain-independent planning applications to allow infinite continuous search space in actions. However, POPCORN can only be used in discrete numeric effects.
}

In general, those numeric planning approaches are limited by discretization \Revised{or} linear optimization methods, lacking the ability to handle complex planning problems, which contain actions whose numeric effects are allowed to be linear and non-linear in convex or non-convex search spaces.

\subsection{Hybrid Planning}
\MajorRevision{
Hybrid planning problems involve discrete and continuous actions, where discrete actions include propositional operations and discrete numeric changes, and continuous actions have continuous effects continuing to be applied over time according to temporal equations. Continuous effects can be linear or non-linear.}\Revised{Discretization is one of the viable approaches to solve hybrid planning problems \cite{DBLP:journals/apin/PennaMM12,DBLP:conf/aaai/PiotrowskiFLMM16,DBLP:conf/ecai/ScalaHTR16}.
\MajorRevision{
However, it is hard for discretization-based planners to scale to larger search space where short and long lived activities coexist in, especially when the problems are not time-discretized but with effects relying on different durations.}Indeed, there are planners that do not rely on discretization. For example, COLIN \cite{DBLP:journals/jair/ColesCFL12} presents a heuristic forward planner with delete relaxations approach to handle temporal planning with continuous time-dependent effects. It does not rely on discretization but linear programs. However, it only supports continuous time-dependent effects with ``constant'' rates of change and it is not capable of  dealing with non-convex problems. POPF \cite{DBLP:conf/aips/ColesCFL10} and OPTIC \cite{DBLP:conf/aips/BentonCC12} convert all numeric constraints to linear programming constraints, then they use forward searches combined with Mixed Integer Programming (MIP) solvers to ensure the temporal and numeric constraints of the problems are met. Although these successful approaches were proposed, they only can solve problems \ignore{whose continuous actions with constant effects}\MajorRevision{with linear continuous effects} and they can't handle real-world planning missions with non-linear numeric changes.
\MajorRevision{ScottyActivity \cite{DBLP:journals/jair/Fernandez-Gonzalez18} was proposed to handle a mix of discrete and continuous actions by making use of convex optimization, as a continuation of COLIN and OPTIC. However, ScottyActivity cannot deal with non-convex problems, e.g., \textbf{with obstacles in planning problems}, as done by our {\ours} approach. \ignore{ScottyPath \cite{ScottyPath} was proposed to handle domains with obstacles, \MajorRevision{it is based on the the opinion of ScottyActivity about using convex optimization to search a shortest path. It first divides a non-convex continuous numeric space into several convex safe regions, then constructs a graph and uses A* algorithm to search a best sequence of safe regions. Finally, it makes use of convex optimization to optimize for a shortest path. ScottyPath, however, cannot handle logical relations but focus on generating paths for a set of ordered or unordered goal region.}}
\ignore{Although safe regions generation can consider obstacles, a shorter plan may exist avoiding all obstacles and going through areas not covered by the safe regions.}}\citet{DBLP:conf/nips/WuSS17} used a framework of RNNs and efficiency of Tensorflow to handle hybrid planning problems with non-linear numeric effects. However, their method cannot handle propositional effects and preconditions. In other words, it cannot be used to solve planning problems mixed with discrete logical relations and continuous numeric changes and with specific propositional goals.}SCIPPlan \cite{DBLP:conf/cpaior/SayS19} plans in metric discrete time hybrid factored planning domains with instantaneous continuous actions, with the purpose of handling discrete and continuous numeric effects and collision avoidance. To handle non-linear continuous effects, SMTPlan+ \cite{DBLP:conf/aips/CashmoreFLM16} makes use of polynomial process models to deal with non-linear polynomial changes. \Revised{And OPTIC++ \cite{DBLP:conf/aips/DenenbergC19} reasons with non-linear domains by generating piecewise linear upper and lower bound approximations for non-linear functions. However, the continuous changes of variables are limited to be monotonic and OPTIC++ also cannot support non-convex domains. }

In general, some of these state-of-the-art approaches make use of discretization, which limits their applications in real-world. Some of them cannot handle non-linear effects in non-convex problems due to their linear programming or convex optimization. {\ours} avoids these problems by gradient descent.

\subsection{\MajorRevision{Path} Planning}
\MajorRevision{Path planning aims at finding paths for agents, such that agents can move along paths from initial states to goal states without colliding with obstacles and other agents \cite{alvarez2004evolutionary,fu2011phase,guo2009path,li2016hybrid,liu2008path,nikolos2003evolutionary,zhang2014automatic,zheng2005evolutionary}. Path planning problems can be regarded as optimization problems with multi goals, which can be solved by graph-based search, mathematical optimization algorithms, artificial potential fields, or sample-based exploring methods. For example, \citet{Trajectory:minimum-time} proposed a minimum time approach, an integrated approach treating obstacle avoidance and envelop protection as inequality constraints in state space formulations, to generate non-linear obstacle-free trajectories.}\citet{Chen2011_trajectory} introduced a global optimization approach to handle kinematic and dynamic constraints based on hybrid genetic algorithms. \MajorRevision{They used particle swarm optimization (PSO) algorithm to estimate initial trajectories and optimized them by local conjugate gradient methods.}\citet{Trajectory:tree} proposed a tree-based algorithm to compute feasible minimum energy trajectories from start positions to distant goals by precomputing sets of branches from the space of allowable inputs. \citet{GRACIA20121} developed a supervisory loop to fulfill workspace constraints caused by robot mechanical limits, collision avoidance, and industrial security in robotic systems with geometric invariance and sliding mode related concepts.
\MajorRevision{ScottyPath \cite{ScottyPath} is based on the the main idea of ScottyActivity \cite{DBLP:journals/jair/Fernandez-Gonzalez18} about using convex optimization to search shortest paths, computing obstacle-free paths in non-convex continuous spaces. It first divides non-convex continuous numeric spaces into several convex safe regions, then constructs graphs and uses A$^*$ algorithm to search best sequences of safe regions. Finally, it makes use of convex optimization to optimize for shortest paths. Those path planners, however, ignore logical relations but focus on generating paths for sets of ordered or unordered goal regions.}

\subsection{\MajorRevision{Gradient-based Planning}}
\MajorRevision{Recently, machine learning approaches have attracted lots of attention, where gradient descent is a preferred way to optimize machine learning algorithms to find local and global minimums. Gradient descent is an optimization algorithm for finding local minimums of differentiable functions, which can be used in neural planners to handle continuous state-action domains. For example, \citet{nguyen1990neural,fairbank2012value,DBLP:conf/icml/SilverLHDWR14,DBLP:conf/nips/HeessWSLET15} and \citet{bueno2019deep} made use of policy gradient algorithms to maximize the expectation of cumulative rewards for discrete-time Markov Decision Processes (MDPs) with continuous states and actions, according to the gradients of this expectation with respect to the policy parameters. \citet{bajpai2018transfer} and \citet{garg2019size} used deep reactive policies and transferred experience from policies of other problems of the same domain, aiming at learning transformations from the state and action spaces to latent state and action spaces. \citet{DBLP:conf/ijcai/CuiK16} used symbolic simulations capturing explicit approximations of the action-value Q function and performed a gradient-based search to compute actions based on states. DreamerV2 \cite{hafner2020mastering} learns policy based on image sequences from Atari games, to predict the outcomes of potential actions to enable planning.

On the other hand, applications based on machine learning in other fields also made progress by combining planning and optimizing via gradient descent, such as in text and image related problems.} For example, in order to generate coherent texts, \citet{DBLP:conf/aaai/YaoPWK0Y19} built a plan-and-write framework to generate stories according to explicit storylines composed of keywords, which are planned out based on given titles. \citet{DBLP:journals/corr/abs-1809-10736} introduced a deep reinforcement learning approach to controllable story plot generation from a predefined start state and a predefined goal state. Gradient-based planning methods were also used to learn models from raw observations to capture the transitions between images. For example, \citet{DBLP:journals/jair/KonidarisKL18} automatically abstracted state representations from raw observations, relied on a prespecified set of skills, and expressed them in the form of PDDL \cite{PDDL}. Causal InfoGAN \cite{NIPS2018_8090} uses Gumbel-Softmax to backprop through transitions of discrete binary states, and leverages the structure of the binary states for planning.

\section{Problem Definition}
In this paper we aim to solve \Revised{sequential} planning problems with both \Revised{discrete logical relations and continuous numeric changes.}Specifically, we aim to consider the planning problem that has the following features:
\begin{itemize}
    \item The \emph{continuous numeric space} (which is composed of a set of numeric variables in each state) is allowed to be non-convex; note that the objective based on the non-convex space is also non-convex.
    \item The effects of actions are allowed to be discrete \Revised{(propositional operations),} continuous \Revised{(numeric changes)} or both.
    \item \MajorRevision{The numeric effects are allowed to be non-linear, linear, or both.}
    \item The numeric variables are allowed in preconditions of actions.
\end{itemize}
We call the above-mentioned planning problem as a \emph{mixed} planning problem.
Formally, we represent our mixed planning problem as a tuple \Revised{$\mathcal{M} = \langle \mathcal{S}, s_0, g, \Actions, \mathcal{B}\rangle$,} where $\mathcal{S}$ is a set of states, each of which is composed of a set of propositions and assignments of \ignore{continuous} numeric variables in the prefix form (e.g., ``(= $v$ 1)'' indicates variable $v$ is assigned to be 1). We define a \emph{continuous numeric space} by the set of numeric variables. Specifically, we use a vector $\nvars$ to denote all of the numeric variables and use $\nvars_i=\langle v_i^1, v_i^2, \ldots, v_i^K\rangle$ to denote their values in state $s_i$. \MajorRevision{Similarly, propositions in state $s_i$ are defined by $\props_i = \langle p_i^1, p_i^2, \ldots p_i^M\rangle$ where $p_i^j$ is 1 or -1, indicating whether a proposition is
included or not.
In other words, a state $s_i$ includes two parts, i.e., $s_i = \langle \props_i, \nvars_i\rangle$.
}$s_0\in\mathcal{S}$ is an initial state, and $g$ is a goal which is composed of a set of propositions.

$\Actions$ is a set of action models,
each of which is composed of a tuple $\langle a, \mathit{pre}(a),\mathit{eff}(a)\rangle$, where $a$ is an action name with zero or more parameters. An action is a grounding of an action model, i.e., every parameter in the action model is an object or a real number. Specifically, we use a vector $\Theta=\langle \Theta_0, \Theta_1, \ldots, \Theta_{N-1}\rangle$ to denote all numeric parameters occurring in $\Actions$ of $N$ steps, where $N$ is the maximal length of the potential solution plan $\plan$. $\Theta_i=\langle \theta_i^1, \theta_i^2, \ldots, \theta_i^T\rangle$ is all of the numeric parameters of actions in $\Actions$ in step $i$. $T$ is the number of different numeric parameters of all actions in $\Actions$. Note that the parameters of actions are different from the set of variables in states. For example, \MajorRevision{``ROV-Navigate (?v1 ?v2 - vehicle $?v_x$ $?v_y$ $?d$ - $\mathbb{R}$)'' indicates a movement of a vehicle ``?v2'', and each movement is determined by three numeric parameters. }
$\mathit{pre(a)}$ is a set of preconditions, each of which is either a proposition or a numeric constraint. For example, ``($\geq$ $?x$ $?l$)" is a numeric constraint indicating variable $?x$ should not be less than $?l$.
Each numeric precondition can also be considered as a real-number interval.
The precondition set can be divided into two sets: propositional preconditions, denoted by $\mathit{pre}^{p}(a)$, and numeric preconditions $\mathit{pre}^{n}(a)$. $\mathit{eff}(a)$ is a set of effects, each of which is either a literal (i.e., a propositional or its negation), or a numeric updating expression. For example, ``(increase location-x ($*$ $?v_x$ $?d$)'' means to increase the value of ``location-x'' by $?v_x \times ?d$. We call the sets of positive literals and negative literals in the effect set $\mathit{eff}(a)$ as positive effects and negative effects, denoted by $\mathit{eff}^+(a)$ and $\mathit{eff}^-(a)$, respectively. Also, we call the set of numeric updating expressions as numeric effects, denoted by $\mathit{eff}^{n}(a)$.
An action model or action is called \MajorRevision{\emph{numeric}} if it has numeric updating effects; otherwise it is called \MajorRevision{\emph{logical}, indicating that the action only includes logical operations.}
An action is applicable in a state if its precondition is satisfied by the state.

\Revised{$\mathcal{B}$ is an interval to constrain numeric parameters, which is defined by $\mathcal{B} = [\underline{\mathcal{B}}, \overline{\mathcal{B}}]$. We use two vectors $\underline{\mathcal{B}} = \langle \underline{\beta}^1,\underline{\beta}^2, \ldots, \underline{\beta}^T \rangle$ and $\overline{\mathcal{B}} = \langle \overline{\beta}^1,\overline{\beta}^2, \ldots, \overline{\beta}^T \rangle$ to denote lower bounds and upper bounds of parameters, respectively. For each parameter $\theta^j\in\Theta$, we assume it is constrained by a lower bound $\underline{\beta}^j$ and an upper bound $\overline{\beta}^j$. It is noted that if a parameter is not limited by any upper bounds, we let $\overline{\beta}^j = \infty$. Similarly, if a parameter is not limited by any lower bounds, we let $\underline{\beta}^j = -\infty$. Different from previous approaches with control parameters, {\ours} is not just based on searching graphs constructed by the bounds of control parameters. {\ours} uses gradient descent to accomplish a finer searching. \ignore{Such that}\MajorRevision{Thus,} {\ours} can handle numeric parameters without being constrained by bounds.
}

Besides,
we use $\psi(a)$ to denote the cost of action $a$. In this paper, \ignore{we focus on benchmarks about path planning, so }we define the cost of an action as the distance yielded by it. We define the cost of a plan $\plan$ as the sum of the cost of all actions in $\plan$, i.e., $C(\plan)=\sum_{a_i \in \plan}\psi(a_i)$.

Given a mixed planning problem $\mathcal{M}$, we aim at computing a plan $\plan = \langle a_1,a_2,\dots,$ $a_n \rangle$, which achieves $g$ from $s_0$ with the minimal cost. Different from classical planning problems, in this paper, we require to not only find appropriate actions but also determine their continuous numeric parameters to minimize the cost.


In this paper, we extend the syntax of PDDL 2.1 \cite{fox2003pddl2} to express our mixed planning problem. Specifically, we use ``\textbf{event}'' to define events as PDDL+  \cite{DBLP:journals/jair/FoxL06} does, events are instantaneous actions which happen as long as their preconditions are met. Events in our paper are similar to the format of action models. Once the preconditions of an event are satisfied, it happens immediately and its effects turn to be true. Similar to numeric planning with control parameters, we use ``\textbf{parameters-bound}'' to define bounds of numeric parameters.
\MajorRevision{Bounds of a parameter aim at restricting value of the parameter into an interval.}It is noted that {\ours} makes no use of the bounds during the gradient descent to compute numeric parameters. Hence, the bounds can be infinite.

\MajorRevision{
Specifically, a mixed planning problem is defined by two parts, a domain file and a problem file. Domain file defines an universal description of the environment, including a domain name, types of objects, predicates, functions, and descriptions of action models. ``:types'' defines what objects can form the parameters of actions and predicates. Notably, in our mixed planning problem, a parameter can be either an object or a numeric value. \emph{For example, ``ROV-Navigate (ship ROV -1 1 1)'' indicates an action about a movement of ``ROV'', which is deployed from ``ship''. ``ship'' and ``ROV'' are objects and their types are ``vehicle''. ``-1'' is a numeric parameter, defined by ``$\mathbb{R}$''.} ``:predicates'' defines propositions by declaring their predicate names and the types of objects they use, e.g., ``(can-move ?v - vehicle)''. Similarly, ``:functions'' defines numeric variables by presenting the names of variables and the types of objects, e.g., ``(location-x ?v - vehicle)''. An action model defines a transformation about updating states, including a name, types of parameters, preconditions and effects. To define an action model introduced by ``:action'', we use ``:parameters'' to define types of parameters, which can be used when executing this action. ``:precondition'' defines a set of preconditions, where each precondition is a condition which must be met in order to execute the action, which can either ask a proposition to be true or false, or require a variable to satisfy a constraint. ``:effect'' is a set of effects, where an effect is to define a proposition should be added into or deleted from the state, or update the value of a variable. On the other hand, problem file includes the details of an instance. It defines existing objects in the instance by ``:objects''. Initial state and goal state are defined by ``:init'' and ``:goal'', respectively. Optionally, problem file can define the bounds of numeric parameters and an objective function. The bounds are global constraints requiring numeric parameters to satisfy, defined by ``:parameters\_bounds'', such as ``-5 $\leq$ ?v$_x$ $\leq$ 5''. Otherwise, the numeric parameter can be any real-valued if not being specifically requested, i.e., ``-$\infty$ $\leq$ ?v$_x$ $\leq$ $\infty$''. Objective function is defined by
``:metric'', which behaves like an optimisation function, it defines a cost value to be minimized or maximized for a plan.

\begin{figure*}[!ht]
 \setlength{\abovedisplayskip}{1pt}
 \setlength{\belowdisplayskip}{1pt}
  \centering
  \includegraphics[width=0.98\textwidth]{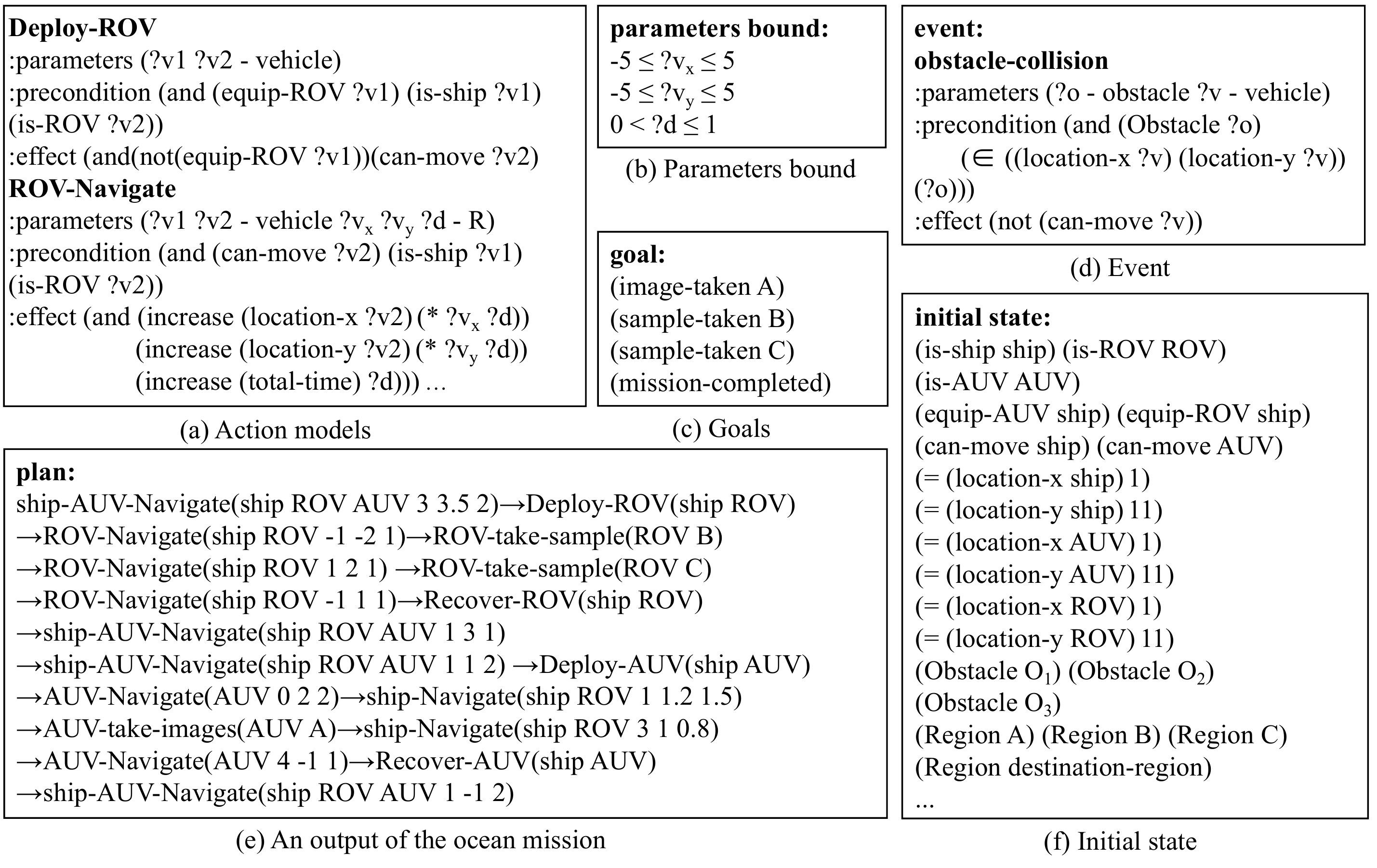}
  \caption{\MajorRevision{An example of the mixed planning problem on ocean mission.}}
  \label{problem_definition:example}
\end{figure*}

Intuitively, we show an example of a mixed planning problem in Figure \ref{problem_definition:example}. In Figure \ref{problem_definition:example}(a), ``Deploy-ROV (?v - vehicle)'' is a \ignore{discrete}\MajorRevision{logical} action model which does not have numeric effects, while ``ROV-navigate (?v1 ?v2 - vehicle $?v_x$ $?v_y$ $?d$ \MajorRevision{- $\mathbb{R}$})'' is a numeric action which has numeric effects such as ``(increase location-x ?v2(* ?$v_x$ ?$d$))''. Figure \ref{problem_definition:example}(b) defines bounds of three numeric parameters. Figure \ref{problem_definition:example}(c) is a goal composed of a set of propositions. Figure \ref{problem_definition:example}(d) describes an event that a collision happens when a vehicle $?v$ enters an obstacle region $?o$ (i.e., ($\in$ ((location-x ?v)(location-y ?v)) ($?o$))). Figure \ref{problem_definition:example}(f) is an initial state which is composed of both propositions and assignments. For example, ``(equip-AUV ship)'' indicates the ship equips a AUV at the beginning, ``(= (location-x ship) 1)'' indicates the initial x-axis location of the ship is 1, and ``(Obstacle O$_1$)'' is a assignment of an obstacle which must be avoided. Figure \ref{problem_definition:example}(e) is a solution plan of the mixed planning problem.

\MajorRevision{
Compared with PDDL 2.1 and PDDL+:
\begin{enumerate}
    \item The parameters of actions in {\ours} can be either objects or real numbers. In PDDL 2.1 and PDDL+, parameters can only be objects, which limits the ability of modeling numeric changes. \emph{Figures \ref{problem_definition:comparison_with_PDDL}(a) and (d) show action models about moving a ROV used in {\ours} and PDDL 2.1 respectively. Figure \ref{problem_definition:comparison_with_PDDL}(a) shows an action of a movement of a ROV. ROV ``?v2'' can move based on three numeric parameters ?v$_x$, ?v$_y$ and ?d if ``?v2'' is not equipped by ship ``?v1''. However, as shown in Figure \ref{problem_definition:comparison_with_PDDL}(d), the effects are discrete and each movement is fixed, defined by ``(at-x ?v ?loc)'' and ``(at-y ?v ?loc)''. }To model unfixed movements in PDDL 2.1 and PDDL+, one way is to use durative actions, whose numeric effects rely on linear functions determined by durations.
    \item In durative action models in PDDL 2.1 and PDDL+, although the introduction of durations improves the ability of expressing numeric changes, it is hard for them to model numeric effects which may have different transitions. Because effects of one action depend on the same numeric parameters, i.e., durations. \emph{As shown in Figure \ref{problem_definition:comparison_with_PDDL}(e), in order to describe the movements of ROV, we have to build more durative actions, because movements towards different directions are hard to be modeled by a general function. As shown in the figure, numeric effects are marked in purple and parameters of actions are marked in blue.}
    \item The numeric parameters in mixed planning problems can be limited by bounds or not, as shown in the red text in the Figure \ref{problem_definition:comparison_with_PDDL}(b). In PDDL 2.1 and PDDL+, owing to relying on durations, the constraints can limited by constraining durations, as shown in Figure \ref{problem_definition:comparison_with_PDDL}(e). It is noted that parameters bounds can help reduce the search space and runtime during searching plans. As for some planners, especially the planners based on mathematical algorithms such as linear programming and convex optimization, the values of bounds are closely associated with the ability to plan. If values are too small, search space are divided finely, resulting in a large search space which makes it hard to compute a plan. On the other hand, large values of bounds enlarge possible value scope of numeric parameters, which essentially increases the difficulty of problem solving, let alone when the bounds are infinite. Differently, mxPlanner makes no use of the bounds, it works whether bounds exist or not.
    \item We use ``event'', which are from PDDL+, to model predictable exogenous events such as a ROV will be damaged if it enters an obstacle region, as shown in Figure \ref{problem_definition:comparison_with_PDDL}(c). However, the descriptions of entering a region, which are marked in green, are unable to be described in PDDL+. On the other hand, in discretized search space, although a region can be indicated by enumerating all discretized values involved, as shown in the green proposition in Figure \ref{problem_definition:comparison_with_PDDL}(d), it has to ignore other real-valued locations.
\begin{figure*}[!ht]
  \centering
  \includegraphics[width=0.98\textwidth]{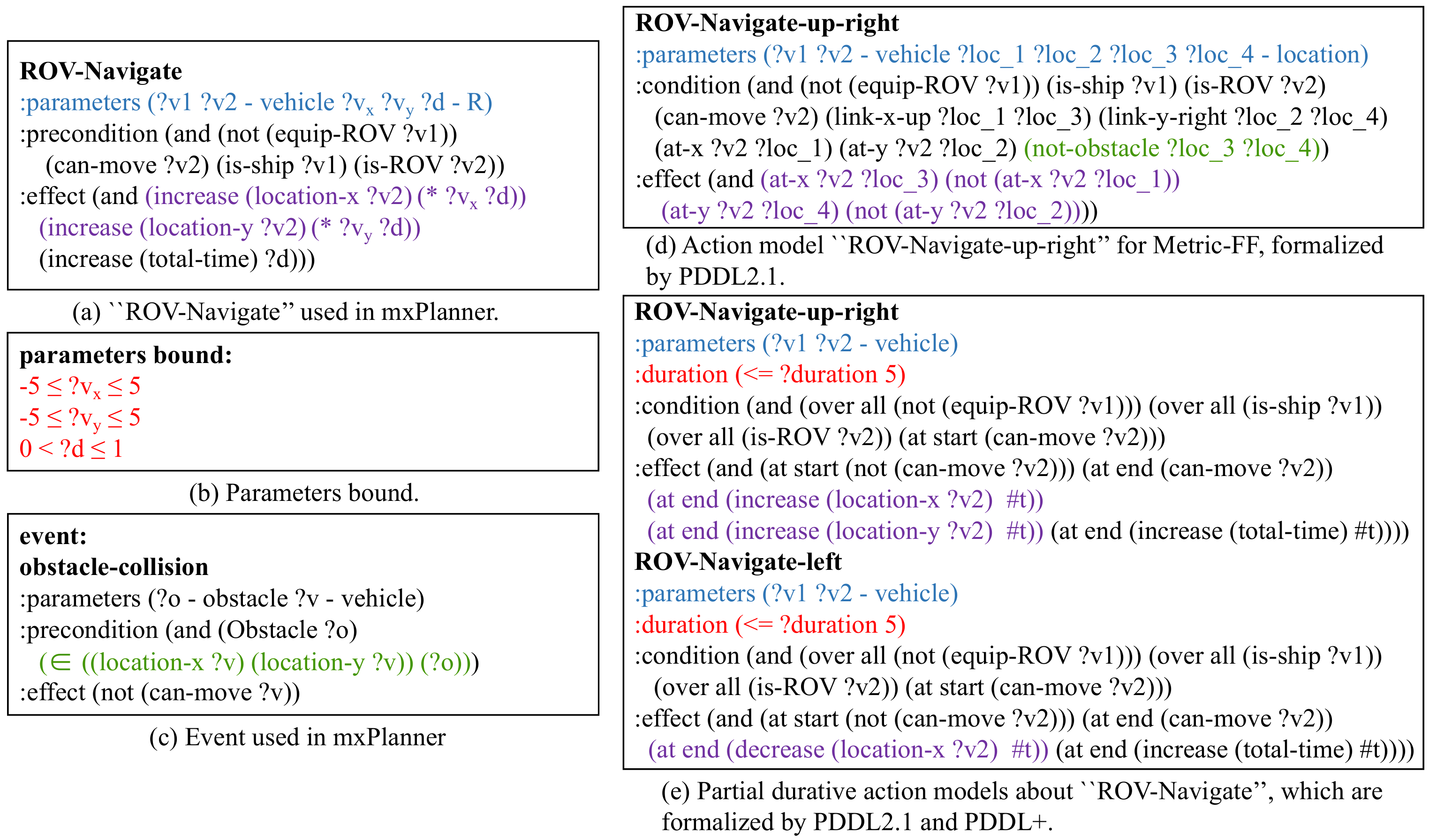}
  \caption{\MajorRevision{Comparison with PDDL2.1 and PDDL+.}}
  \label{problem_definition:comparison_with_PDDL}
\end{figure*}
\end{enumerate}
}

Based on the extensions of syntax, {\ours} aims at handling planning problems mixed with logical relations and numeric changes, where numeric changes can be determined by multiple parameters. Compared with numeric planners handling problems with control parameters, whose numeric effects are determined by durations and control parameters, {\ours} is not only able to solve these problems as well as computing exact values of parameters, but also can handle non-convex continuous space, i.e., containing obstacles. Intuitively, we compare our approach against the previous approaches in Table \ref{tab:comparison} according to the following four features:
\begin{itemize}
    \item \textbf{Non-discretization}: This feature indicates whether the corresponding approach is based on discretization or not.
    \item \Revised{\textbf{Mixed effects:} This feature indicates whether the corresponding approach handles problems mixed with discrete logical relations and numeric changes or not.}
    \item \textbf{Non-linear effects}: This feature indicates whether the corresponding approach can handle \MajorRevision{actions with non-linear numeric effects or not}.
    \item \Revised{\textbf{Non-convexity}:} This feature indicates whether the corresponding approach can handle non-convex continuous numeric space or not.
\end{itemize}
\begin{table}[!ht]
    \centering
    \caption{The comparison with previous approaches} \label{tab:comparison}
    \begin{tabular}{|l|l|p{1cm}|p{1cm}|p{1.2cm}|}
        \hline
         Approaches &  Non-discretization & \Revised{Mixed effects} & Non-linear effect & Non-convexity \\
         \hline
         Metric-FF \cite{DBLP:journals/jair/Hoffmann03}  & N & Y & N & N \\
         \hline
         LPRPG \cite{DBLP:conf/aips/ColesFLS08a} & N & Y & N & N\\
         \hline
         Kongming \cite{DBLP:conf/aips/LiW08}  & N & Y & N & Y \\
         \hline
         \Revised{UPMurphi \cite{DBLP:journals/apin/PennaMM12}} & N & Y & Y & N\\
         \hline
         COLIN \cite{DBLP:journals/jair/ColesCFL12}  & Y & Y & N & N \\
         \hline
         POPF \cite{DBLP:conf/aips/ColesCFL10} & Y & Y & N & N\\
         \hline
         OPTIC \cite{DBLP:conf/aips/BentonCC12} & Y & Y & N & N\\
         \hline
         POPCORN \cite{DBLP:conf/ecai/SavasFLM16}  & Y  & Y & N & N\\
         \hline
         \Revised{DiNo \cite{DBLP:conf/aaai/PiotrowskiFLMM16}} & N & Y & Y & N \\
         \hline
         \Revised{ENHSP \cite{DBLP:conf/ecai/ScalaHTR16}} & N & Y & Y & N\\
         \hline
         \Revised{SMTPlan+ \cite{DBLP:conf/aips/CashmoreFLM16}} & Y & Y & Y & N \\
         \hline
         Wu et al. \cite{DBLP:conf/nips/WuSS17}  & Y & N & Y  & Y \\
         \hline
         ScottyActivity \cite{DBLP:journals/jair/Fernandez-Gonzalez18}  & Y & Y & Y & N \\
         \hline
         OPTIC++ \cite{DBLP:conf/aips/DenenbergC19} & Y & Y  & Y & N\\
         \hline
         {\ours} & Y & Y & Y & Y \\
         \hline
    \end{tabular}
\end{table}

First, we compare the planners on the feature ``non-discretization''. Discretization means that the planner only considers discretized actions and the solution space is only composed of fixed discrete numeric variables.
In consequence, the planner cannot take into account the values that are between two neighbor discretized values.
It will significantly reduce the completeness of the planner. Also, we compare them on the features about the application scopes of planners: ``Mixed effects'', ``Non-linear effect'' and ``Non-convexity''.
The more features the planners possess, the larger the application scopes are.
The results are shown in Table \ref{tab:comparison}, where ``Y'' indicates the corresponding feature is \emph{true}, while ``N'' indicates the corresponding feature is \emph{false}. For example, the feature ``Non-discretization'' of {\ours} is ``Y'', indicating {\ours} is not based on discretization of the continuous search space.
From the table, we can see that only our {\ours} approach allows all of the above-mentioned four features to be true.

}

\section{Our {\ours} Approach}
In this section, we address our {\ours} approach in detail. We show a framework of our {\ours} in Figure \ref{figure:overview} in the form of unfolded RNN Cells. Each RNN Cell represents a step in a plan, which is composed of a heuristic module, a transition module, and a loss module. The heuristic module is an algorithm to output an action, the transition module is used to update a state according to action models, and the loss module is composed of functions to calculate losses. It is notable that \MajorRevision{we just ``borrow'' the framework of RNN and take advantage of its features about sharing the same RNN Cell across all steps. The heuristic module and the transition module are implemented as algorithmic procedures or functions instead of neural networks. Using RNN framework offers an efficient way to build sequential plans via gradient descent. Note that, different from RNNs, we optimize the input numeric parameters with gradient descent, rather than weights of neural networks in RNN Cells, which have been replaced by three modules in this paper.} \ignore{the heuristic module and the transition module are not implemented by neural networks. \Revised{We just ``borrow'' the framework of RNN}}We assume the number of RNN cells, $N$, is sufficiently large to compute a plan. The overall procedure of computing a plan is as shown below:
\begin{figure}[!ht]
  \centering
  \includegraphics[width=0.98\textwidth]{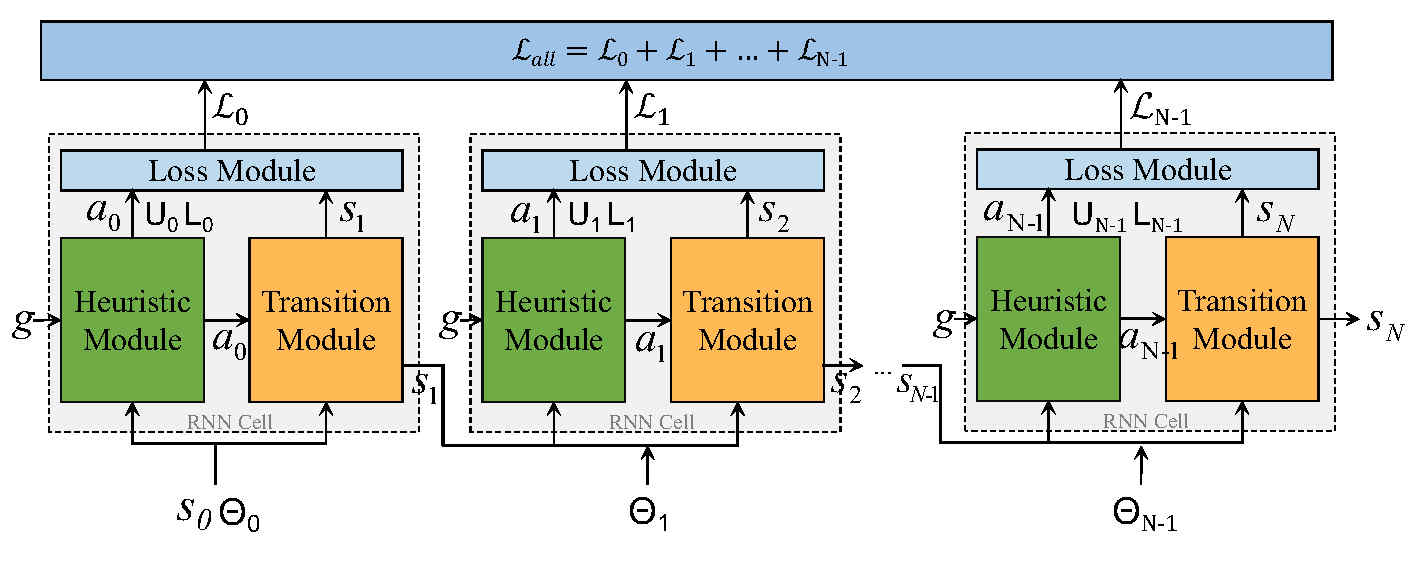}
  \caption{\Revised{The Gradient-based framework of {\ours}, which follows the RNN framework to perform forward simulation.}}
  \label{figure:overview}
\end{figure}
\begin{enumerate}
\item We first randomly initialize numeric parameters $\Theta$ of actions which will be optimized by the subsequent steps.
\item The heuristic module takes a state $s_i$, numeric parameters $\Theta_i$, and the goal $g$ as inputs, and outputs an action $a_i$, upper bound vector $\ubounds_i$ and lower bound vector $\lbounds_i$ of numeric variable values. The upper bound vector and lower bound vector are two real number vectors with the size of the number of numeric variables. These two bound vectors designate the value ranges of all numeric variables in the next state.
\item The transition module takes a state $s_i$, action $a_i$, and numeric parameters of actions $\Theta_i$ in $a_i$ as inputs, and outputs the next state $s_{i+1}$.
\item The loss module takes an action $a_i$, upper bound vector $\ubounds_i$, lower bound $\lbounds_i$ and state $s_{i+1}$ as inputs, and outputs the loss of the step $\mathcal{L}_i$. We calculate the total loss $\mathcal{L}_{all}$ by accumulating losses from all of the RNN Cells.
\item We inversely optimise numeric parameters in each step $\Theta_0, \Theta_1, \ldots, \Theta_{N-1}$ by minimizing $\mathcal{L}_{all}$.
\Revised{It is noted that, after updating numeric parameters, the transferred states at each step are also updated in the next iteration, along with the numeric parameters updated. Hence, the actions computed by the heuristic module may be different in the following iterations. With the values of numeric parameters updated, the heuristic module repeatedly estimates actions based on current updated states until a plan with the minimal cost is achieved.}
\item We repeat the second to the fifth steps until we find a valid plan $\plan$. Note that actions in $\sigma$ are not fixed and they can be changed during iterations. That is, we do not only update numeric parameters, but also update actions at the same time.
\end{enumerate}

In the following subsections, we will introduce in detail about the heuristic module, the transition module, the loss module and the inverse optimization procedure.

\subsection{Heuristic Module}
The heuristic module aims to find an appropriate action towards the goal and to predict the value range of numeric variables in the next state. We build the heuristic module based on the relaxed planning graph \cite{DBLP:journals/ai/BlumF97} and the interval-based relaxation \cite{DBLP:conf/ecai/ScalaHTR16}. Relaxed planning graphs are widely applied in heuristics searching planning approaches and the interval-based relaxation is used to capture the numeric effects of actions. \Revised{We first give some simple backgrounds about relaxed planning graphs and interval-based relaxation.}

\Revised{
Relaxed planning graph is built form a state $s$ only containing propositions. It contains actions and propositions that are possibly reachable from state $s$. In other words, the graph doesn't include unreachable actions or propositions. A relaxed planning graph is composed of interlaced nodes at state-level and nodes at action-levels. Nodes at state-level are propositions possibly being true. Nodes at action-level are actions that might be possible to be executed. Two nodes are connected by an edge, indicating preconditions or effects. A relaxed planning graph is first started from a level of state nodes, each node is a proposition in state $s$. Then a level of action nodes follows, each node is an action whose preconditions are satisfied in $s$. If a node at the first state-level is a precondition of a node at the first action-level, we link them by an edge. Then we add all effects of actions at the first action-level into $s$, and regard the updated propositions as the nodes at the second state-level nodes. If a state node at the second state-level is an effect of an action at the first action-level. We connect the state node and action node by an edge. We repeat the procedure until all goal propositions are in state nodes at the same state-level.

Interval-based relaxation is an extension of the principle of monotonic relaxation to numeric planning. In the interval-based relaxation, each numeric variable is defined by an interval, representing the set of values that the variable can have. The interval can be extended following a series of rules, which do not only formulate simple binary operations, but also complex mathematical functions.

\begin{figure}[!ht]
    \centering
    \includegraphics[width=0.99\textwidth]{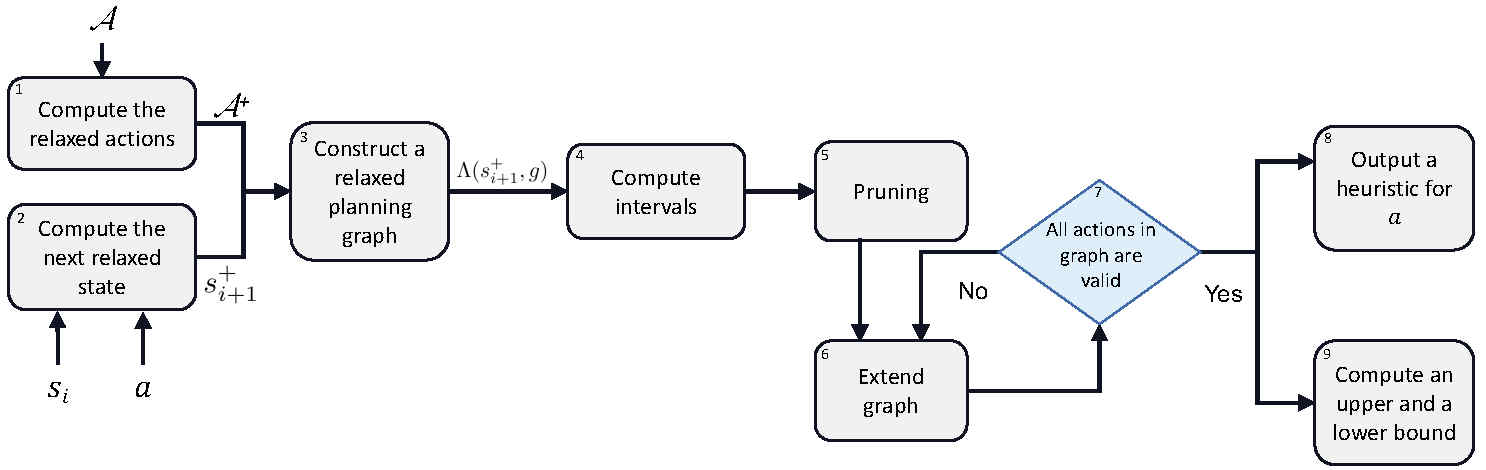}
    \caption{\MajorRevision{The procedure of the heuristic module.}}
    \label{fig:heuristic_procedure}
\end{figure}

Formally, given the current state $s_i$ and goal $g$, the heuristic module $\mathcal{H}$ selects an appropriate action $a_i$ which leads to a state $s_{i+1}$ with the minimal heuristic value, an upper bound vector $\ubounds_i$ and a lower bound vector $\lbounds_i$ which make up the intervals that all numeric variables in $s_{i+1}$ should satisfy. As shown in Figure \ref{fig:heuristic_procedure}, we first introduce a set of relaxed actions and update state $s$ to the next relaxed state $s^+_{i+1}$ with an action $a$ (Boxes 1 and 2 in Figure \ref{fig:heuristic_procedure}). Then we construct a relaxed planning graph and compute intervals based on the graph (Boxes 3 and 4 in Figure \ref{fig:heuristic_procedure}). After that, we revise the graph by pruning and repeatedly extending actions (Boxes 5, 6 and 7 in Figure \ref{fig:heuristic_procedure}). When all relaxed actions in the graph are valid, we compute a heuristic value for action $a$, an upper bound and a lower bound (Boxes 8 and 9 in Figure \ref{fig:heuristic_procedure}).
}

For a state $s_i$ and one of its applicable actions $a$ whose numeric parameters are given by $\Theta_i$, we define the next relaxed state $s^+_{i+1}$ by adding the positive effect of $a$ and updating numeric variables with numeric effects $\mathit{eff}^{n}$($a$) computed by $\Theta_i$. Then we introduce a set of relaxed actions, denoted by $\Actions^+$, which are obtained from actions by removing the negative effects and numeric preconditions and effects. 
Following Graphplan \cite{DBLP:journals/ai/BlumF97}, based on the relaxed action set $\Actions^+$, we construct a relaxed planning graph from $s^+_{i+1}$ to $g$, denoted by $\Lambda(s^+_{i+1},g)$. As the relaxed action set does not contain any numeric preconditions and effects, we need to extend the planning graph.
For a planning graph $\Lambda(s^+_{i+1},g)$ which has $h$-level, we use $\Actions_n(s^+_{i+1},g)$ and $\props_n(s^+_{i+1},g)$ to denote the sets of actions and propositions in the $n$-th level at $\Lambda(s^+_{i+1},g)$.
Specifically, we conduct the following steps to extend $\Lambda(s^+_{i+1},g)$:
\begin{itemize}
\item 
Following the interval-based relaxation \cite{DBLP:conf/ecai/ScalaHTR16},
\Revised{we restrict the value of each numeric variable in a state within a real number interval and extend the interval according to the numeric effects of actions and the parameter bounds $\mathcal{B}$. We use $\intervals_n(s^+_{i+1},g)$ to denote the intervals in the $n$-th level. Noted that $\mathcal{B}$ are used to estimate the number of actions which are possibly to be executed, rather than divided the continuous search space to construct graphs.
}

\item We then remove the redundant actions $a'$ from $\Lambda(s^+_{i+1},g)$. We consider an action $a' \in \Actions_m(s^+_i,g)$ is redundant, if its positive effects do not contain any goal and any proposition that belongs to the preconditions $\mathit{pre}^p(a'')$ of actions in $\Actions_n(s^+_i,g)$ with $n > m$. 
We repeat this step until there is no such redundant action in $\Lambda(s^+_{i+1},g)$.

\item It is possible that there exists an action $b$ whose numeric preconditions are not satisfied. 
That is, there exists a numeric variable $v$ related to a numeric precondition of $b$ in $\Actions_n(s^+_{i+1},g)$ that is disjoint with $\intervals_n(s^+_{i+1},g)$.
Then we add all relaxed actions $\Actions^+(v)$ in $\Actions^+$, which are applicable to $\props_{n-1}(s^+_{i+1},g)$ and include a numeric effect related to $v$,
into $\Actions_{n-1}(s^+_i,g)$. 
After that, we update $\props_{n}(s^+_i,g)$ and $\intervals_{n}(s^+_i,g)$.
If the numeric precondition about $v$ of $b$ is not satisfied, we repeatedly add $\Actions^+(v)$ into $\Actions_{n-1}(s^+_i,g)$, until it is satisfied.
As each interval is expanded monotonically, the numeric precondition finally will be satisfied.
We repeat this step until all numeric preconditions of all actions in $\Lambda(s^+_{i+1},g)$ are satisfied.
\end{itemize}

Intuitively, the actions in the extended relaxed planning graph can make up a valid plan. Thus, we consider the number of actions in $\Lambda(s^+_{i+1},g)$ as the heuristic value of the relaxed state $s^+_{i+1}$. For a current state $s_i$, from all actions applicable in $s_i$, we select an action $a_i$ which leads to the relaxed state $s^+_{i+1}$ with the least heuristic value.

\begin{figure}[!ht]
    \centering
    \includegraphics[width=0.98\textwidth]{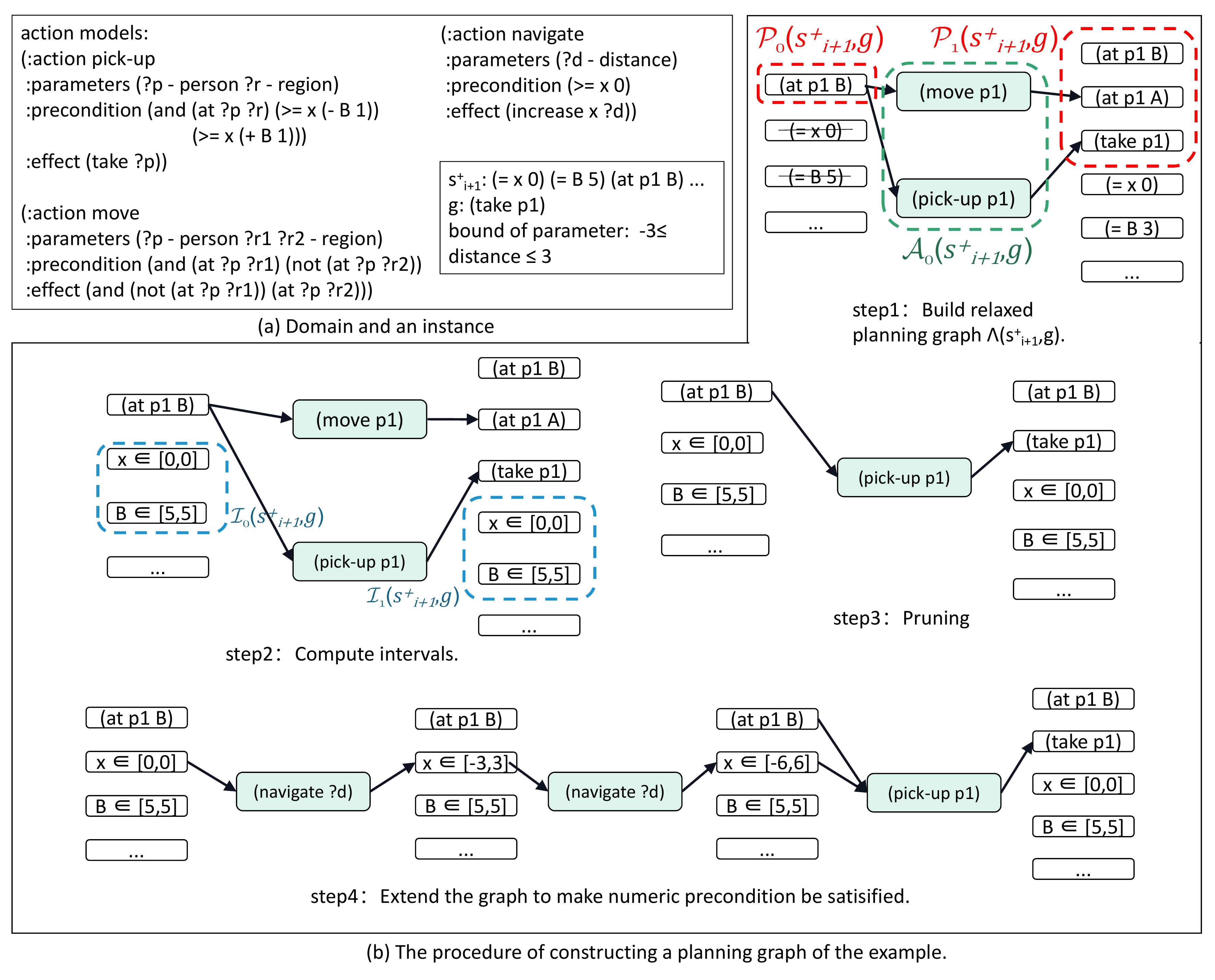}
    \caption{\MajorRevision{An example for calculating heuristic value.}}
    \label{fig:heuristic_example}
\end{figure}

\Revised{
\emph{An example is shown in Figure \ref{fig:heuristic_example} where Figure \ref{fig:heuristic_example}(a) enumerates a simple domain containing three action models and an instance. Numeric parameter ``distance'' is limited by a bound. In step 1 of Figure \ref{fig:heuristic_example}(b), we first ignore numeric variables, effects and preconditions, and we construct a relaxed planning graph $\Lambda(s^+_{i+1},g)$ based on Graphplan. The propositions at the first and second state-level are denoted by $\props_0(s^+_{i+1},g)$ and $\props_1(s^+_{i+1},g)$, which are framed by red squares. The actions in the first action-level are denoted by $\Actions_0(s^+_{i+1},g)$, which are framed by a green square. As shown in step 2, we rewrite variables by intervals and extend the intervals following the Interval-based relaxation based on $\Lambda(s^+_{i+1},g)$. Similarly, we denote them by $\intervals_0(s^+_{i+1},g)$ and $\intervals_1(s^+_{i+1},g)$ and use blue squares to frame them. In step 3 of Figure \ref{fig:heuristic_example}(b) we remove the redundant actions from $\Lambda(s^+_{i+1},g)$ whose effects are not required in goal or any actions at following action-levels. At last, in step 4 of Figure \ref{fig:heuristic_example}(b) we extend the graph by adding actions into $\Lambda(s^+_{i+1},g)$ until all numeric preconditions of actions are satisfied. And the heuristic value is the number of actions in $\Lambda(s^+_{i+1},g)$, which is 3 for the example.
}

}

To guide the computation of parameters, we also estimate an upper bound vector $\ubounds_i = \langle u^1_i, u^2_i, ..., u^K_i  \rangle$ and a lower bound vector $\lbounds_i = \langle l^1_i, l^2_i, ..., l^K_i  \rangle$ of numeric variables at the next state $s_{i+1}$ updated by action $a_i$ we have chosen. Generally, the actions occurring in the smaller levels of the planning graph are more probably to be selected in the beginning of a valid plan than those actions in the bigger levels. Thus, we should guarantee the numeric preconditions of such actions in the smaller levels to be satisfied in priority.
For every numeric variable $v^k$ in $\nvars$, we compute its upper bound $u^k_i$ and lower bound $l^k_i$ in $s_{i+1}$ as follows:
\begin{itemize}
    \item Given an extended relaxed planning graph $\Lambda(s^+_{i+1},g)$, we find the actions whose preconditions relate to $v^k$ and which occur at the smallest level, denoted by $\Actions(v^k)$.

    \item Suppose $\hat{a} \in \Actions(v^k)$ has a numeric precondition $[\hat{l},\hat{u}]$ related to numeric variable $v^k$. For the current state $s_i$, we denote the distance to the numeric precondition of $\hat{a}$ as $||v^k_i - \frac{\hat{l}+\hat{u}}{2}||$, where $v^k_i$ is the value of $v^k$ in $s_i$. For every numeric variable $v^k$, we choose the action $\hat{a}$ with the smallest distance w.r.t. $v^k$, and set its bounds in the next state $s_{i+1}$ by the numeric precondition of $\hat{a}$, i.e., $l^k_i = \hat{l}$ and $u^k_i = \hat{u}$.
\end{itemize}


\subsection{Transition Module}
\MajorRevision{The transition module essentially is a transition function $\gamma$ that transforms $s_i$ to $s_{i+1}$ according to an action $a_i$ with parameters, where numeric parameters are given by $\Theta_i$.}The transition function is designed according to action models, which are taken as input\Revised{, including logical updating and numeric changes.} Compared with the other planners about updating variables directly, we first compute numeric effects which is defined by $\Theta_i$ and $a_i$, then we update variables following the action model. Note that only the numeric parameters related to action $a_i$ are used. Formally, the transition module is defined as follows:
\begin{equation}
\setlength{\abovedisplayskip}{2pt}
\setlength{\belowdisplayskip}{1pt}
\label{equation:transition}
\Revised{s_{i+1} = \gamma (s_i, a_i; \Theta_i)}
\end{equation}
\emph{In the example of Figure \ref{problem_definition:example}, a ship navigates according to the parameters ``vel\_x'', ``vel\_y'', and ``duration''. An effect of \MajorRevision{``ROV-Navigate (?v1 ?v2 - vehicle $?v_x$ $?v_y$ $?d$ - $\mathbb{R}$)''} is ``(increase location-x ($*$ $?v_x$ $?d$))'' indicating that next x-axis of the robot is the product of $?v_x$ and $?d$ plus current x-axis. Assuming the axis of robot location-x $=$ 0, and numeric parameters $\Theta_i=\langle 2, -2, 1\rangle$, the x-axis of ROV will be updated into 2 after \MajorRevision{``ROV-Navigate (ship ROV 2 -2 1)''}.  }

\MajorRevision{
To construct Equation (\ref{equation:transition}) from domain file, we first instantiate all actions by enumerating objects which are able to form actions, and define each grounding action $a$ by a $X$-dimensional one-hot vector $\Vec{a}$, where $X$ is the number of grounding actions. We denote a set of grounding actions by $\ddot{\Actions} = \langle a_0, a_1, \dots, a_{X-1} \rangle$. \emph{Taking Figure \ref{fig:heuristic_example} as an example, there are 5 grounding actions, i.e., ``pick-up (p1 A)'', ``pick-up (p1 B)'', ``move (p1 A B)'', ``move (p1 B A)'' and ``navigate (?d)''. Vectors of ``pick-up (p1 A)'' and ``navigate (?d)'' are (0,0,0,0,1) and (1,0,0,0,0) respectively.} Specifically, due to containing propositions and variables ($s_i = \langle \props_i, \nvars_i \rangle$), transition function $\gamma(.)$ is composed of logical operational updating $\gamma_l(.)$ and numeric updating $\gamma_n(.)$. Equation (\ref{equation:transition}) can be updated by Equation (\ref{equation:transition_extension}):
\begin{align}
\notag
s_{i+1} &= \gamma (\langle  \props_{i}, \nvars_{i}  \rangle, a_i; \Theta_i) \\
&= \langle \gamma_l (\props_i, \Vec{a}_i), \gamma_n (\nvars_i, \Vec{a}_i;\Theta_i) \rangle
\label{equation:transition_extension}
\end{align}

Logical operational updating takes an action $a_i$ and current propositional state $\props_i$ as input, and computes next propositions $\props_{i+1}$, which is defined by:
\begin{align}
\notag
\props_{i+1} &= \gamma_l (\props_i, \Vec{a}_i) \\
&= \Vec{a}_i\mathsf{E}_l + ( \Vec{1} - (\Vec{a}_i\mathsf{E}_l)^2)\props_i
\label{equation:transition_discrete}
\end{align}
where $\Vec{a}_i$ is the one-hot vector of action $a_i$, and $\mathsf{E}_l \in \mathbb{R}^{X \times M} $ indicates propositional effects of all actions. Each row $\mathbf{e}_x = \langle \mathbf{e}_x^1, \ldots, \mathbf{e}_x^M \rangle$ of $\mathsf{E}_l = [\mathbf{e}_0, \mathbf{e}_1, \ldots, \mathbf{e}_{X-1}]$ is a vector indicating propositional effects of an grounding action $a_i$ (i.e., $\mathit{eff}^+(a_i)$ and $\mathit{eff}^-(a_i)$). If $\mathbf{e}_x^m = 1$, one of positive literals in $\mathit{eff}^+(a_i)$ is to add the proposition into state. If $\mathbf{e}_x^m = -1$, one of positive literals in $\mathit{eff}^-(a_i)$ is to delete the proposition from state. Otherwise, neither $\mathit{eff}^+(a_i)$ nor $\mathit{eff}^-(a_i)$ updates this proposition. Equation (\ref{equation:transition_discrete}) aims at only updating propositions corresponding to propositional effects and ensuring the others to be unchanged.

Numeric updating is to update variables $\nvars_i$ in current state $s_i$ according to an action $a_i$, which is defined by Equation (\ref{equation:transition_numeric}), where $\kappa \in \mathbb{R}^{X\times K}$ indicates whether an action increases or decreases the value of a variable. If an effect $\mathit{eff}^n(a_x)$ of grounding action $a_x$ is to increase the value of variable $v^k$, we let $\kappa_x^k = 1$. Otherwise, $\kappa_x^k = -1$.
\begin{align}
\label{equation:transition_numeric}
\notag
\nvars_{i+1} &= \gamma_n (\nvars_i, \Vec{a}_i;\Theta_i) \\
&= \nvars_i + \Vec{a}_i(\kappa\odot(\mathsf{E}_{ind} + \mathsf{E}_{d}))
\end{align}
Generally, numeric updating includes parameters-independent numeric updating (i.e., to increase or decrease a constant or a value computed by a function based on variables) and parameters-dependent numeric updating (i.e., to increase or decrease a value computed by functions based on numeric parameters). We use $\mathsf{E}_{ind} \in \mathbb{R}^{X \times K}$ and $\mathsf{E}_{d} \in \mathbb{R}^{X \times K}$ to denote them, respectively. Each element $\dot{\mathbf{e}}_x^k \in \mathsf{E}_{ind}$ indicates the value that is decided by constants or current variables $v_i$, being used to update variables $v_i^k$ by the $x$th grounding action. Each $\hat{\mathbf{e}}_x^k \in \mathsf{E}_{d}$ indicates the value that is decided by numeric parameters $\Theta_i$, which is defined by:
\begin{align}
\label{equation:transition_numeric_parameters}
\hat{\mathbf{e}}_x^k &= f_x^k(\Theta_i,\nvars_i)
\end{align}
where $f_x^k(.)$ is a function based on effects of the $x$th grounding action which updates variable $v^k$. Note that $\nvars_i$ is not required. \emph{Taking Figure \ref{fig:heuristic_example} as an example, for variable ``x'', action ``navigate (?d)'' has an effect ``(increase x ?d)'' to update it. In this example, there is only one numeric parameter, ``distance''. Assuming $?d = 1$, we let $\hat{\mathbf{e}}_x^k = 1$.}}

\subsection{Inverse Optimization of Parameters}
To calculate an optimal plan, we repeatedly update action parameters $\Theta$ via gradient descent. In our framework, the model of the RNN cells \MajorRevision{is not formed from neural networks, but algorithmic modules which is fixed when updating $\Theta$}
\Revised{We view the parameters $\Theta$ as the inputs of the framework at each step, and name the procedure of updating the input as an \emph{Inverse} optimization, which is different from RNN approaches that aim at learning parameters of models.}
Similar work on inverse optimization has been demonstrated effective by \citet{DBLP:conf/nips/WuSS17}, which \Revised{ made use of the framework of RNNs and calculated continuous action sequences via optimizing the input.} Compared to their work, our problem is more challenging since we need to consider logical relations and symbolic heuristic information when computing solution plans, \Revised{while they focus on discrete and continuous numeric change and cannot handle propositions.}
\MajorRevision{
\emph{For example, in the Navigation domain as shown in Figure \ref{figure:compare_wu_example},
an agent starts from an initial location, aiming at reaching a goal destination $g$ with the purpose of maximizing a reward function. The domain only contains a moving action whose transition function is shown in Figure \ref{figure:compare_wu_example}(b), where $d_t$ is the distance from location $s_t$ to deceleration zone $z$ in the middle of the map, $\lambda$ is a velocity reduction factor which is determined by $d_t$. The initial state $s_0$ is a two-dimensional location of the agent. The goal is to maximize the reward, i.e., to minimize the Manhattan distance from the agent to the goal location $g$, as shown in Figure \ref{figure:compare_wu_example}(c). When the approach terminates, it returns a sequence of two-dimensional vector, each of which indicates a movement on x-axis and y-axis, as shown in Figure \ref{figure:compare_wu_example}(d). As shown in the figure, the method proposed by Wu et al. does not involve any logical relations.
In contrast, the description of {\ours} contains two action models, as illustrated in Figure \ref{figure:compare_wu_example}(e),
\begin{figure}[!ht]
\centering
\includegraphics[width=0.92\textwidth]{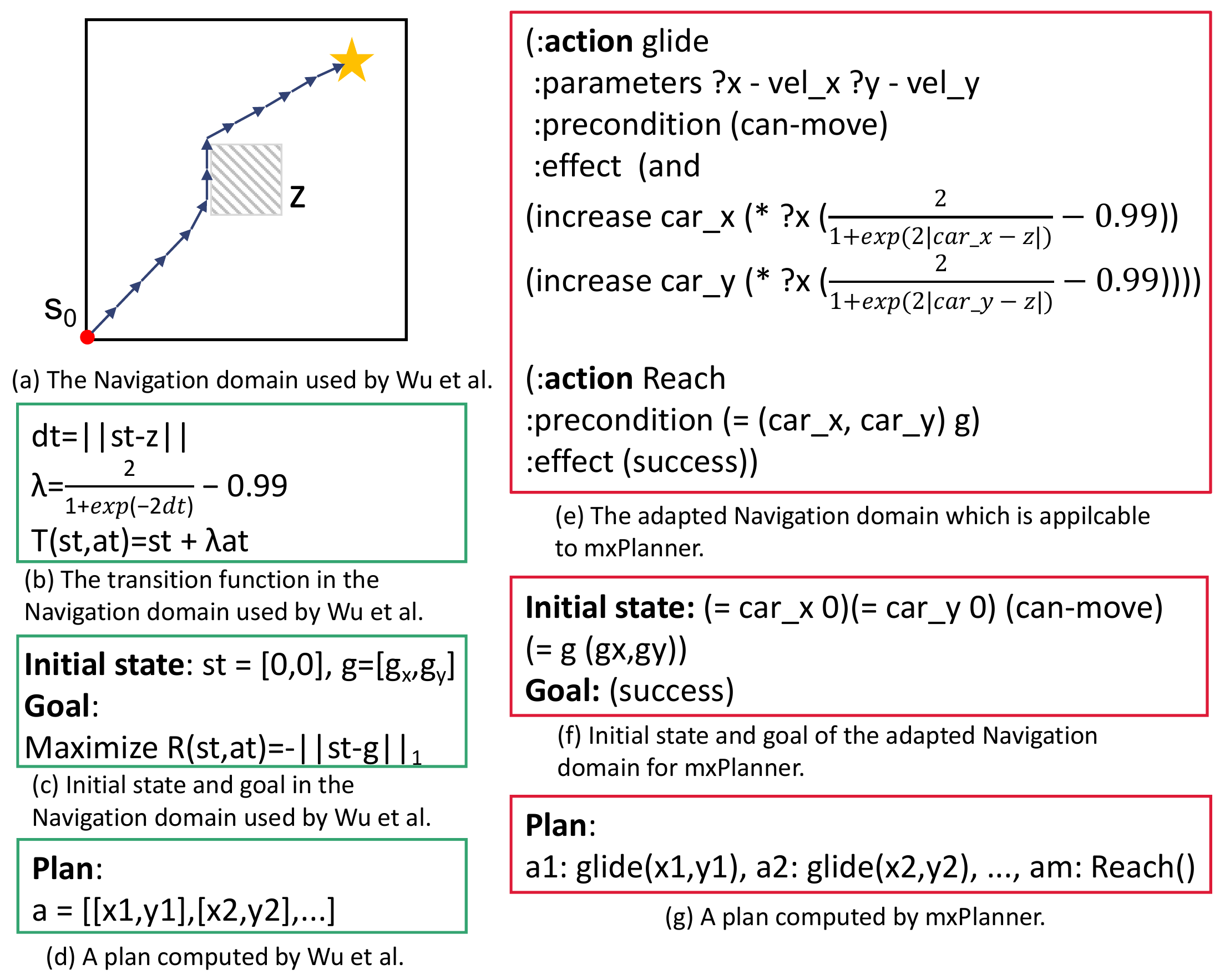}
\caption{\MajorRevision{The description of the Navigation domain from \citet{DBLP:conf/nips/WuSS17} and  {\ours}, respectively.}}
\label{figure:compare_wu_example}
\end{figure}
including numeric parameters (e.g., ``vel$\_$x''), propositions (e.g., ``(can-move)''), and preconditions, (e.g., ``(= (car$\_$x, car$\_$y) g)''). As shown in Figure \ref{figure:compare_wu_example}(f), the initial state is composed of numeric variables and propositions, and the goal is composed of a set of propositions. \MajorRevision{Finally,} {\ours} computes a plan reaching goals, as shown in Figure \ref{figure:compare_wu_example}(g), including action names and their parameters.}

}

To do the inverse optimization, we design a novel objective function with three different losses, as shown in Equation (\ref{equation:loss_instantaneous}):
\begin{align}
\setlength{\abovedisplayskip}{1pt}
\setlength{\belowdisplayskip}{1pt}
    \mathcal{L}_i = ~&w_1\mathcal{L}_{b_i}+ w_2\mathcal{L}_{o_i} + w_3\psi(a_i)
    \label{equation:loss_instantaneous}
\end{align}
where $w_1,w_2$ and $w_3$ are hyperparameters. The three losses are shown as follows:
\begin{itemize}
\item $\mathcal{L}_{b_i}$ is a loss to let next state satisfy numeric bounds computed by heuristic module, defined by Equation (\ref{equation:loss_bound}):
\begin{align}
\setlength{\abovedisplayskip}{1pt}
\setlength{\belowdisplayskip}{1pt}
\mathcal{L}_{b_i} = ||\ReLu(\nvars_{i+1} - \ubounds_i)||_2+ ||\ReLu(\lbounds_i - \nvars_{i+1})||_2
\label{equation:loss_bound}
\end{align}
where $\ReLu(x)=\maximum(0,x)$, $\nvars_{i+1}$ denotes values of numeric variables in state $s_{i+1}$. In heuristic module, $\lbounds_i$ and $\ubounds_i$ were calculated to require variables $\nvars_{i+1}$ satisfy $\lbounds_i\leq\nvars_{i+1}\leq\ubounds_i$. Once a numeric variable $v_{i+1}^k$ ($1 \leq k \leq K$) in state $s_{i+1}$ exceeds its upper bound $\ubounds_i^k$, a loss is generated. The case for lower bounds is similar.

\item $\mathcal{L}_{o_i}$ is a loss for capturing the scenario of avoiding obstacles, which is defined by:
\begin{align}
\setlength{\abovedisplayskip}{-1pt}
\setlength{\belowdisplayskip}{1pt}
\label{equation:loss_obstacle}
\mathcal{L}_{o_i} =&\sum_{\alpha=1}^{M}m_{\alpha} ||y^\prime_{\alpha} - p_{i+1}||_2
\end{align}
where $m_\alpha$ is:
\begin{align}
\setlength{\abovedisplayskip}{1pt}
\setlength{\belowdisplayskip}{1pt}
m_{\alpha} = &
    \begin{cases}
      1, & \text{if $p_ip_{i+1} \cap O_{\alpha} \neq \emptyset$} \\
      0, & \text{otherwise}
    \end{cases}
\end{align}
where $p_i$ is the position of the agent in state $s_i$, \Revised{$p_{i+1}$ is the next position of the agent in state $s_{i+1}$, $p_ip_{i+1}$ indicates a line between $p_i$ and $p_{i+1}$, and} $y^\prime_\alpha$ is a selected target position that guides the agent to avoid obstacle $O_{\alpha}$ ($\alpha = 1, \dots, M$). To avoid $O_{\alpha}$, we define a convex cone whose vertex is $p_i$, covering $O_{\alpha}$. We use $Y_\alpha$ to denote the intersection of \Revised{vertices} in obstacle $O_{\alpha}$ and the convex cone. Let ${y}_\alpha \in Y_\alpha$ be the closest vertex to next position $p_{i+1}$ of the robot in next state $s_{i+1}$, i.e., ${y}_\alpha = \arg\min_{y \in Y_\alpha}||p_{i+1} - y||_2$. We define $y^\prime_{\alpha}$ as a position that satisfies $||y^\prime_{\alpha} - y_{\alpha}||_2 = \varepsilon$ where $\varepsilon$ is a small positive real number and $y^\prime_{\alpha}$ is not in the convex cone.
An example is shown in Figure \ref{figure:obstacle_example}. Intuitively, when the robot tries to go through the obstacle, $\mathcal{L}_{o_i}$ aims to guide its destination $p_{i+1}$ getting close to $y'_\alpha$ for getting rid of $O_\alpha$.
\begin{figure}[!ht]
  \centering
  \includegraphics[width=0.55\textwidth]{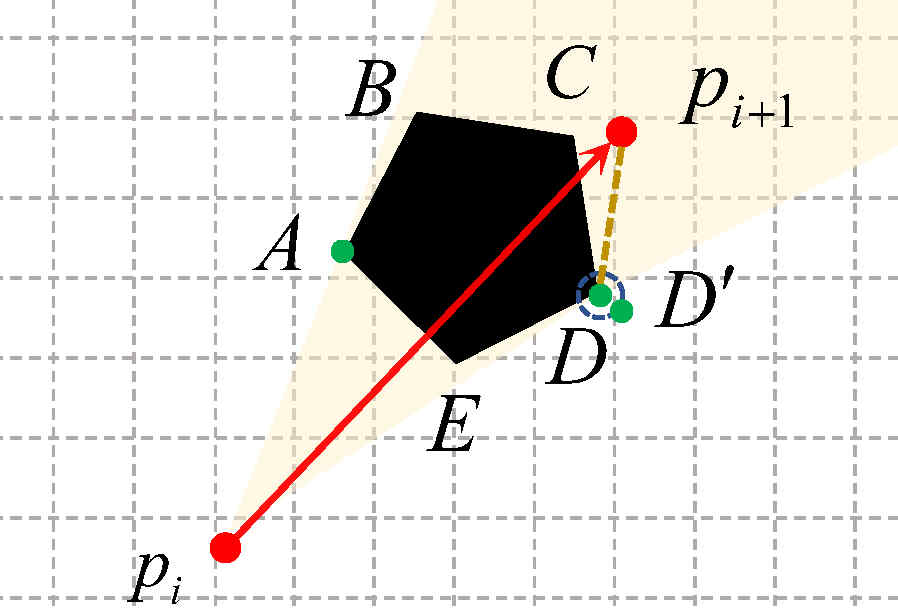}
  \caption{
  The dark pentagon ABCDE is an obstacle $O_\alpha$, $p_i$ is the current position and $p_{i+1}$ is  next position. The yellow area is the convex cone which intersects with $O_\alpha$ at A and D. $y_\alpha = D$ is closer to $p_{i+1}$. $D'$ is a position outside $O_\alpha$ that is $\varepsilon$ away from $D$ and can be seen as a candidate of $y'_\alpha$.
  }
  \label{figure:obstacle_example}
\end{figure}

\item $\psi(a_i)$ is the cost of $a_i$, which we view as the navigating distance in this paper. For example, the effects of \MajorRevision{``ROV-Navigate (?v1 ?v2 - vehicle $?v_x$ $?v_y$ $?d$ - $\mathbb{R}$)''} are ``(increase location-x ($*$ $?v_x$ $?d$))'', ``(increase location-y ($*$ $?v_y$ $?d$))'' and ``(increase total-time $?d$)'', so its cost is $\sqrt{(?d ?v_x)^2 + (?d ?v_y)^2}$.
\end{itemize}

We define the accumulated loss $\mathcal{L}_{all}$ as the sum of instantaneous losses until the goal is achieved, i.e.,
\begin{align}
\setlength{\abovedisplayskip}{-3pt}
\setlength{\belowdisplayskip}{-3pt}
    \mathcal{L}_{all} =~&\sum_{i=0}^{\mu-1}\mathcal{L}_i, \hspace{5mm}
    \text{s.t.} \hspace{2mm} a_\mu = end
    \label{equation:loss_all}
\end{align}
where $\mu$ is the final step where the goal is achieved and a$_\mu$ is a terminator indicating the goal is arrived.

We then compute the partial derivatives of the accumulated loss based on Equation (\ref{partial_derivative}), \Revised{as shown below:

\begin{align}
\setlength{\abovedisplayskip}{1pt}
\setlength{\belowdisplayskip}{1pt}
\label{partial_derivative}
  \notag
  \frac{\partial \mathcal{L}_{all}}{{\partial \Theta_i}}
  &= \sum_{\lambda=i}^{\mu-1}\frac{\partial \mathcal{L}_{all}}{\partial \mathcal{L}_{\lambda}} \frac{\partial \mathcal{L}_{\lambda}}{\partial \Theta_i}\\\notag
  &= \sum_{\lambda=i}^{\mu-1}(\frac{\partial \mathcal{L}_{all}}{\partial \mathcal{L}_{\lambda}} \frac{\partial \mathcal{L}_{\lambda}}{\partial \nvars_{\lambda +1}} \frac{\partial \nvars_{i+1}}{\partial \Theta_{i}} \prod_{k=i+1}^{\lambda} \frac{\partial \nvars_{k+1}}{\partial \nvars_{k}})\\
  &= \frac{\partial \nvars_{i+1}}{\partial \Theta_{i}} (
  \sum_{\lambda=i}^{\mu-1}(\frac{\partial \mathcal{L}_{all}}{\partial \mathcal{L}_{\lambda}}
  (\frac{\partial \mathcal{L}_{all}}{\partial \mathcal{L}_{b_\lambda}}
  \frac{\partial \mathcal{L}_{b_\lambda}}{\partial \nvars_{\lambda+1}} \!+\!\frac{\partial \mathcal{L}_{all}}{\partial \mathcal{L}_{o_\lambda}}
   \frac{\partial \mathcal{L}_{o_\lambda}}{\partial \nvars_{\lambda+1}} \!+\!
   \frac{\partial \mathcal{L}_{all}}{\partial \psi(a_i)}
   \frac{\partial \psi(a_i)}{\partial \nvars_{\lambda+1}} )
  \!\prod_{k=i+1}^{\lambda}\! \frac{\partial \nvars_{k+1}}{\partial \nvars_{k}}))
\end{align}
}

\begin{figure}[!ht]
    \centering
    \includegraphics[width=0.98\textwidth]{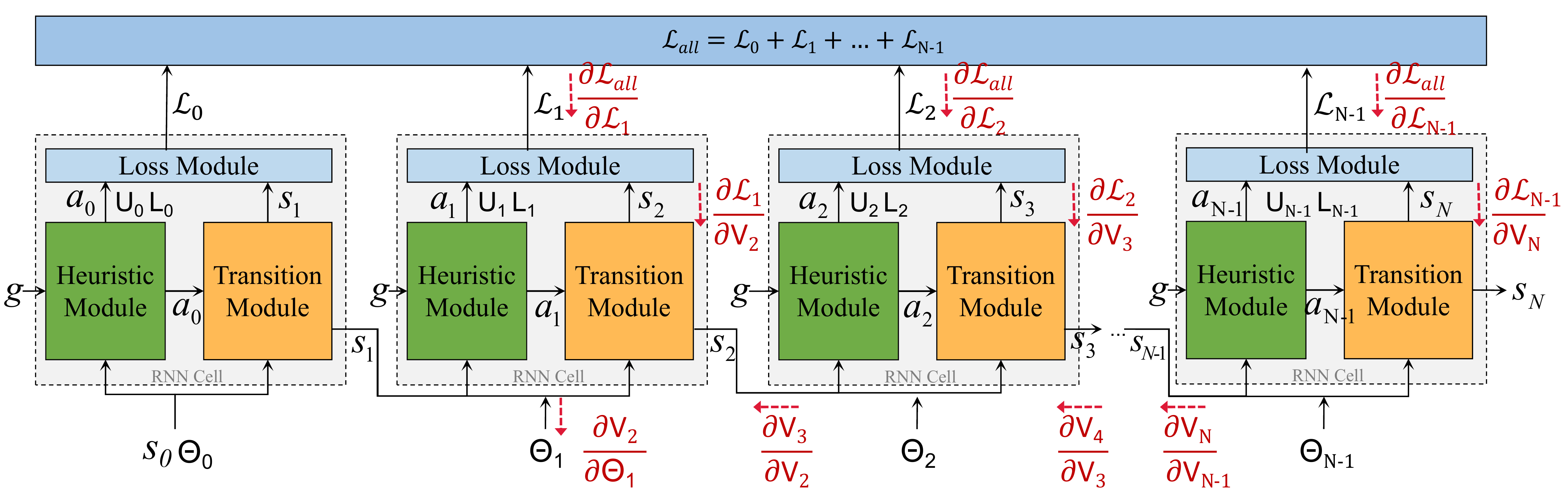}
    \caption{\MajorRevision{The procedure of back propagation in {\ours}.}}
    \label{fig:loss_derivative}
\end{figure}

An example of back propagation procedure \MajorRevision{ to compute $\frac{\partial \mathcal{L}_{all}}{{\partial \Theta_1}}$} is shown in Figure \ref{fig:loss_derivative}, where the gradient flow propagates across time steps. Intuitively, the gradient of $\Theta_i$ is determined by the loss from $\mathcal{L}_{i}$ to $\mathcal{L}_{\mu-1}$. The gradient of the numeric parameters irrelated to $a_i$ is zero.
\MajorRevision{
Note that, although a state is composed of propositions and variables, the loss $\mathcal{L}_{all}$ is determined by variables (according to Equation (\ref{equation:loss_bound}) and (\ref{equation:loss_obstacle})). Therefore, the partial derivatives and the procedures of back propagation focus on variables instead of the whole states. Via gradient descent, we then optimize the numeric parameters of each action in the plan to minimize the loss. After that we recompute a new plan based on the new heuristic module and transition module with the updated parameters. We iteratively optimize the numeric parameters and recompute new plans until the stop condition is satisfied.}

 The parameters $\Theta_i$ in each step are updated by the gradient from the total loss, as shown in Equation (\ref{update_parameters}):
\begin{equation}
\Theta_i = \Theta_i - \omega \frac{\partial \mathcal{L}_{all}}{\partial \Theta_i},\label{update_parameters}
\end{equation}
\Revised{where $\omega$ is a learning rate. We set $\omega = 0.001$ in this paper.}

\subsection{Overview of {\ours}}
An overview of {\ours} is shown in Algorithm \ref{algorithm:code}. We first initialize numeric parameters $\Theta$ of $N$ steps and build a heuristic module and a transition module to iteratively estimate action $a_i$ (Line 5 of Algorithm \ref{algorithm:code}) and update $s_{i+1}$ (Line 6) to attain a candidate plan $\plan$. After that we update the parameters $\Theta$ by optimizing $\mathcal{L}_{all}$ (Line 9 and 10). We repeat the above procedure until the stop requirement is satisfied \MajorRevision{or the running time exceeds the cutoff time} (Line 2) and output the solution plan $\plan$.

\begin{algorithm}[!ht]
\caption{{\ours}}
\label{algorithm:code}
\textbf{input:}  \Revised{$\mathcal{M} = \langle\Actions,\mathcal{S},s_0,g,\mathcal{B}\rangle$.}\\
\textbf{output:} $\plan$.
\begin{algorithmic}[1]
\STATE initialize numeric parameters in $N$ steps $\Theta = \langle \Theta_0,\dots,\Theta_{N-1} \rangle$ randomly;
\WHILE{$\mathcal{L}_{stop} \neq 0$ (Equation (\ref{equation:stop})) \MajorRevision{ $\And$ runtime $\leq$ cutoff time}}
\STATE $\plan = \langle  \rangle$;
\WHILE{i = 0, \dots, N-1}
\STATE \Revised{predict $a_i$, $\ubounds_i$, $\lbounds_i$ with $\Actions, s_i, g, \Theta_i$;}
\STATE update $s_{i+1}$ with $a_i$, $\Theta_i$ and $s_i$ (Equation (\ref{equation:transition}));
\STATE $\plan = [\plan|a_i]$;
\ENDWHILE
\STATE calculate accumulated loss $\mathcal{L}_{all}$ (Equation (\ref{equation:loss_all}));
\STATE update $\Theta$ by $\Theta_i = \Theta_i - \omega \frac{\partial \mathcal{L}_{all}}{\partial \Theta_i}$ (Equation (\ref{update_parameters}));
\STATE \Revised{update $\Theta$ by $\Theta = max(min(\Theta,\overline{\mathcal{B}}),\underline{\mathcal{B}})$ (Equation (\ref{bounds}));}
\ENDWHILE
\STATE return $\plan$;
\end{algorithmic}
\end{algorithm}

\Revised{
Particularly, to ensure the numeric parameters fall within the bounds $\mathcal{B} = [\underline{\mathcal{B}}$,$\overline{\mathcal{B}}$], we limit numeric parameters by Equation (\ref{bounds}).
\begin{equation}
\Theta = max(min(\Theta,\overline{\mathcal{B}}),\underline{\mathcal{B}}).\label{bounds}
\end{equation}

To guarantee the solution plan $\plan$ is executable and can achieve the goal by avoiding all obstacles, we define $\mathcal{L}_{stop}$ in Equation (\ref{equation:stop}). When $\mathcal{L}_{stop}=0$, a solution plan is found.
\begin{align}
\setlength{\abovedisplayskip}{1pt}
\setlength{\belowdisplayskip}{-1pt}
    \mathcal{L}_{stop}=~&
    \begin{cases}
      \sum_{i=0}^{\mu-1}\mathcal{L}_{o_i}, & \mbox{if}
      ~g \subseteq s_{\mu}\mbox{ and}\\& \plan\mbox{ is executable} \\
      \infty, & \mbox{otherwise}
    \end{cases}
    \label{equation:stop}
\end{align}
}
{\ours} has a soundness property as follows:
\begin{theorem}
The action sequence computed by {\ours} is a valid plan for the planning problem.
\end{theorem}
\Revised{
\emph{Proof:} According to Algorithm \ref{algorithm:code}, {\ours} outputs a plan $\plan$ when the loss $\mathcal{L}_{stop}=0$. As shown in Equation (\ref{equation:stop}), we get $\mathcal{L}_{stop}=0$ if and only if plan $\plan$ is executable, the accumulated sum of $\mathcal{L}_{o}$ is 0, and $g \subseteq s_{\mu}$. An executable plan $\plan$ indicates all preconditions of actions in $\plan$ are satisfied and the states are updated correctly. We get $\sum_{i=0}^{\mu-1}\mathcal{L}_{o_i} = 0$ if and only if all goals are achieved after executing the $\mu$th action in $\plan$ and sum of $\mathcal{L}_{o}$ of each step is 0. We get $g \subseteq s_{\mu}$ when the targeted propositions are involved in $s_{\mu}$ and the numeric variables in $s_{\mu}$ fall within the required intervals. In other words, all obstacles are avoided when reaching the goal, and the plan is executable, i.e., preconditions of actions are satisfied at the states where they are executed. Thus, the output action sequence is a solution plan for the problem.
}


\section{Experiments}
In this section, we evaluated our {\ours} approach in three domains: AUV\footnote{It is from ScottyActivity \cite{DBLP:journals/jair/Fernandez-Gonzalez18}}, Taxi\footnote{http://agents.fel.cvut.cz/codmap/} and Rover\footnote{https://ipc02.icaps-conference.org/}, which were also used in \cite{DBLP:journals/jair/Fernandez-Gonzalez18,DBLP:conf/ecai/SavasFLM16,Hoeller2020HDDL}.
\Revised{Since} there is no planner \Revised{that} can directly handle our \emph{mixed} planning problems,
we compare {\ours} with three state-of-the-art planners with modifications, Metric-FF \cite{DBLP:journals/jair/Hoffmann03}, POPCORN \cite{DBLP:conf/ecai/SavasFLM16}, and ScottyActivity \cite{DBLP:journals/jair/Fernandez-Gonzalez18} with different settings, as shown below:

\begin{enumerate}
    \item For convex domains, we compare {\ours} against Metric-FF, POPCORN, and ScottyActivity. We ran ScottyActivity with the Enforced Hill-Climbing \cite{DBLP:journals/jair/HoffmannN01}. Since the original domains include non-linear numeric effects which POPCORN is unable to handle, we adapted POPCORN by the following two steps:
\begin{itemize}
    \item 
    We assigned the duration of actions with a fixed value, which makes each movement of x-axis or y-axis be a product of a fixed duration and another numeric parameter, respectively, i.e., x-velocity and y-velocity. Such a modification makes the effect become linear, which can be handled by POPCORN. 
    \item Based on the modified domains, we configured POPCORN with Enforced Hill-Climbing and best-first searching strategy and used CPLEX \cite{cplex2009v12} to optimize the plans, aiming at minimizing our evaluation criterion by updating numeric parameters.
\end{itemize}
We denote the adapted POPCORN by POPCORN$^+$. 

\item For non-convex domains, we also compare {\ours} against Metric-FF with discretized numeric effects. 
We do not compare with POPCORN and ScottyActivity in non-convex domains. 
\Revised{It is nontrivial to adapt these two planners to handle non-convex continuous numeric space, as they are optimization algorithms focusing on convex continuous numeric space.}
To make Metric-FF deal with our problem, we adapted Metric-FF following the idea of ScottyPath \cite{ScottyPath} about constructing safe regions from non-convex space.
\MajorRevision{ScottyPath divides non-convex space into several convex safe regions and computes a sequence of those regions via heuristic search, such that convex optimization can be used to generate the shortest sub-path between convex regions. Since ScottyPath focuses on geometric path planning without logical relations and extending ScottyPath to handle logical relations is nontrivial, it is not appropriate to compare {\ours} to it.}Specifically, we adapted Metric-FF by the following steps:
\begin{itemize}
    \item As in this paper we only evaluated approaches on path planning domains \MajorRevision{with logical relations and numeric changes}, we discretized numeric actions by stipulating each movement to have a fixed step length.
    \item \MajorRevision{We introduce three action models (as shown in Figure \ref{fig:metric-ff-domain}) to model movements towards eight angles, i.e., two horizontal movements, two vertical movements, and four diagonal movements. Each action updates the position according to the fixed step length.
    }

\begin{figure}[!ht]
    \centering
    \includegraphics[width=0.88\textwidth]{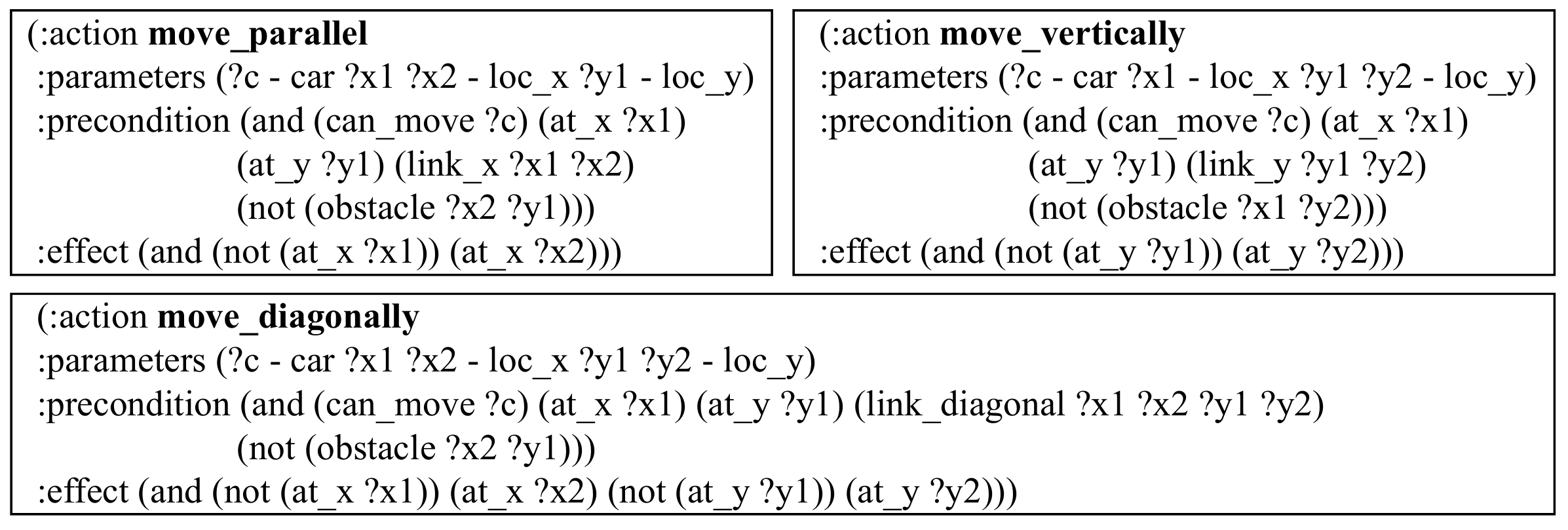}
    \caption{\MajorRevision{Three action models for modeling movements in the AUV domain for Metric-FF.}}
    \label{fig:metric-ff-domain}
\end{figure}

    \item As each movement is linear, we can estimate whether a movement crashes an obstacle by checking whether the line between the current position to the stop position crosses the regions of the obstacles.
    \ignore{Then we enumerate all lines with a fixed length that do not crash with any obstacle and write them into a PDDL file. }\MajorRevision{
    We enumerated all movements that do not crash with any obstacles, and formalized those movements with propositions in the form of ``(link\_x ?x1 ?x2)'',  ``(link\_y ?y1 ?y2)'', and ``(link\_diagonal ?x1 ?y1 ?x2 ?y2)''. Specifically, if a movement is horizontal, i.e., from (?x1, ?y1) to (?x2, ?y1), we denote it by a proposition ``(link\_x ?x1 ?x2)''. If a movement is vertical, i.e., from (?x1, ?y1) to (?x1, ?y2), we denote it by ``(link\_y ?y1 ?y2)''. If a movement is from (?x1, ?y1) to (?x2, ?y2), we denote it by ``(link\_diagonal ?x1 ?y1 ?x2 ?y2)''. After mapping all movements to propositions, we write those propositions into the initial state in problem PDDL file.
    }
 \end{itemize}
\Revised{We denote the adapted Metric-FF by Metric-FF$^+$. We ran Metric-FF \cite{DBLP:journals/jair/Hoffmann03} with six canonical configurations (i.e., Standard-FF, best-first search, best-first search with helpful actions pruning, Weighted A*, A* epsilon, Enforced Hill-Climbing with helpful actions pruning and A* episilon) and chose the best solution as the final plan.

\item  Finally, we evaluated how the hyperparameters influenced the performance and iterations of {\ours}. Note that
we did not compare {\ours} to \citet{DBLP:conf/nips/WuSS17}, 
as their approach cannot handle mixed planning problems with propositional effects and preconditions.}
\end{enumerate}

\subsection{Benchmarks}
For each domain, we randomly generated 15 planning instances with a $150 \times 150$ sized map. For each instance, we randomly generated one to five objective regions which do not overlap each other. Those objective regions are squares ranging from $5 \times 5$ to $20 \times 20$. To generate non-convex domains, we randomly placed obstacles in the map to make the continuous search space be non-convex. The obstacles were also randomly generated in the form of rectangle regions ranging from  $5 \times 5$ to $25 \times 25$. Note that obstacles do not overlap each other or objective regions.

\subsubsection{The AUV Domain}

\begin{figure*}[!ht]
 \setlength{\abovedisplayskip}{1pt}
 \setlength{\belowdisplayskip}{1pt}
  \centering
  \includegraphics[width=0.98\textwidth]{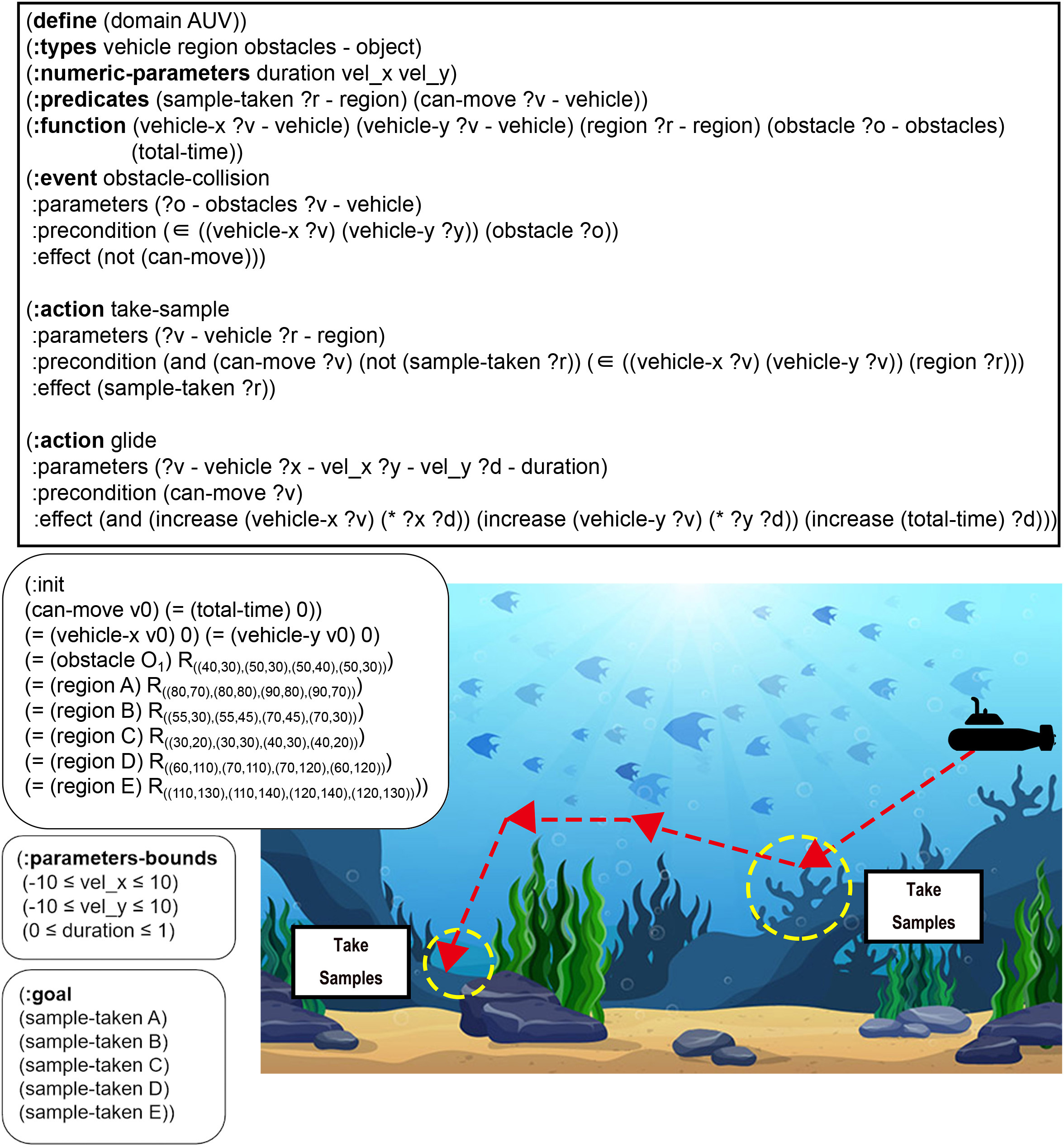}
  \caption{\Revised{The AUV domain and an example planning instance.}}
  \label{figure:plan_auv}
\end{figure*}

In the AUV domain, an AUV (automated underwater vehicle) aims to reach each objective region and take samples without touching obstacles. We modified the AUV domain used in ScottyActivity \cite{DBLP:journals/jair/Fernandez-Gonzalez18} by defining the effects of the action ``glide'' with the product of the velocity in the x-axis or y-axis (``vel\_x'' or ``vel\_y'') and attaching the execution time of ``glide'' with ``duration''. 
\Revised{
We also added an event to describe scenarios that the AUV collides with obstacles.}
Note that we denote a region in the form of $R_L$, where $L$ is a list of vertices indicating the region surrounded by the edges connecting vertices sequentially in the list.
The whole description of the AUV domain is shown in Figure \ref{figure:plan_auv}, which includes the domain description and a planning instance. In this instance, an AUV ``v0'' aims to take samples in five regions avoiding obstacle ``O$_1$'', 
which is represented by ``(Obstacle O$_1$)'', indicating an obstacle region ``O$_1$'' surrounded by edges with vertices  (40,30), (50,30), (50,40), and (50,40).

\subsubsection{The Taxi Domain}

\begin{figure}[!ht]
 \setlength{\abovedisplayskip}{1pt}
 \setlength{\belowdisplayskip}{1pt}
  \centering
  \includegraphics[width=0.98\textwidth]{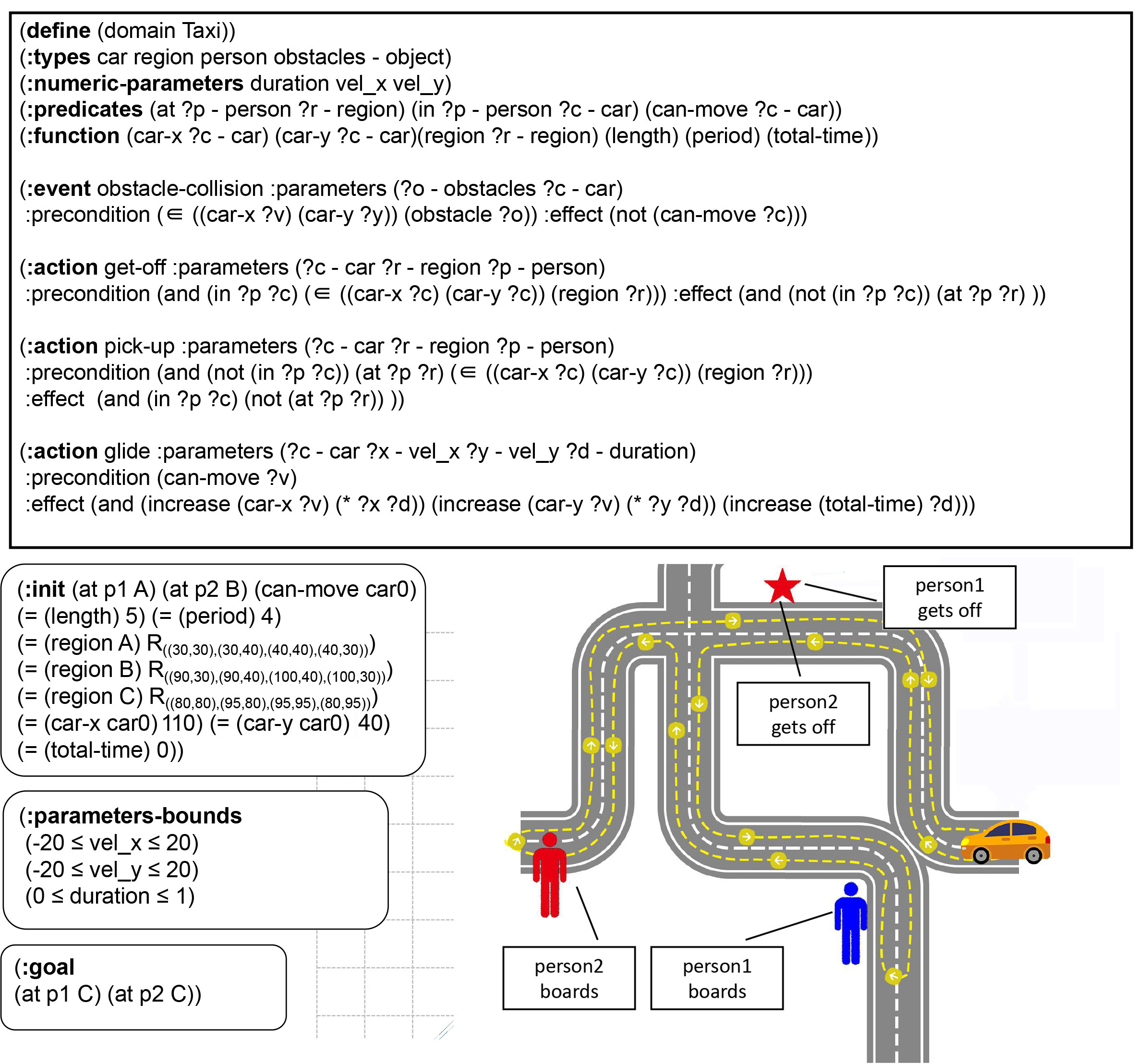}
  \caption{\Revised{The Taxi domain and an example planning instance.}}
  \label{figure:plan_taxi}
\end{figure}

In the Taxi domain, an agent drives a taxi to pick up passengers to destinations. The domain description of Taxi is shown in Figure \ref{figure:plan_taxi}, which includes three action models, ``pick-up'', ``get-off'', and ``glide'' with three numeric parameters ``vel\_x'', ``vel\_y'' and ``duration''. A passenger can be picked up by a taxi when he is in the region the same as the taxi. \Revised{Note that the taxi cannot cross obstacles when moving.}
In the instance (including an initial state, parameters bounds and the goal) shown in Figure \ref{figure:plan_taxi}, two passengers ``person1 (or p1)'' and ``person2 (or p2)'' are in regions A and B, respectively, aiming to get to the same destination ``C''.

\subsubsection{The Rover domain}
\begin{figure*}[!ht]
 \setlength{\abovedisplayskip}{1pt}
 \setlength{\belowdisplayskip}{1pt}
  \centering
  \includegraphics[width=0.98\textwidth]{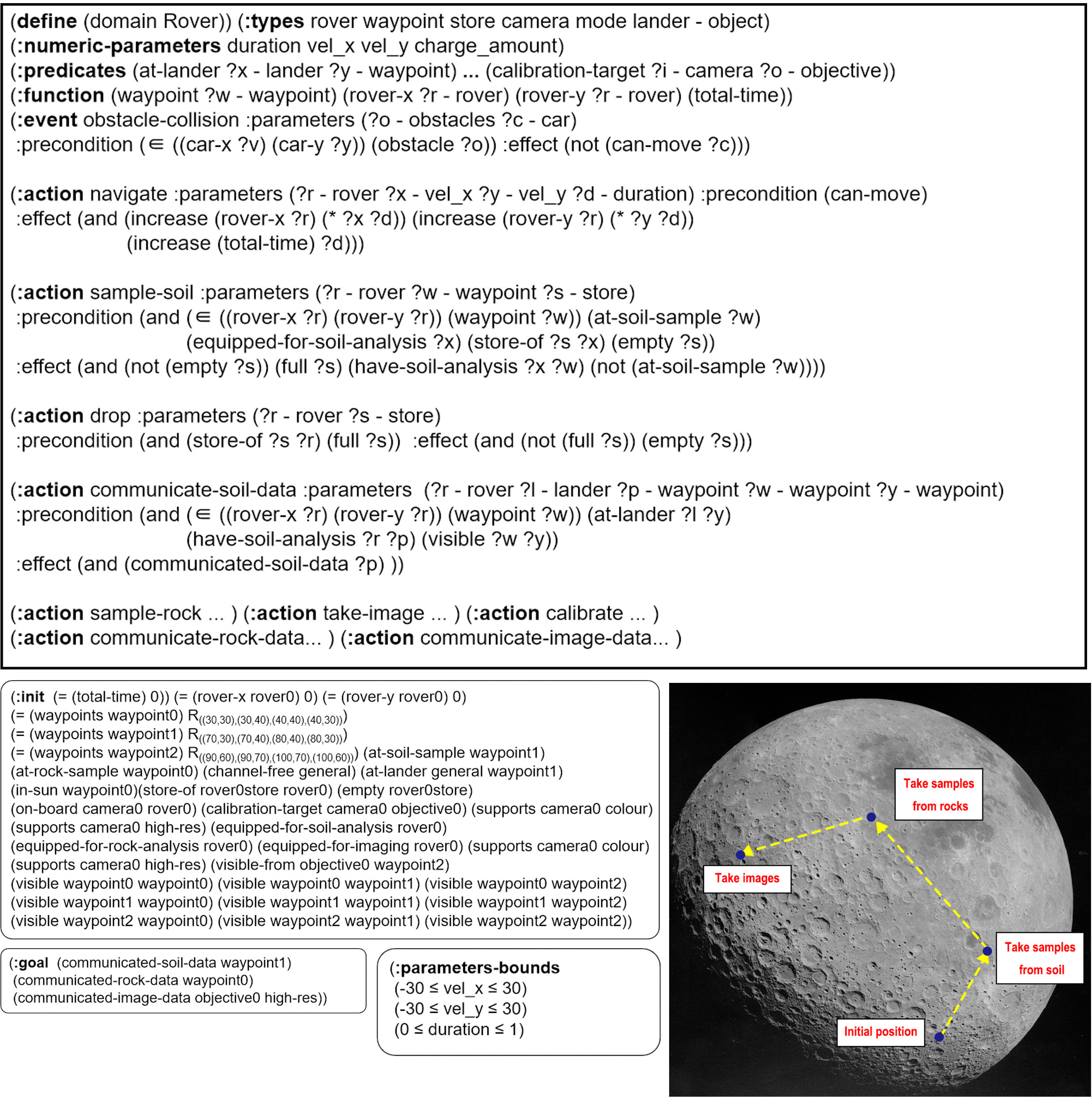}
  \caption{\Revised{The Rover domain and an example planning instance.}}
  \label{figure:plan_rover}
\end{figure*}

In the Rover domain, a rover navigates on a planet surface, aiming at taking samples of soil and rock, taking images and communicating them back to the lander. The domain description of Rover is shown in Figure \ref{figure:plan_rover}, which includes a numeric action model ``navigate'' with numeric parameters (``vel\_x'', ``vel\_y'' and ``duration'') and eight logical action models such as ``sample-soil'' and ``communicate-rock-data''.

Figure \ref{figure:plan_rover} also shows a planning instance including an initial state, parameters-bounds and goals. In this instance, a rover ``rover0'' aims to take samples at ``waypoint0'' and ``waypoint1''  and an image at ``waypoint2'', starting from the ``initial position'' shown in the figure.

\Revised{
\subsubsection{Comparison of three domains}
All three domains are \MajorRevision{different} from each other with respect to the following features: (1) the number of object types, (2) the number of predicates, and (3) the number of action models, as shown in Table \ref{table:domains}.

\begin{table*}[!ht]
\centering
\scriptsize
\begin{tabular}{ccccc}
  \toprule
Domain & Types of objects & Predicates & Action models
  \\
  \hline
AUV & 2 & 2 & 2 \\
Taxi & 3 & 2 & 3 \\
Rover& 7 & 22 & 9 \\
  \bottomrule
 \end{tabular}
 \caption{\Revised{Features of the three domains.}}
\label{table:domains}
 \end{table*}

From Table \ref{table:domains}, we can see that AUV is a domain containing two action models and two types of objects. One of the action models contains numeric preconditions, which requires the corresponding numeric variables within a range of real values. The other one includes non-linear effects, guiding an agent to move from one position to another. Compared with the AUV domain, the Taxi domain may have solutions with round paths, due to the randomness of passengers and destinations. A round path means that a taxi first reaches region ``B'' from region ``A'' to pick up a passenger, then goes back to region ``A'' to drop him off.
Round paths significantly increase the difficulties of solving problems. 
For the Rover domain, it is more complicated than the other two domains, as it includes more object types, predicates and action models. As the samples are required to \MajorRevision{be} returned back to the lander, the Rover domain may have solutions with round paths as well.
Although the three domains in some sense are path planning domains, they have their own properties and difficulties. \MajorRevision{Compared to path planning problems solved by path planners, our planning problems include both logical relations and numeric changes, where the logical relations make the planning problems even more challenging, since we need to consider the constraints posed by logical relations, e.g., the order of objective regions, when considering the numeric changes.}
}

\subsubsection{Evaluation Criterion}
In this paper, we consider the navigation distance as the optimization objective for the three domains. We only define the cost with respect to action models that make changes to locations, such as ``glide'', which changes the location of the agent by effects: ``increase x-location (* $?v_x$ $?d$)'' and ``increase y-location (* $?v_y$ $?d$)''. Given a plan $\plan = \langle a_0, a_1, \dots, a_{N-1} \rangle$, we define the evaluation criterion $C(\plan)$ by \MajorRevision{summing} the cost of each action in $\plan$:
\begin{align}
    \notag
    C(\sigma) &= \sum_{a_i \in \plan}\psi(a_i) ~= \sum_{a_i \in \plan}\sqrt{(?v_x ?d)^2 + (?v_y ?d)^2}
\end{align}
Note that our approach also works with other evaluation criteria, such as makespan or plan length.

\subsection{Experimental Results}
\Revised{We use ``$\mxplanner$'', ``$\metricff$'', ``$\ScottyActivity$'', and ``$\popcorn$'' to denote {\ours}, Metric-FF$^+$, ScottyActivity and POPCORN$^+$, respectively.} We use subscripts to denote bounds of numeric parameters. For example, ``$\mxplanner_{10}$'' and ``$\ScottyActivity_{10}$'' indicate all numeric parameters in {\ours} and ScottyActivity are restricted within [-10,10], respectively.
In general, we denote the subscript by $\bound$, i.e., $\mxplanner_{\bound}$, indicating all numeric parameters are restricted within $[-\bound,+\bound]$. Different from {\ours} and ScottyActivity, Metric-FF$^+$ and POPCORN$^+$ can only handle discrete parameters. As we mentioned in the procedures of adapting Metric-FF and POPCORN at the beginning of Section 5, we assign the numeric parameters with values in advance in order to make action effects discrete.
We use $\metricff_{\bound}$ to denote the value of numeric parameters in Metric-FF. For example, ``$\metricff_{10}$'' indicates that all numeric parameters of Metric-FF$^+$ are 10 (e.g., ``vel\_x'' = 10).
\MajorRevision{Specifically,} we set parameter ``duration'' of {\ours} and ScottyActivity to be $0 \leq \text{duration} \leq 1$, and set ``duration'' of Metric-FF$^+$ and POPCORN$^+$ to be $\text{duration} = 1$. 
We set the cutoff time as 36000 seconds.

We evaluated our {\ours} with respect to the following aspects:
\Revised{
\begin{itemize}
    \item \textbf{Costs with parameter bounds in convex problems:}
We evaluated the performance of {\ours} in problems without obstacles, with comparison to Metric-FF$^+$, ScottyActivity and POPCORN$^+$.

\item \textbf{Costs with parameter bounds in non-convex problems:}
We evaluated Metric-FF$^+$ and {\ours} by varying $\bound$ in instances with obstacles.

\item \textbf{Costs without parameter bounds in non-convex problems:}
Note that it is not necessary for our {\ours} to indicate numeric parameter bounds. We use $\mathcal{R}$ as a subscript to denote the case that parameters are unconstrained.
We evaluated {\ours} on problems without restriction of parameter bounds.

\item \textbf{Costs with the number of obstacles increasing:}
To see the change of costs with respect to the number of obstacles, we evaluated the performance of {\ours} and Metric-FF$^+$ on problems with different number of obstacles.

\item \textbf{Costs with different hyperparameters in {\ours}:}
To see the impact of hyperparameters in {\ours}, we evaluated the performance of {\ours} by varying the values of the three hyperparameters, i.e., $w_1$, $w_2$, and $w_3$, in the loss function $\mathcal{L}_{all}$ (Equation (\ref{equation:loss_instantaneous})).

\item \textbf{Iterations with different hyperparameters in {\ours}:}
We also evaluated the number of iterations of {\ours} by varying the values of the three hyperparameters, i.e., $w_1$, $w_2$, and $w_3$, in the loss function (Equation (\ref{equation:loss_instantaneous})).

\MajorRevision{ \item \textbf{Running time:}
Finally, we show the running time of {\ours}, Metric-FF, ScottyActivity, and POPCORN$^+$ with respect to different parameter bounds.
}
\end{itemize}
We will present the experimental results in detail with respect to the above-mentioned aspects in the following subsections.
}

\subsubsection{Costs with parameter bounds in convex problems}

\Revised{
We first compare {\ours} to Metric-FF$^+$, POPCORN$^+$, and ScottyActivity, with respect to different parameter bounds in convex problems. We count the number of instances successfully solved by the four planners with different parameter bounds. The results are shown in Table \ref{table:count_no_obstacle}. From Table \ref{table:count_no_obstacle} we can see that both {\ours} and ScottyActivity with different parameter bounds ($\bound=10$ and $\bound=20$) can solve all instances, while $\popcorn_{10}$, $\metricff_{20}$, and $\popcorn_{20}$ fail to solve all instances of all three domains. 
This indicates that flexible searching, as done by both {\ours} and ScottyActivity, can indeed help improving the success rate of generating solutions to planning problems. This fact is also verified by the result of $\metricff_{10}$. When the step length, i.e., $\bound$, is set to be 10, Metric-FF$^+$ can solve all planning problems successfully. This is because when the step length is small enough (compared to $\bound=20$), Metric-FF$^+$ is flexible enough to successfully compute solutions to all planning problems.
In some domains, POPCORN$^+$ fails to generate plans within the cut-off time. This is because POPCORN$^+$ is a forward-searching heuristics planner, which builds up all linear constraints in terms of numeric parameters. When the bounds of numeric parameters is small, it is difficult for POPCORN$^+$ to solve problems since the smaller the bound is, the larger the graph POPCORN$^+$ needs to build is. Hence, in domains AUV and Taxi, $\popcorn_{20}$ successfully solves more instances than $\popcorn_{10}$. On the other hand, bigger bounds do not always mean to be better, since the bigger the bounds are, the more possible the values of numeric parameters are. Enlarging the possible value scope of numeric parameters will increase the difficulty of solving the problems.
\begin{table}[h]
\centering
 \caption{The number of planning problems successfully solved by different planners.} \label{table:count_no_obstacle}
\begin{tabular}{c|cccc|cccc}
  \toprule
  & $\mxplanner_{10}$ &  $\metricff_{10}$ & \Revised{$\popcorn_{10}$} &  $\ScottyActivity_{10}$ & $\mxplanner_{20}$ &  $\metricff_{20}$ & \Revised{$\popcorn_{20}$} &  $\ScottyActivity_{20}$ \\
  \hline
  AUV & 15 & 15 & \Revised{14} & 15 & 15 & 14 & \Revised{15} & 15 \\
  Taxi & 15 & 15 & \Revised{7} & 15 & 15 & 13 & \Revised{11} & 15 \\
  Rover & 15 & 15 & \Revised{15} & 15 & 15 & 12 & \Revised{15} & \Revised{15} \\
  \bottomrule
 \end{tabular}
 \end{table}
}

\begin{figure}[!ht]
\setlength{\abovedisplayskip}{0pt}
\setlength{\belowdisplayskip}{0pt}
\centering
\subfigcapskip=-5pt
\subfigcapmargin = .05cm
\subfigure[AUV]{
  \begin{minipage}[b]{0.31\textwidth}
    \includegraphics[width=\textwidth]{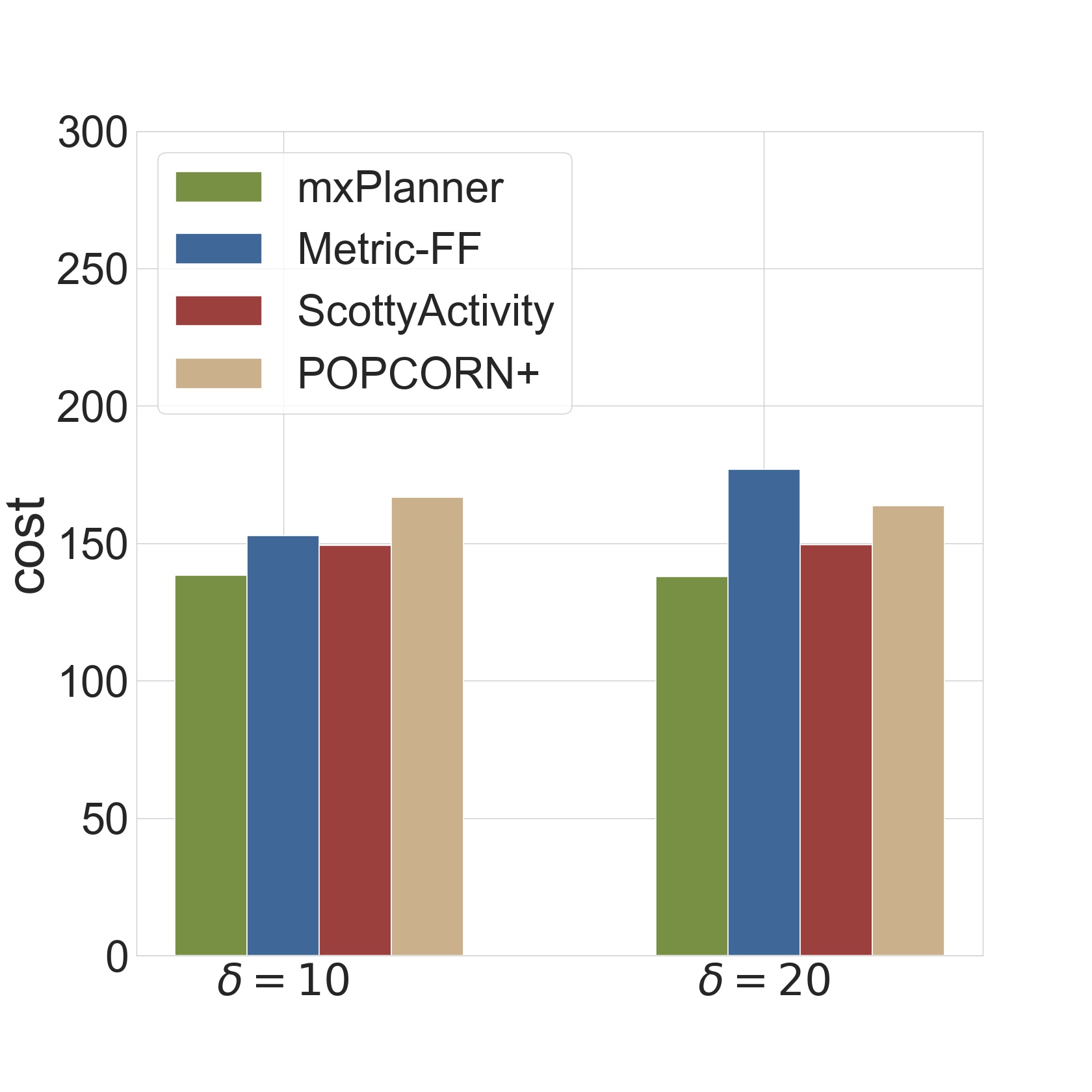}
  \end{minipage}}
\subfigure[Taxi]{
    \begin{minipage}[b]{0.31\textwidth}
\centering
    \includegraphics[width=\textwidth]{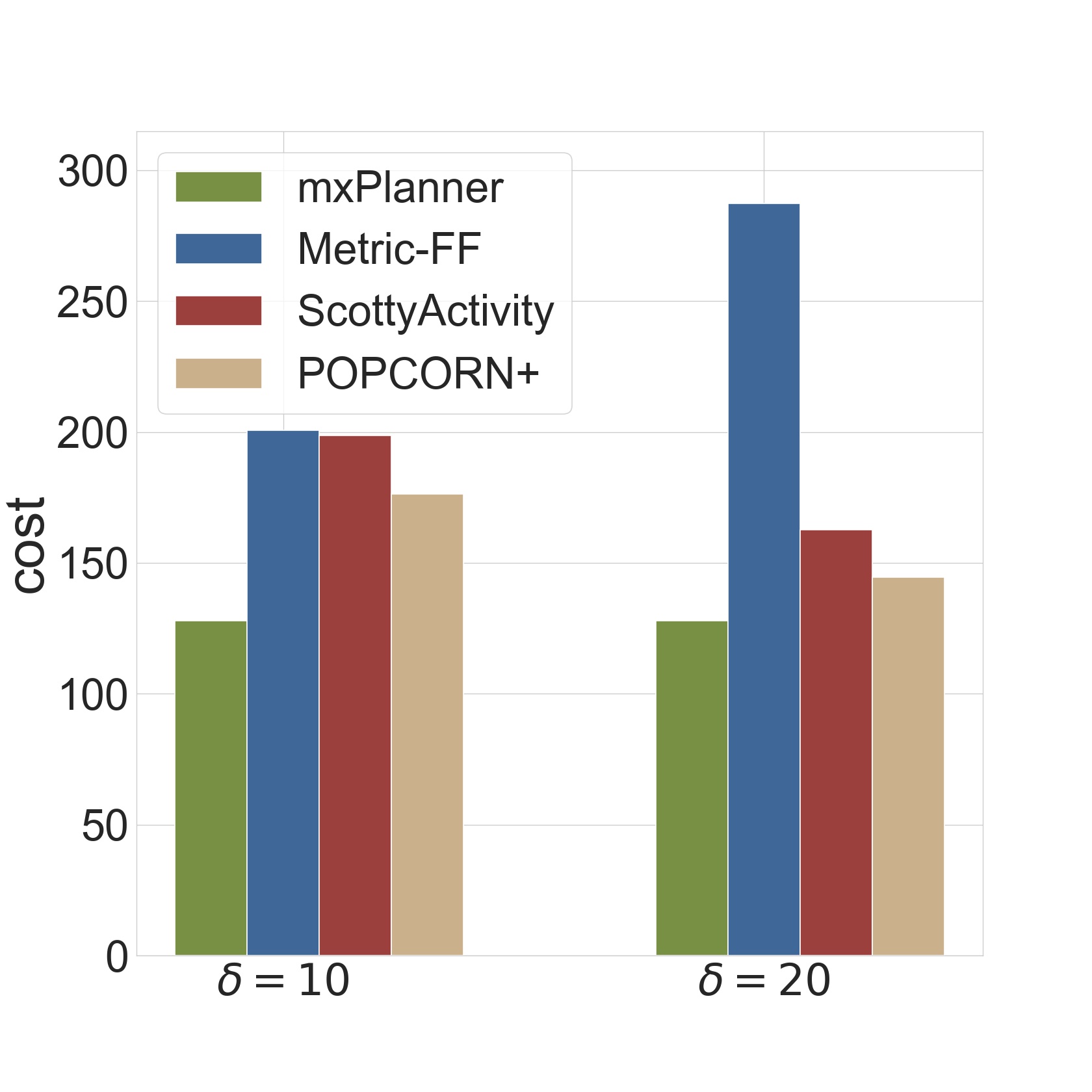}
  \end{minipage}}
\subfigure[Rover]{
    \begin{minipage}[b]{0.31\textwidth}
\centering
    \includegraphics[width=\textwidth]{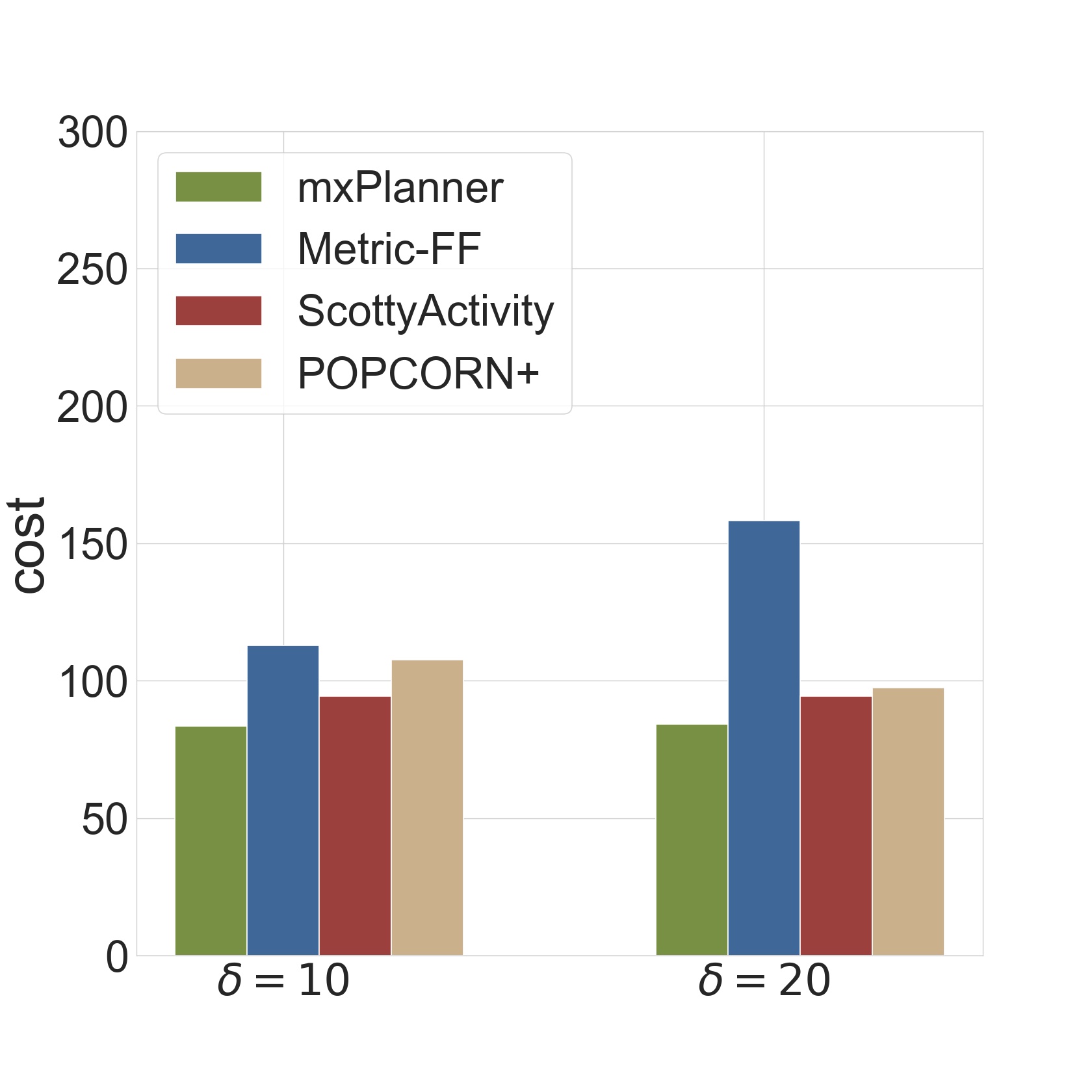}
  \end{minipage}}
\caption{\Revised{Average cost of obstacle-free instances in the three domains with different parameter bounds.}} 
\label{figure:problems_without_obstacles}
\end{figure}

\Revised{
To evaluate the costs of {\ours}, Metric-FF$^+$, POPCORN$^+$, and ScottyActivity with respect to different parameter bounds, we set the parameter bounds to be 10 and 20, respectively. The results are shown in Figure \ref{figure:problems_without_obstacles}.
\ignore{We only consider those instances that are solved by four planners with both $\delta=10$ and $\delta=20$. More specifically, in the AUV domain, the Taxi domain and the Rover domain, we only consider 13, 6, 12 instances respectively. }\MajorRevision{Particularly, as shown in Table \ref{table:count_no_obstacle}, some of the 15 instances in each domain cannot be solved by all four planners with both parameter bounds (i.e., 10 and 20). When calculating the average cost of each planner in each domain, we only consider those instances that were solved by all four planners with both $\delta=10$ and $\delta=20$. More specifically, we only consider 13, 6, 12 instances that were successfully solved by all four planners in domains AUV, Taxi, and Rover, respectively.}
The results show that {\ours} has the top performance. 
Especially, {\ours} performs similarly with ScottyActivity in AUV and Rover, whose instances are simpler than those of Taxi. 
The usage of convex optimization makes ScottyActivity be capable of computing plans of high-quality for some simple instances. However, the quality of plans significantly depends on the heuristic searching procedure on the plan skeleton before invoking a convex optimization solver. 
This also explains why ScottyActivity performs much worse than {\ours} in Taxi.
Additionally, in Taxi, the average \MajorRevision{plan costs computed by $\ScottyActivity_{10}$ and $\popcorn_{10}$ are significantly higher than those plans with respect to $\delta=20$. The reason is that some instances in Taxi contain round trips, which means that the car needs to go forth and back. When problems contain multiple persons waiting \ignore{for being moved}\MajorRevision{to be moved,}} $\ScottyActivity_{10}$ and $\popcorn_{10}$ compute plans with repeated loops, resulting in a larger cost. The average plan costs of Metric-FF$^+$ are the largest in most cases of three domains. Because the discretization of Metric-FF$^+$ lets agents move more compared with the planners using an optimization algorithm. In addition, $\metricff_{10}$ performs better than $\metricff_{20}$ in all domains, because a large step may lead to a detour from the path.
}

\subsubsection{Costs with bound of parameters in problems with obstacles}

\Revised{We also evaluate {\ours} on more complicated planning problems with obstacles, comparing with Metric-FF$^+$. 
The results are shown in Table \ref{table:obstacles}. 
From the table, we observe that {\ours} solves all instances successfully, while Metric-FF$^+$ fails on several instances within the cutoff time. }For the instances solved by the two planners, {\ours} also outperforms Metric-FF$^+$. 
\MajorRevision{
This is because that numeric effects used in Metric-FF$^+$ are discretized, resulting in all movements with the same lengths. The fixed movements make Metric-FF$^+$ be impossible to dynamically adapt step lengths when planning. In contrast,} {\ours} computes numeric parameters for each step, which avoids unnecessary paths due to fixing the step length. The results indicate that {\ours} is able to deal with complex changes and compute plans with lower costs without discretization.

\begin{table*}[ht]
\centering
\scriptsize
\begin{tabular}{p{1mm}|p{5.2mm}p{5.6mm}|p{5.2mm}p{5.6mm}|p{5.2mm}p{5.6mm}|p{5.2mm}p{5.6mm}|p{4mm}p{5.2mm}|p{4mm}p{4mm}}
  \toprule
  & \multicolumn{4}{c|}{AUV domain}
  & \multicolumn{4}{c|}{Taxi domain}
  & \multicolumn{4}{c}{Rover domain}
  \\
  \hline
 & $\mxplanner_{10}$ & $\metricff_{10}$
 & $\mxplanner_{20}$ & $\metricff_{20}$
 & $\mxplanner_{10}$ & $\metricff_{10}$
 & $\mxplanner_{20}$ & $\metricff_{20}$
 & $\mxplanner_{10}$ & $\metricff_{10}$
 & $\mxplanner_{20}$ & $\metricff_{20}$ \\
 \hline
 1 & \textbf{106.89} & 126.57 & \textbf{107.72} & 124.85 & \textbf{144.24} & 164.85 & \textbf{143.24} & 153.14 & \textbf{35.53} & 54.14 & \textbf{35.53} & $\backslash$\\
 2 & \textbf{127.76} & 158.28 & \textbf{127.62} & 156.57 & \textbf{113.79} & 154.85 & \textbf{113.20} & 156.57 & \textbf{117.45} & 134.85 & \textbf{117.45} & 136.57\\
 3 & \textbf{117.83} & 144.14 & \textbf{116.79} & 148.28 & \textbf{117.04} & 221.42 & \textbf{116.95} & $\backslash$ & \textbf{42.75} & 78.28 & \textbf{42.82} & 68.28\\
 4 & \textbf{188.69} & 381.13 & \textbf{188.05} & 442.84 & \textbf{165.24} & 275.56 & \textbf{154.20} & 261.42 & \textbf{82.40} & 130.71 & \textbf{83.39} & 136.57\\
 5 & \textbf{135.78} & 136.57  & \textbf{130.00} & 136.57& \textbf{134.01} & 213.14 & \textbf{135.13} & $\backslash$ & \textbf{53.63} & $\backslash$ & \textbf{53.72} & $\backslash$\\
 6 & \textbf{97.37} & 106.57 & \textbf{98.15} & 116.57& \textbf{132.12} & 241.42 & \textbf{133.49} & $\backslash$ & \textbf{79.57} & 120.71 & \textbf{79.34} & 108.28 \\
 7 & \textbf{183.32} & 194.85
    & \textbf{178.09} & 216.57 & \textbf{169.55} & 287.99 & \textbf{177.57} & 329.71 & \textbf{66.28} & 108.28 & \textbf{91.72} & $\backslash$\\
 8 & \textbf{193.23} & 208.99 & \textbf{181.84} & 216.57 & \textbf{130.56} & 266.27  & \textbf{129.79} & 221.42 & \textbf{78.38} & 82.43 & \textbf{130.84} & $\backslash$\\
 9 & \textbf{152.91} & 245.56 & \textbf{174.00} & $\backslash$ & \textbf{171.72} & 293.85 & \textbf{165.14} & $\backslash$ & \textbf{106.26} & 148.99 & \textbf{117.14} & $\backslash$\\
 10 & \textbf{153.58} & 397.99 & \textbf{152.27} & 386.27 & \textbf{103.91} & 414.56 & \textbf{105.69} & $\backslash$ & \textbf{131.09} & 251.42 &\textbf{118.95} & $\backslash$\\
 11 & \textbf{220.28} & 567.70  & \textbf{303.90} & 477.99 & \textbf{93.52} & 98.28 & \textbf{93.52} & 88.28 & \textbf{141.13} & 217.28 & \textbf{141.50} & $\backslash$\\
 12 & \textbf{264.90} & 766.69 & \textbf{281.73} & 642.84 & \textbf{148.52} & 363.85 & \textbf{137.96} & $\backslash$ & \textbf{59.78} & 104.14 & \textbf{67.92} & 100\\
 13 & \textbf{85.11} & 92.43  & \textbf{84.61} & 108.28 & \textbf{111.69} & 122.43 & \textbf{106.19} & 140 & \textbf{79.78} & 102.43 & \textbf{109.26} & 148.28 \\
 14 & \textbf{102.92} & 114.14  & \textbf{102.97} & 148.28 & \textbf{181.42} & 378.70 & \textbf{183.42} & 321.42 & \textbf{84.74} & 174.85 & \textbf{85.28} & $\backslash$\\
 15 & \textbf{148.90} & 180.71  & \textbf{148.73} & 228.28 & \textbf{95.83} & 156.57 & \textbf{90.65} & 140 & \textbf{104.22} & 134.85 & \textbf{106.08} & 184.85\\
$A$ & \textbf{151.96} & 254.82 & \textbf{157.32} & 253.63& \textbf{134.21} & 243.58 & \textbf{132.42} & 201.33 & \textbf{86.38} & 122.04 & \textbf{83.20} & 99.71\\
  \bottomrule
 \end{tabular}
 \caption{Costs of instances with obstacles in the AUV domain, the Taxi domain, and the Rover domain with different parameter bounds. Each domain contains 15 instances. Each problem contains less than five obstacles. $A$ in the table indicates the average cost of the solved problems.}
\label{table:obstacles}
 \end{table*}

As shown in Table \ref{table:obstacles}, compared with Metric-FF$^+$, the performance of {\ours} is less influenced by the parameter bound. 
\MajorRevision{In contrast,} the larger the parameters are, the fewer problems are solved by Metric-FF$^+$. 
This is because when the parameters are larger, the step length becomes longer, making it more possible to cross an objective region instead of entering it. \Revised{In other words, for discretized planners, it is more likely to enter an objective region with a smaller step length.} On the other hand, it becomes increasingly difficult to solve such a searching problem when parameter bounds get smaller, since the searching space would become much larger. Compared with discretized planner, {\ours} is not limited to fixed parameters, neither heuristic searching nor updating parameters.

\RevisedByHankz{From Table \ref{table:obstacles} we can also see that the average of plan costs in Rover domain is less than that in \ignore{domains AUV and Taxi}\MajorRevision{the AUV and Taxi domains.}
This is because all three domains require arriving at some regions and executing actions to accomplish goals. The regions in the Rover domain are more concentrated than those in \ignore{domains AUV and Taxi}\MajorRevision{the AUV and Taxi domains,} resulting in the average of plan costs in the Rover domain is a little lower than both AUV and Taxi.}

\subsubsection{Costs without parameter bounds in non-convex problems}

\begin{table*}[!ht]
\centering
\scriptsize
\begin{tabular}{p{1mm}|p{4.2mm}p{4.2mm}p{4.2mm}c|p{4.2mm}p{4mm}p{4.2mm}c|p{4.2mm}p{4.2mm}p{4.2mm}c}
  \toprule
  & \multicolumn{4}{c|}{AUV domain}
  & \multicolumn{4}{c|}{Taxi domain}
  & \multicolumn{4}{c}{Rover domain}
  \\
  \hline
 & $\metricff_{2}$ & $\metricff_{10}$ & $\metricff_{20}$ & $\mxplanner_{\mathcal{R}}$
 & $\metricff_{2}$ & $\metricff_{10}$ & $\metricff_{20}$ & $\mxplanner_{\mathcal{R}}$
 & $\metricff_{2}$ & $\metricff_{10}$ & $\metricff_{20}$ & $\mxplanner_{\mathcal{R}}$
 \\
 \hline
 1 & 122.85 & 126.57 & 124.85 & \textbf{107.15} & 165.42 & 164.85 & 153.14 & \textbf{143.51} & 40.28 & 54.14 & $\backslash$ & \textbf{35.68}\\
 2 & 142.31 & 158.28 & 156.57 & \textbf{128.78} & 295.97 & 154.85 & 156.57 & \textbf{111.19} & 125.62 &  134.85 & 136.57 & \textbf{112.36}\\
 3 & 130.08 & 144.14 & 148.28 & \textbf{118.70} & 169.82 & 221.42 & $\backslash$ & \textbf{122.89} & 47.60 & 78.28 & 68.28 & \textbf{44.11}\\
 4 & 206.68 & 381.13 & 442.84 & \textbf{189.91} & 333.30 & 275.56 & 261.42 & \textbf{147.47} & 140.37 & 130.71 & 136.57 & \textbf{91.19}\\
 5 & 143.60 & 136.57 & 136.57 & \textbf{127.58} & 385.42 & 213.14 & $\backslash$ & \textbf{139.28} & 84.08 & $\backslash$ & $\backslash$ & \textbf{53.36}\\
 6 & \textbf{103.05} & 106.57 & 116.57 & 105.29 & 538.76 & 241.42 & $\backslash$ & \textbf{132.26} & 156.65 & 120.71 & 108.28 & \textbf{79.82}\\
 7 & 194.85 & 194.85 & 216.57 & \textbf{175.81} & 518.22 & 287.99 & 329.71 & \textbf{178.40} & 95.40 & 108.28 & $\backslash$ & \textbf{70.92}\\
 8 & 266.05 & 208.99 & 216.57 & \textbf{169.41} & 417.02 & 266.27 & 221.42 & \textbf{135.84} & 81.05 & 82.43 & $\backslash$ & \textbf{78.25}\\
 9 & 176.91 & 245.56 & $\backslash$ & \textbf{155.62} & 840.57 & 293.85 & $\backslash$ & \textbf{102.03} & 139.97 & 148.99 & $\backslash$ & \textbf{103.92}\\
 10 & 201.62 & 397.99 & 386.27 & \textbf{155.1} & 537.53 & 414.56 & $\backslash$ & \textbf{110.73} & 195.97 & 251.42 & $\backslash$ & \textbf{118.95}\\
 11 & 391.47 & 567.7 & 477.99 & \textbf{239.68} & 135.89 & 98.28 & 88.28 & \textbf{82.77} & 244.68 & 217.28 & $\backslash$ & \textbf{143.21}\\
 12 & 405.33 & 766.69 & 642.84 & \textbf{270.09} & 339.57 & 363.85 & $\backslash$ & \textbf{146.51} & 99.05 & 104.14 & 100.00 & \textbf{59.28}\\
 13 & 98.14 & 92.43 & 108.28 & \textbf{85.25} & 225.54 & 122.45 & 140 & \textbf{96.68} & 122.02 & 102.43 & 148.28 & \textbf{79.42}\\
 14 & 110.63 & 114.14 & 148.28 & \textbf{101.53} & 524.56 & 378.7 & 321.42 & \textbf{185.13} & 126.37 & 174.85 & $\backslash$ & \textbf{100.65}\\
 15 & 166.14 & 180.71 & 228.28 & \textbf{153.99} & 128.85 & 156.57 & 140 & \textbf{102.27} & 218.59 & 134.85 & 184.85 & \textbf{106.48}\\
$A$ & 191.93 & 255.48 & 253.63 & \textbf{140.04} & 304.97 & 211.72 & 201.33 & \textbf{131.48} & 129.99 & 115.14 & 126.12 & \textbf{81.81}\\
  \bottomrule
 \end{tabular}
 \caption{\Revised{Costs of problems with obstacles in the AUV domain, the Taxi domain, and the Rover domain. ``$\mxplanner_\mathcal{R}$'' indicates {\ours} with no parameter bounds except for ``duration''. The columns of ``\metricff'' are the same as Table \ref{table:obstacles}.}} 
\label{table:obstacles_no_step_length}
 \end{table*}
\Revised{
Table \ref{table:obstacles_no_step_length} shows the results of {\ours} on the instances with obstacles without parameter bounds, i.e. the lower bounds and upper bounds of parameters are from $-\infty$ to $\infty$. 
As shown in the table, without parameter bounds, {\ours} still solves all instances and outperforms Metric-FF$^+$.
It also shows that {\ours} is able to solve mixed planning problems without prior knowledge on discretizing parameters.

\Revised{
To evaluate how the step length influences Metric-FF$^+$, we also show the results of Metric-FF$^+$ with different step lengths. As we mentioned before, a larger step length is more likely to result in crossing obstacles or more cost. Hence, we also apply Metric-FF$^+$ with step length being one and two. However, we do not show the results \emph{w.r.t.} one, because the discretized problems contain too many propositions and Metric-FF$^+$ fails to solve them. In the AUV domain, Metric-FF$^+$ with a smaller step length generally performs better than a larger step length.
Especially, for the 6-th instance, $\metricff_2$ even has the top performance, as the instance has a more direct solution plan that can be found more simply.
However, for other two domains, $\metricff_2$ performs worse, compared against the larger step lengths. It is because that Metric-FF$^+$ has to face larger and larger search space when step lengths get smaller increasingly, and it is getting more difficult for discretized planners to solve them.
Different from Metric-FF$^+$, {\ours} does not rely on discretization and it dynamically searches step lengths and step angles.} 
All the results demonstrate that {\ours} computes solution plans with lower costs making no use of prior knowledge on parameter discretization.
}

Compared to the discrete planner Metric-FF$^+$, which must be fed with fixed numeric parameters, {\ours} not only can plan without restrictions of step lengths, but also can compute plans with lower costs.

\subsubsection{Costs with the number of obstacles increasing}

\begin{figure}[!ht]
 \setlength{\abovedisplayskip}{1pt}
 \setlength{\belowdisplayskip}{1pt}
  \centering
  \includegraphics[width=0.8\textwidth]{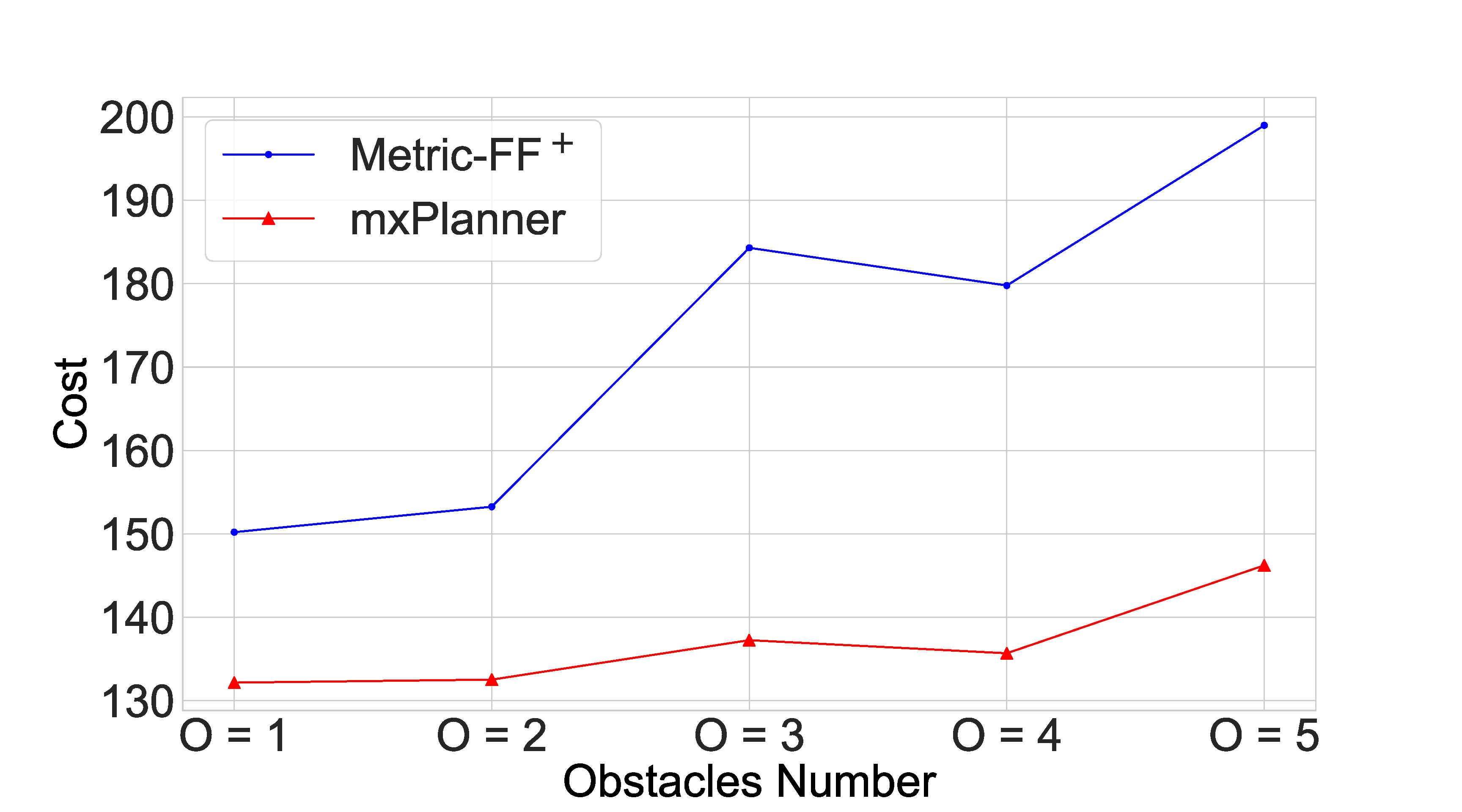}
  \caption{Average costs of ten instances with increasing numbers of obstacles in the AUV domain.}
  \label{figure:obstacle_increase}
\end{figure}

Figure \ref{figure:obstacle_increase} shows the average costs of 10 instances with $\bound = 10$ in the AUV domain. For different obstacle numbers, we generate afresh 10 instances with keeping the target regions. Surprisingly, the increasing number of obstacles does not lead \Revised{to} higher costs.
Counter-intuitively, for both planners, the average cost \emph{w.r.t.} three obstacles is a \Revised{little}\ignore{few} higher than that of the problems with four obstacles.
The reason is that in some instances with three obstacles, the places of obstacles make a \ignore{detour of path}\Revised{detour from the path}, which leads a higher cost.

Compared to Metric-FF, the trend of average costs computed by {\ours} grows more slowly. It shows that our approach is less influenced by the number of obstacles.

\subsubsection{Costs with different hyperparameters}\label{sec:costs_with_different_hyperparameters}
In this experiment, we would like to investigate the influence of different hyperparameters (i.e., $w_1$, $w_2$, $w_3$ in Equation (\ref{equation:loss_instantaneous})) on the performance of {\ours}.
We first randomly selected 10 instances in the AUV domain. We calculated the average cost over all the instances by {\ours} under different hyperparameters with $\bound = 10$.

\begin{figure}[!ht]
\setlength{\abovedisplayskip}{0pt}
\setlength{\belowdisplayskip}{0pt}
\centering
\subfigure[$w_1$ varied,  $w_2 = w_3 = 1$.]{
  \begin{minipage}[b]{0.31\textwidth}
    \includegraphics[width=\textwidth]{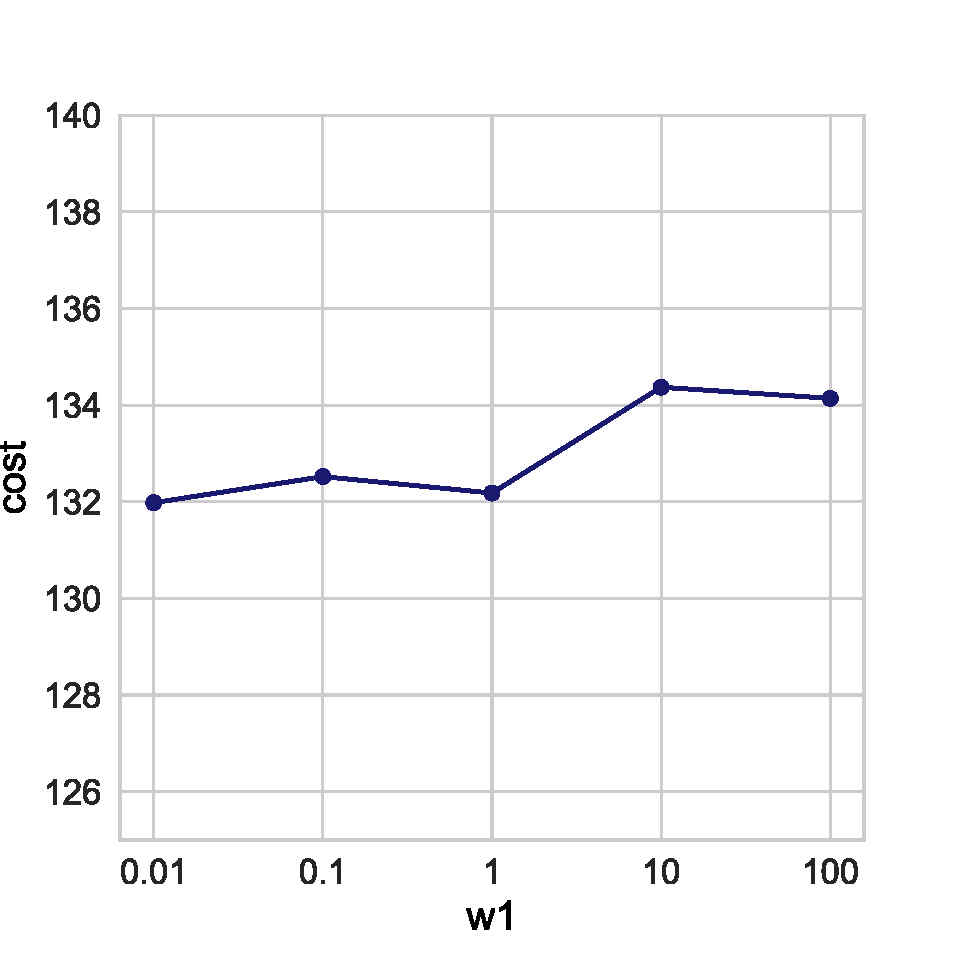}
  \end{minipage}
}
\subfigure[$w_2$ varied,  $w_1 = w_3 = 1$.]{
    \begin{minipage}[b]{0.31\textwidth}
    \includegraphics[width=\textwidth]{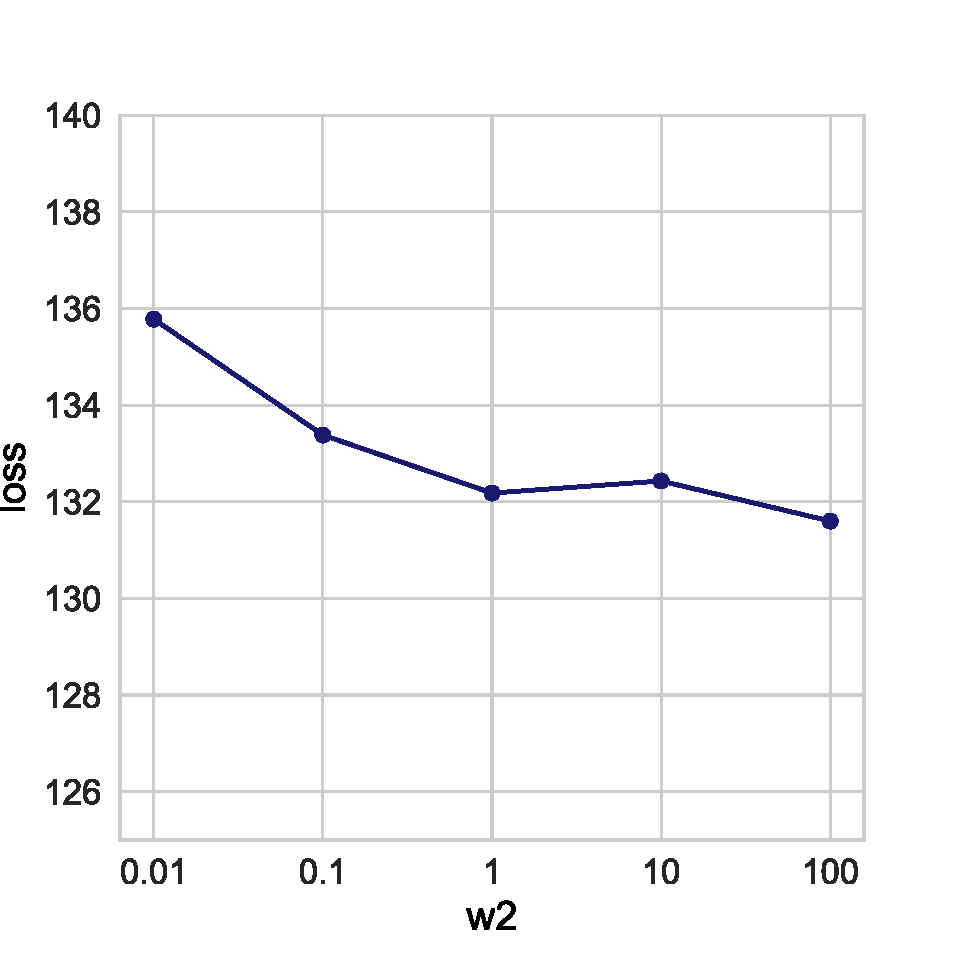}
    \end{minipage}
}
\subfigure[$w_3$ varied,  $w_1 = w_2 = 1$.]{
    \begin{minipage}[b]{0.31\textwidth}
    \includegraphics[width=\textwidth]{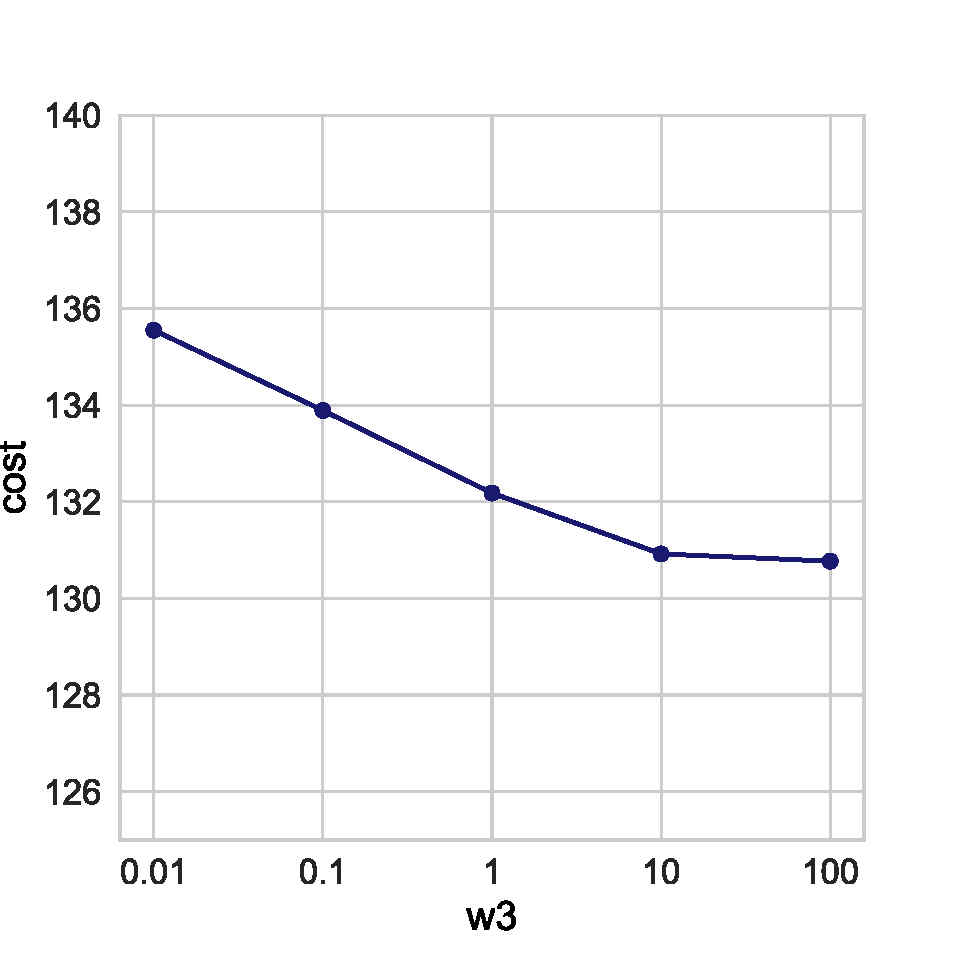}
    \end{minipage}
}
\caption{Average costs of 10 instances in the AUV domain with different hyperparameters.}
\label{figure:hyperparameters_cost}
\end{figure}

As illustrated in Figure \ref{figure:hyperparameters_cost}(a), the weight $w_1$ has a negligible effect on the planning performance. Because $w_1$ is the weight related to $\mathcal{L}_b$ which guides variables to satisfy numeric preconditions and to achieve goals, it does not affect the cost directly. Figure \ref{figure:hyperparameters_cost}(b)
demonstrates that the cost tends to decrease when $w_2$ gets larger. It is because that 
when $w_2$ increases, the propagated gradient about crashing an obstacle becomes larger.
It further leads {\ours} to prioritize avoiding obstacles above meeting numeric preconditions.
In consequence, it is more possible to find a detour path.
\Revised{Figure \ref{figure:hyperparameters_cost}(c) shows a negative correlation between costs and the weight $w_3$, the weight of the loss about plan costs. Intuitively, the bigger weight of cost is, the bigger the propagated gradient about cost is.}

\begin{figure}[t]
\setlength{\abovedisplayskip}{0pt}
\setlength{\belowdisplayskip}{0pt}
  \centering
  \includegraphics[width=0.8\textwidth]{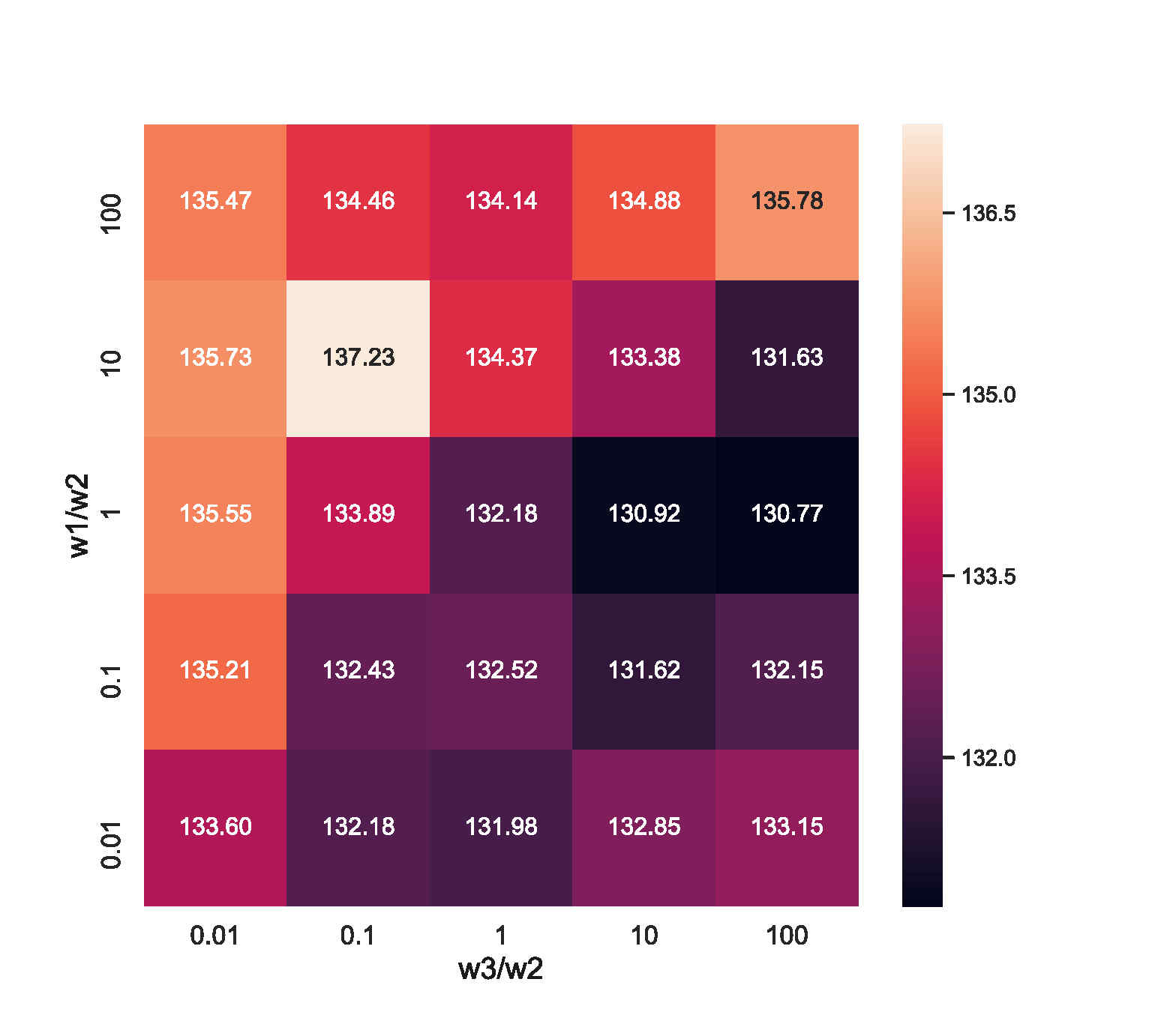}
  \caption{The heatmap of the cost with different hyperparameters.}
  \label{figure:heatmap_loss}
\end{figure}

We also give a more intuitive exhibition about how hyperparameters influence plan costs in a heatmap, as shown in Figure \ref{figure:heatmap_loss}. 
The weight $w_2$ is set to 1.
The x-axis is the ratio of $w_3$ to $w_2$ and the y-axis is the ratio of $w_1$ to $w_2$, which both range from 0.01 to 100. The darker the area is, the smaller the average cost is.
Obviously, when $w_1:w_2:w_3 = 1:1:100$, {\ours} has the best performance and has an average cost of 130.77.

From the heatmap, we can observe that from right to left, the cost tends to decrease. That is, the weight $w_3$ is getting larger.

\subsubsection{Iterations with respect to various hyperparameters}
\begin{figure}[!t]
\setlength{\abovedisplayskip}{0pt}
\setlength{\belowdisplayskip}{0pt}
\centering
\subfigcapskip=-5pt
\subfigcapmargin = .05cm
\subfigure[$w_1$ varied,  $w_2 = w_3 = 1$. ]{
  \begin{minipage}[b]{0.31\textwidth}
    \includegraphics[width=\textwidth]{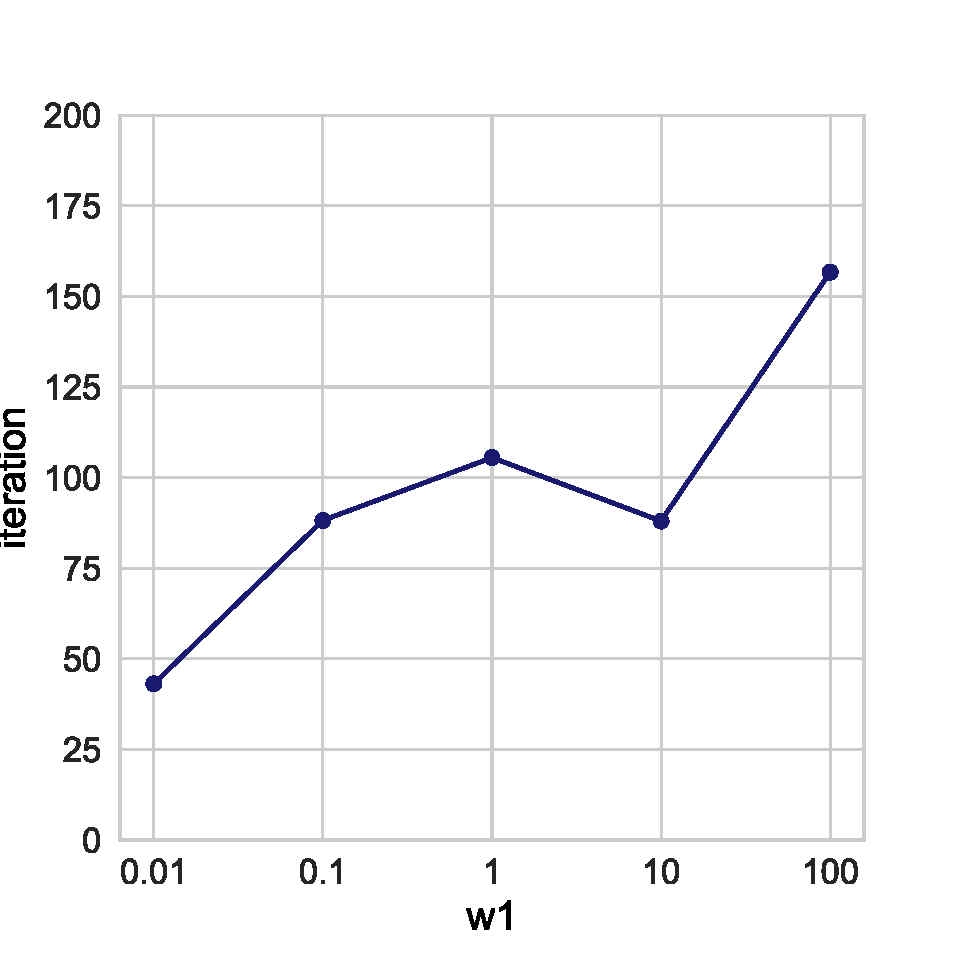}
  \end{minipage}}
\subfigure[$w_2$ varied, $w_1 = w_3 = 1$.]{
    \begin{minipage}[b]{0.31\textwidth}
    \includegraphics[width=\textwidth]{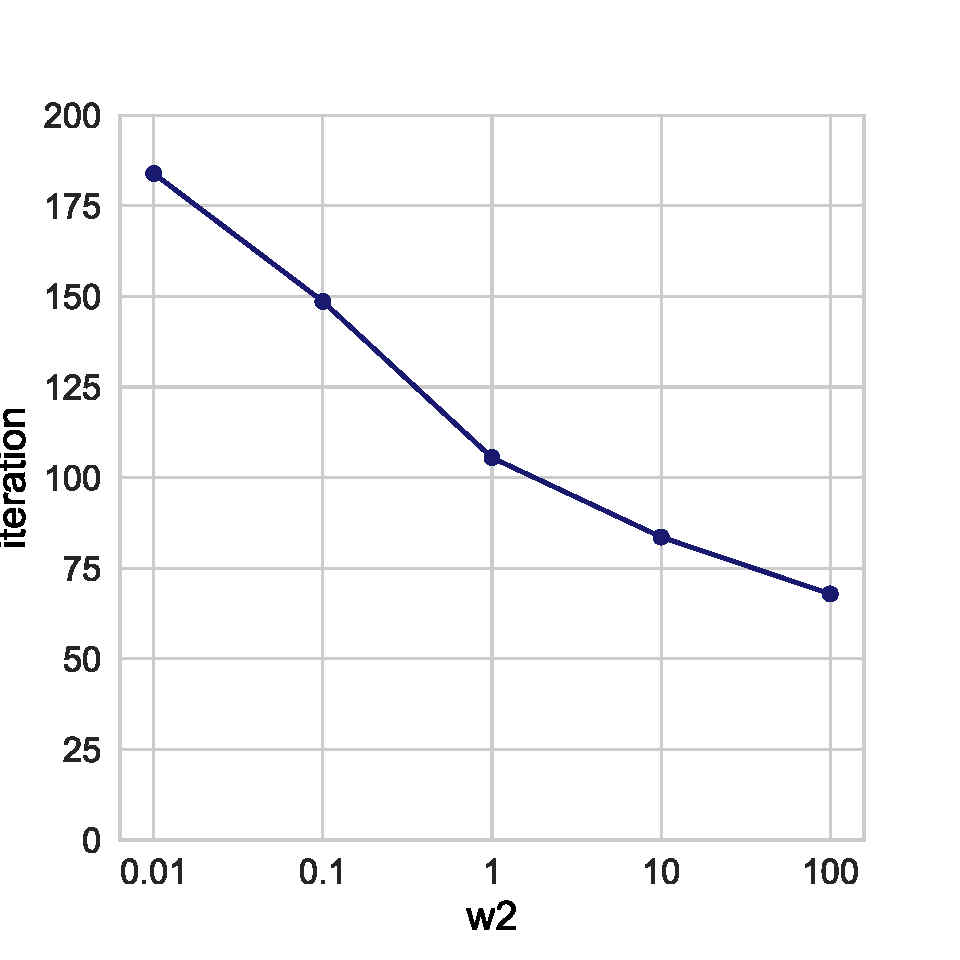}
    \end{minipage}}
\subfigure[$w_3$ varied, $w_1 = w_2 = 1$.]{
    \begin{minipage}[b]{0.31\textwidth}
    \includegraphics[width=\textwidth]{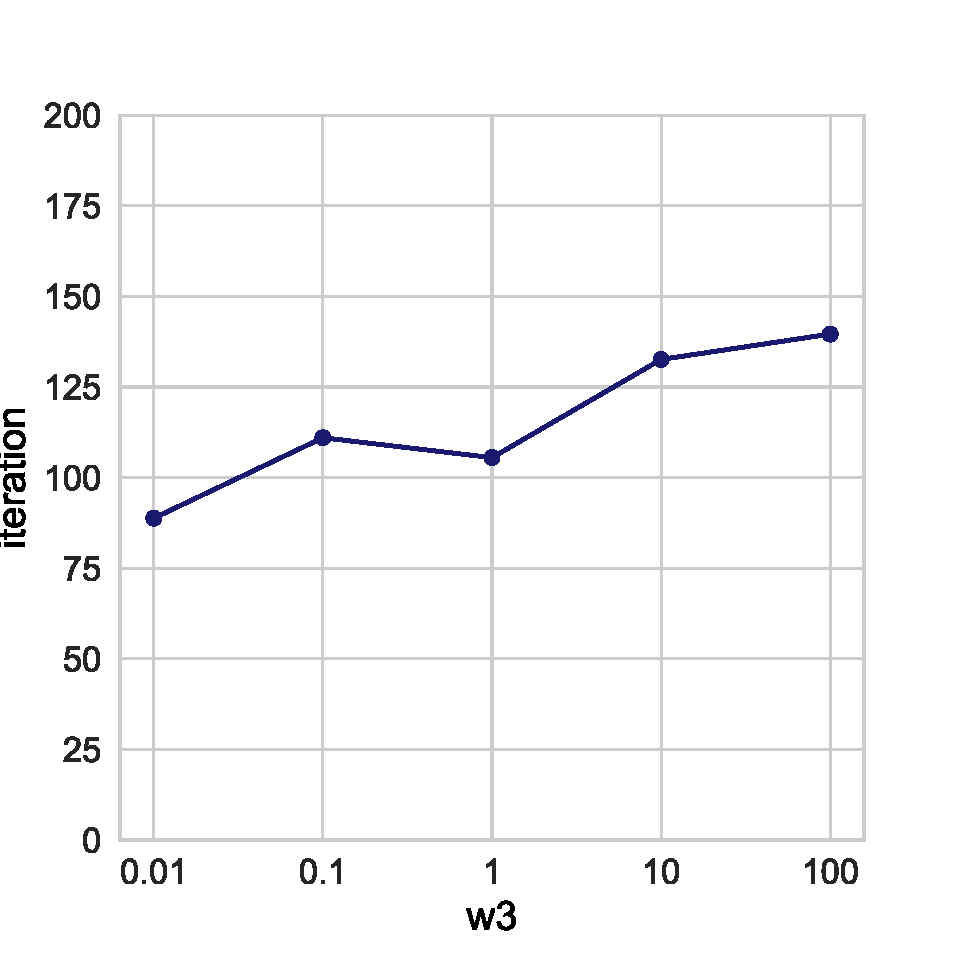}
    \end{minipage}}
\caption{Average computing iterations of 10 instances in the AUV domain with different hyperparameters.}
\label{figure:hyperparameters_iteration}
\end{figure}

To see the impact of hyperparameters, we recorded the numbers of iterations of {\ours} with respect to various values of hyperparameters.
We used the same benchmarks \MajorRevision{as} in Section \ref{sec:costs_with_different_hyperparameters}. The results are shown in Figure \ref{figure:hyperparameters_iteration}.
From Figure \ref{figure:hyperparameters_iteration},
we can see that {\ours} generally takes fewer iterations to generate a plan with $w_2$ increasing. \MajorRevision{In contrast,} with $w_1$ and $w_3$ increasing, the iteration tends to increase as well.

\begin{figure}[!h]
 \setlength{\abovedisplayskip}{1pt}
 \setlength{\belowdisplayskip}{1pt}
  \centering
  \includegraphics[width=0.8\textwidth]{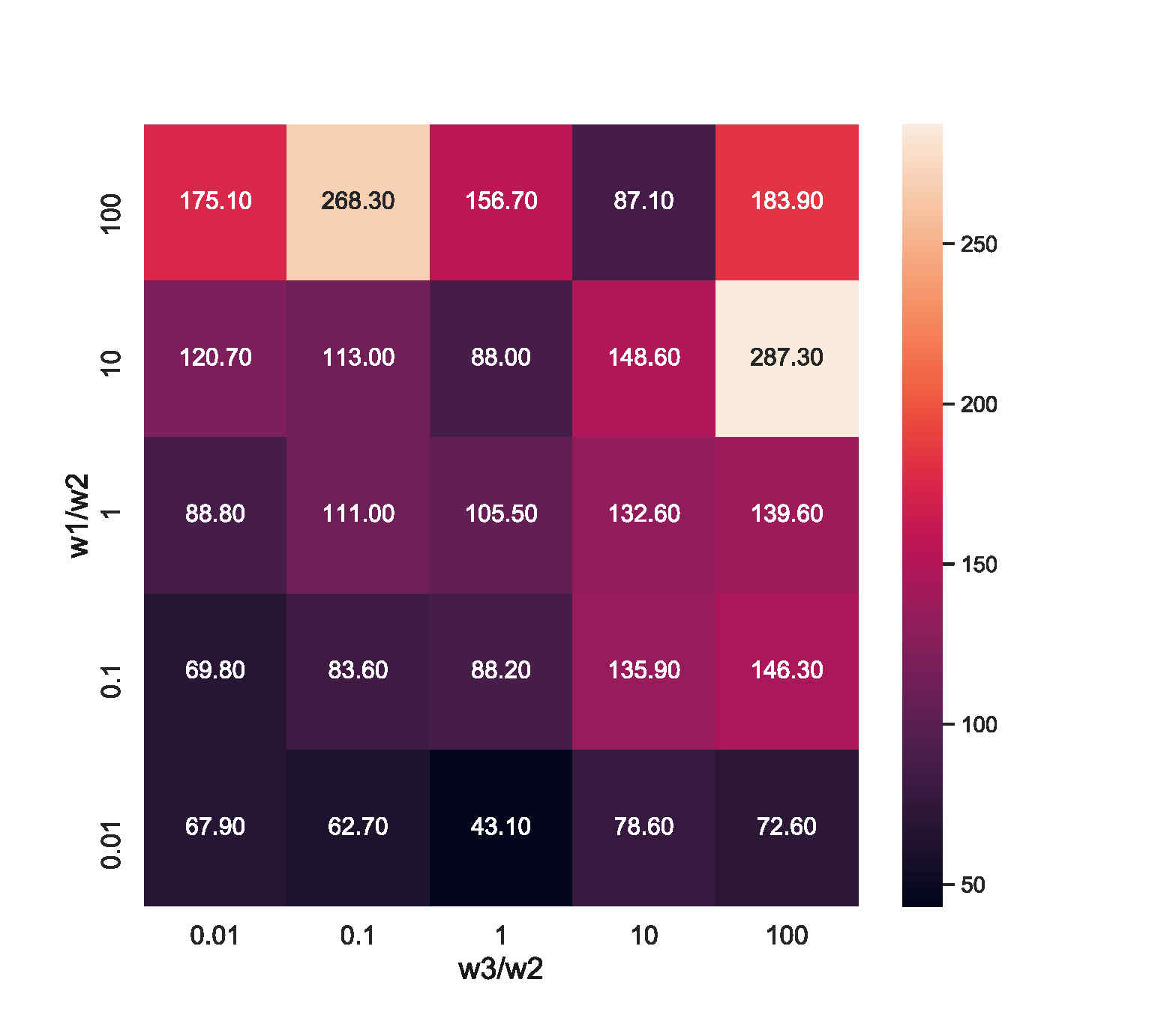}
  \caption{The heatmap of average computing iterations  of 10 instances with different hyperparameters.}
  \label{figure:heatmap_iteration}
\end{figure}

Similarly, we also used a heatmap to exhibit the influences of the hyperparaemters on the running iterations of {\ours}, as shown in
Figure \ref{figure:heatmap_iteration}. Different from the average costs, when $w_1$ is small, the number of iterations also becomes small. When $w_1:w_2:w_3 = 0.01 :1:1$, the number of iterations of {\ours} is the smallest oval all approaches.
This is because $w_1$ relates to the gradient of $\mathcal{L}_b$ which guides to find a relaxed plan. A small value of $w_1$ means to be a modest modification of the relaxed plan. In other words, when the relaxed plan is similar with a valid plan, {\ours} is able to find a valid plan quickly.

\subsubsection{Running time:}
\MajorRevision{
\begin{table}[!ht]
\centering
\scriptsize
\begin{tabular}{p{1mm}|cccc|cccc}
\toprule
   & $\mxplanner_{10}$ & $\metricff_{10}$  & $\popcorn_{10}$      & $\ScottyActivity_{10}$  & $\mxplanner_{20}$ & $\metricff_{20}$  & $\popcorn_{20}$      & $\ScottyActivity_{20}$   \\
 \hline
1  & 11.63   & 0.07 & 0.137            & 0.11 & 6.89   & 0.06             & 0.148 & 0.19 \\
2  & 28.38   & 0.09 & 0.19             & 0.48 & 16.81  & 0.06             & 0.263 & 0.37 \\
3  & 78.23   & 0.07 & 0.353            & 0.76 & 46.45  & 0.06             & 0.192 & 0.34 \\
4  & 104.88  & 0.11 & 2.441            & 0.19 & 102.47 & 0.06             & 0.163 & 0.76 \\
5  & 11.05   & 0.08 & 0.172            & 0.12 & 12.06  & 0.05             & 0.126 & 0.22 \\
6  & 16.79   & 0.07 & 0.15             & 0.37 & 7.39   & 0.05             & 0.114 & 0.20 \\
7  & 75.16   & 0.1  & 1.424            & 1.96 & 73.86  & 0.06             & 0.343 & 0.62 \\
8  & 134.70  & 0.1  & 0.659            & 4.88 & 129.10 & 0.06             & 0.365 & 1.17 \\
9  & 49.30   & 0.1  & 0.235            & 0.54 & 41.85  & \textbackslash{} & 0.185 & 0.28 \\
10 & 38.59   & 0.08 & 0.174            & 0.21 & 27.49  & 0.05             & 0.116 & 0.67 \\
11 & 1198.58 & 0.09 & 8.069            & 0.18 & 896.77 & 0.06             & 0.531 & 0.94 \\
12 & 649.38  & 0.13 & \textbackslash{} & 0.30 & 271.03 & 0.05             & 1.649 & 1.28 \\
13 & 15.96   & 0.06 & 0.139            & 0.13 & 13.37  & 0.04             & 0.123 & 0.12 \\
14 & 47.81   & 0.06 & 0.259            & 0.12 & 39.94  & 0.05             & 0.128 & 0.37 \\
15 & 239.73  & 0.07 & 0.66             & 0.17 & 131.12 & 0.05             & 0.165 & 2.47 \\
\emph{A}  & 180.01  & 0.09 & 1.08             & 0.70 & 121.11 & 0.05             & 0.31  & 0.67     \\
\bottomrule
\end{tabular}
\caption{\MajorRevision{Running time of instances without obstacles in the AUV domain with different bounds. $\emph{A}$ in the table indicates the average time of the solved problem.}}
\label{table:running_time_no_obstacle}
\end{table}

Although {\ours} is superior to the other approaches on navigating distance, its running time generally exceeds theirs. The running time of {\ours} varies significantly depending on the complexity of planning problems. Table \ref{table:running_time_no_obstacle} shows results of running time in the AUV domain with different bounds. \ignore{Notably, the experiments of running time of four methods are implemented on the same machine, i.e. Ubuntu 16.04 on a 50GB of memory, using different machines does not affect the qualities of plans about navigating distance. }Specifically, Metric-FF took the least time to generate plans. The reason is due to the discretization of Metric-FF, it ignored durations and control parameters, decreasing difficulties of planning problems. Compared to the discretized planning problems, POPCORN$^+$ and ScottyActivity handle numeric planning problems with control parameters and durations, which are more complex. Therefore, they took more time to generate plans compared to Metric-FF. As shown in the Table \ref{table:running_time_no_obstacle}, $\ScottyActivity_{10}$ took less time than $\popcorn_{10}$. However, $\popcorn_{20}$ generated plans faster when bound was set to be 20. Thus, performance of POPCORN$^+$ is more affected by different bounds. Different from Metric-FF, POPCORN$^+$, and ScottyActivity, {\ours} is an iteration-based planning approach which aims at computing plans and minimizing loss iteratively. In each iteration, {\ours} calculates heuristic values that determine whether agents collide with obstacles or not. Therefore, the running time of {\ours} is generally higher than other planners that divide the search space into subspaces before planning, which helps reduce the running time. It would be interesting to investigate the possibility of improving the efficiency of {\ours} in the future.
}
\ignore{
\Revised{
Unfortunately, {\ours} is superior to the other approaches but its running time generally exceeds theirs. The running time of {\ours} varies significantly, depending on the complexity of planning problems. More specifically, its running time ranges from 43 seconds to 7892 seconds for AUV, from 36 seconds to 15269 seconds for Taxi, and from 25 seconds to 23647 seconds for Rover, respectively. Compared with ScottyActivity, {\ours} is an iteration-based planning approach which needs to find a relaxed plan and to minimize the loss in each iteration. ScottyActivity invokes an off-the-shelf solver of convex optimization which is efficient in problem solving.
Its efficiency helps ScottyActivity to find a valid plan quickly.
Also, in each iteration, {\ours} calculates heuristic value and it determines whether the agent collides with each obstacle or not. 
Different from Metric-FF, POPCORN, and ScottyActivity, {\ours} spends more time in precisely avoiding obstacles by using gradient descent to iteratively minimize the loss. Its running time is generally higher than other planners that divide the search space into subspaces before planning, which helps reduce the planning time. We will seek more efficient algorithms or revise codes for greater efficiency in the future.
}}

\subsubsection{Analysis}

\ignore{As shown above, {\ours} is able to solve planning problems
\Revised{mixed with discrete logical relations and continuous numeric change which can be linear or non-linear. Also, the combination of heuristic searching and gradient-based framework enables {\ours} to handle collision-avoidance.} The above experimental results also show its performance in convex or non-convex planning problems.}

\Revised{We analyze the merits and demerits of {\ours} with comparison to other approaches.}

\begin{itemize}
    \item \emph{Comparison with Metric-FF:}
    {\ours} dynamically adjusts the step lengths and searching angles in each step to generate flexible plans, instead of fixing step length determined by prior knowledge, as done by Metric-FF. Figure \ref{figure:experiments_obstacles} shows two valid plans computed by $\metricff_{10}$ (i.e., Figure \ref{figure:experiments_obstacles}(a)) and $\mxplanner_{10}$ (i.e., Figure \ref{figure:experiments_obstacles}(b)) for the same instance with obstacles in the AUV domain, respectively. 
As the parameters are fixed to be 10,
each movement of Metric-FF has an identical step length, which leads a higher plan cost than {\ours}, while from Figure \ref{figure:experiments_obstacles}(b), it is not difficult to observe that in the plan computed by {\ours} the length of each step is different.

\begin{figure}[!ht]
\setlength{\abovedisplayskip}{0pt}
\setlength{\belowdisplayskip}{0pt}
\centering
\subfigcapskip=-5pt
\subfigcapmargin = .05cm
\subfigure[~~~$\metricff_{10}$]{
  \begin{minipage}[b]{0.35\textwidth}
    \includegraphics[width=\textwidth]{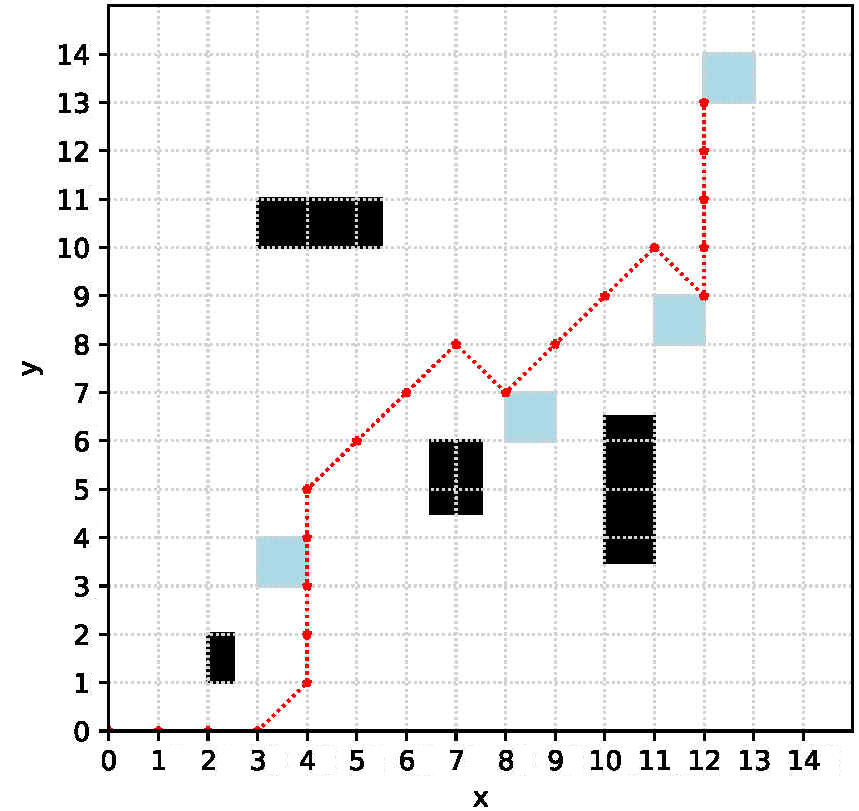}
  \end{minipage}
}
\subfigure[~~~$\mxplanner_{10}$]{
    \begin{minipage}[b]{0.35\textwidth}
    \includegraphics[width=\textwidth]{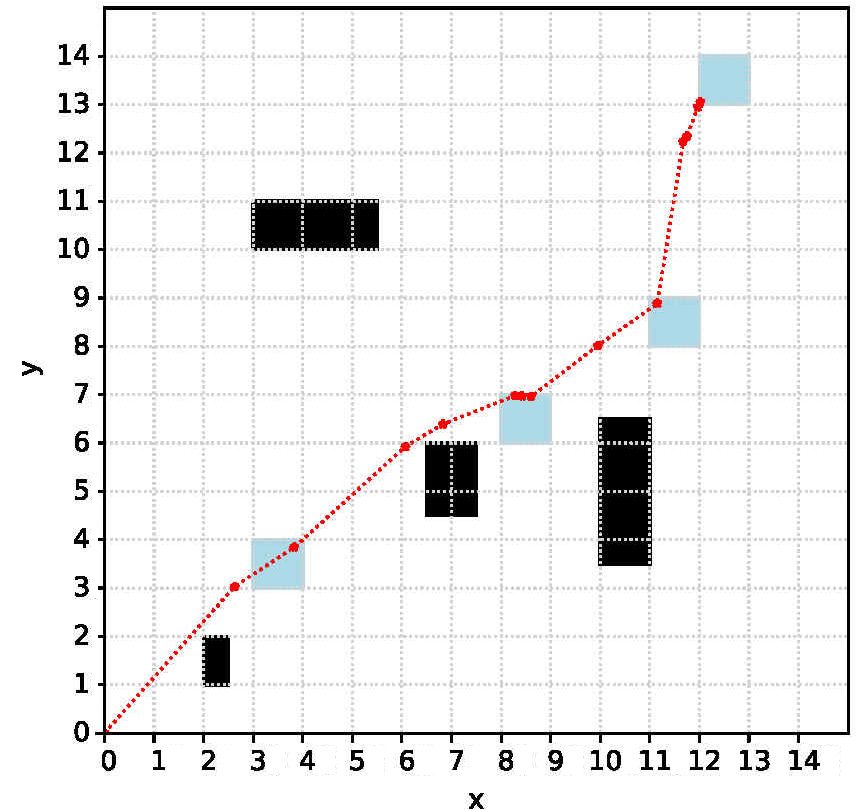}
    \end{minipage}
}
\caption{\MajorRevision{Two example plans in the AUV domain computed by Metric-FF and {\ours} with parameter bounds fixed to be 10, respectively. The bold red dot is the stop position of each movement. Blue areas are target regions and black areas are obstacles. Each value of axes labels indicates the value is multiplied by $10$.}}
\label{figure:experiments_obstacles}
\end{figure}

\item \emph{Comparison with ScottyActivity:}
ScottyActivity optimizes a computed plan, which may lead it to fall into a local optimum. Differently, {\ours} searches candidate plans and optimizes parameters simultaneously, which, to a great extent, reduces the probability of local optimum. Figure \ref{figure:experiments_no_obstacles} shows two valid plans computed by ScottyActivity (a) and {\ours} (b) respectively for the same obstacle-free instance in the AUV domain. 
We can see that the trajectory of ScottyActivity makes a detour and is significantly longer than that of {\ours}.
Notably, in Figure \ref{figure:experiments_no_obstacles}(a), movements each are too close, making the trajectory look like a bold line.

\begin{figure}[!ht]
\setlength{\abovedisplayskip}{0pt}
\setlength{\belowdisplayskip}{0pt}
\centering
\subfigcapskip=-5pt
\subfigcapmargin = .05cm
\subfigure[~~~SA$_{10}$]{
  \begin{minipage}[b]{0.35\textwidth}
    \includegraphics[width=\textwidth]{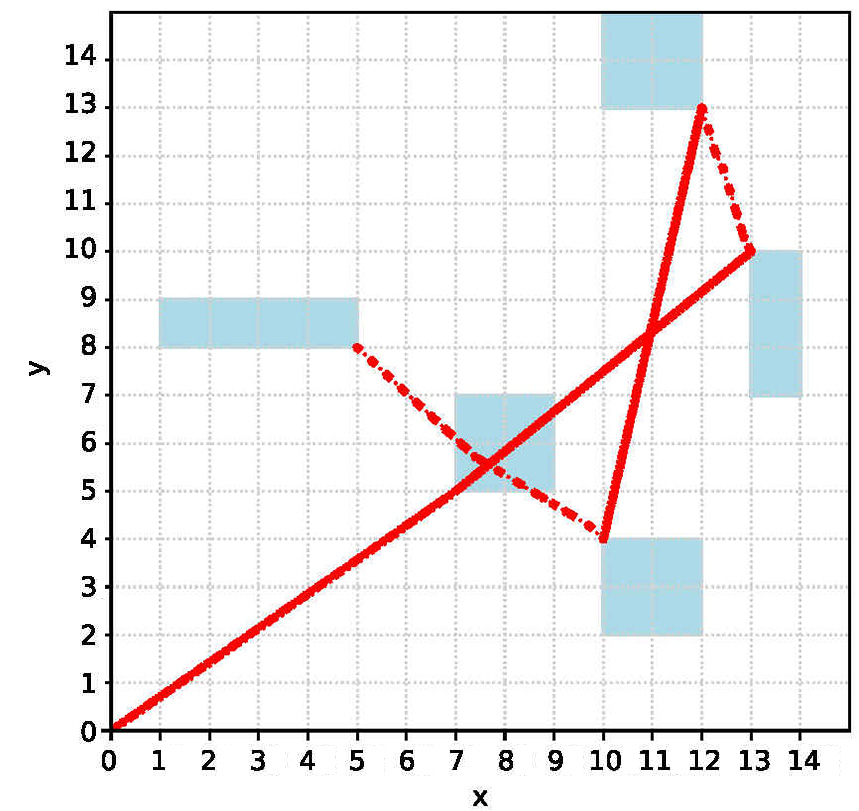}
  \end{minipage}
}
\subfigure[~~~$\mxplanner_{10}$]{
    \begin{minipage}[b]{0.35\textwidth}
    \includegraphics[width=\textwidth]{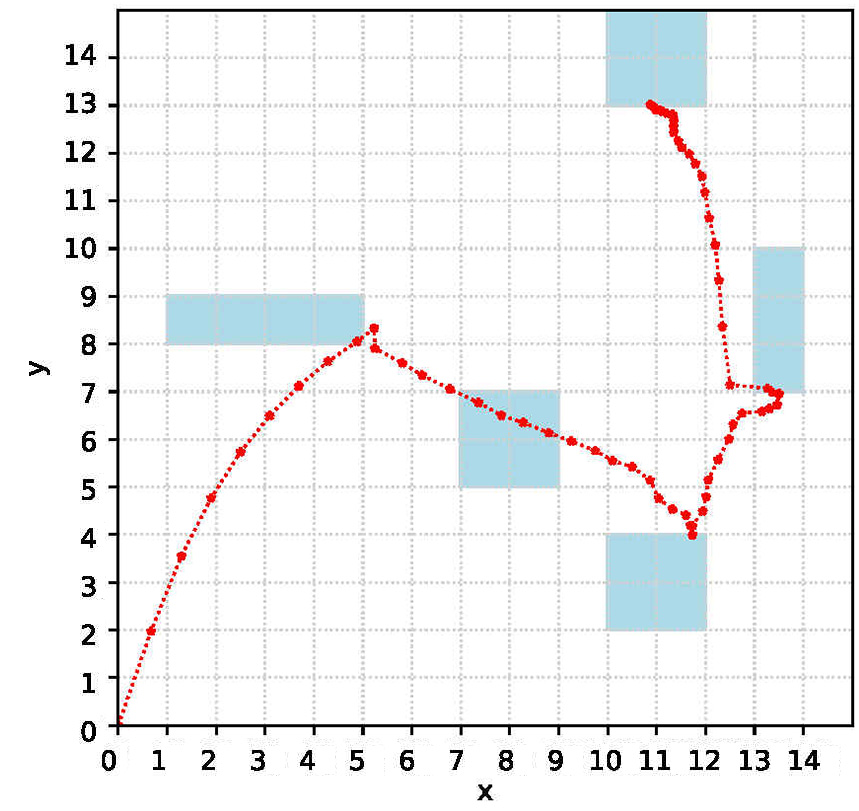}
    \end{minipage}
}
\caption{\MajorRevision{Two example plans of an obstacle-free instance in the AUV domain computed by ScottyActivity and {\ours} with parameter bounds fixed to be 10, respectively. The bold red dot is the stop position of each movement. Blue areas are target regions and black areas are obstacles. Each value of axes labels indicates the value is multiplied by $10$. }
}
\label{figure:experiments_no_obstacles}
\end{figure}

\item
\Revised{\emph{Challenging instances:}The Taxi domain contains a more complicated problem setting which may include round paths, compared with the other two domains. As shown in Figure \ref{fig:mx_back}, the plan trajectory goes back and forth on the map.\begin{figure}[!ht]
    \centering
    \includegraphics[width=0.45\textwidth]{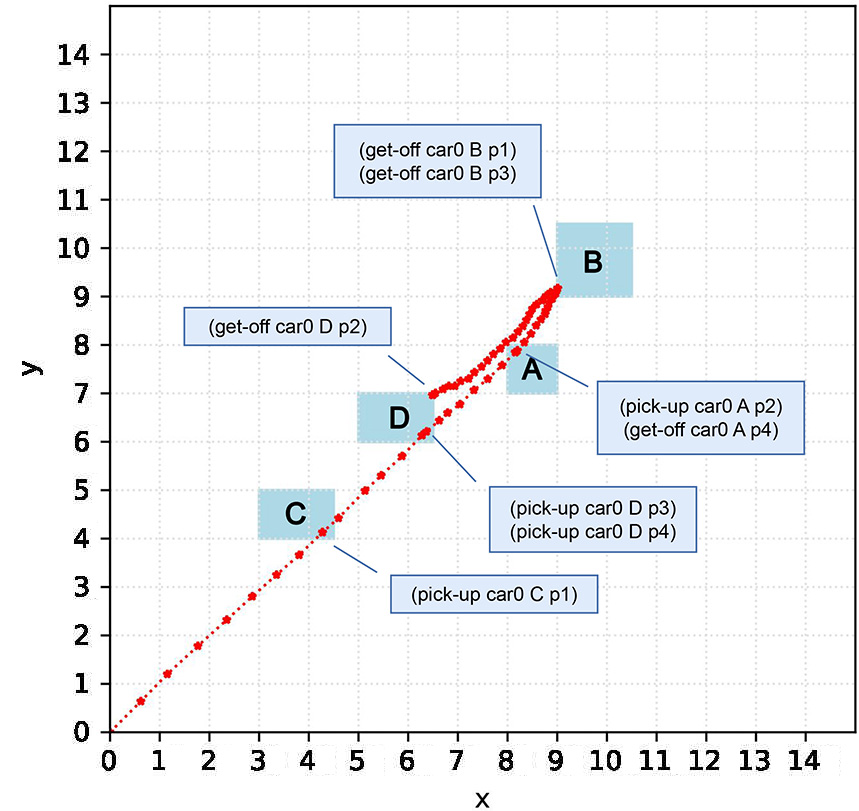}
    \caption{\MajorRevision{An example plan in the Taxi domain calculated by {\ours} with parameter bounds fixed to be 10. The bold red dot is the stop position of each movement. Blue areas labeled with letters are target regions. Logical actions are shown in text-boxes. Each value of axes labels indicates the value is multiplied by $10$.}}
    \label{fig:mx_back}
\end{figure} In particular, the step lengths in the beginning of the plan are longer than latter steps. The reason is that, according to the gradient in Equation (\ref{partial_derivative}), parameters in each step are affected by the instantaneous loss about the action sequence from the current step to the last step. In other words, parameters in latter steps are less influenced by the instantaneous loss.
}

\item \emph{Failures to the best:}
    In some cases prior knowledge on parameter discretization indeed helps computing better solution plans than {\ours}, since the heuristic module in {\ours} may fall into a local optimum. Meanwhile, since {\ours} tries to satisfy the nearest numeric preconditions of candidate actions during the heuristic searching procedure, {\ours} does not guarantee plan optimality.

\end{itemize}


\section{Conclusion}
In this paper, \Revised{we introduce \emph{mixed} planning problems, which are extended from numeric planning problems with control parameters. In order to handle \emph{mixed} planning problems, }we present {\ours}, \MajorRevision{a gradient-based approach that borrows the framework of RNNs by replacing the neural cells with heuristic search and transition modules.}
We evaluate {\ours} in three domains and the experimental results show the superiority of  {\ours} on plan quality, especially in obstacles avoidance problems compared against state-of-the-art approaches. Also, we evaluate the influence of the hyperparameters on {\ours}.


Compared with the previous works, the advances introduced by our planner are shown as follows.
First of all, the combination of heuristic searching and gradient-based framework gives {\ours} the ability to handle \emph{mixed} planning problems without discretization. Especially, {\ours} performs well when handling \emph{mixed} planning problems with non-convex continuous numeric space, i.e., containing obstacles. Second, {\ours} is also competitive in planning problems without obstacles compared with ScottyActivity, a state-of-the-art planner on planning missions with convex continuous numeric space.

Finally, we present several directions to improve our approach in the future.
First, {\ours} spends a lot of time on gradient descent, iteratively update numeric parameters for minimizing loss.
It is promising to create a more efficient approach of parameter updating.
Second, plans computed by {\ours} may not be optimal, the reason is that {\ours} always attempts to satisfy the nearest numeric preconditions of candidate actions in heuristic module. Last but not the least, it is also interesting to investigate more efficient approaches to improve heuristic searching. 

\MajorRevision{
In this paper, {\ours} focuses on three \emph{mixed} navigation domains with obstacles. In the future, we will consider improving {\ours} from three challenging aspects. First of all, in this paper, for each variable, its numeric sub-goal was computed according to the upper and lower bounds of numeric preconditions. It could be more efficient if we improve our heuristic module by considering long-term numeric preconditions, especially for variables of numeric preconditions of actions that are executed many times. It is, however, challenging since the heuristic module computes a relaxed planning graph without a detailed and accurate order. The choice of outputting actions is based on the current state and numeric parameters, which could be different from the computation of relaxed planning graph. Next, {\ours} stops when it finds a valid plan or its running time exceeds the cut-off time. If {\ours} fails to output the most appropriate action, or falls into a local minimum during updating, {\ours} may be disable to compute a valid plan. It could be helpful to add a backtracking algorithm to help {\ours} compute plans efficiently. Specifically, a computed plan may contain a series of numeric actions with the same action name and objects but different numeric parameters. It is difficult to distinguish which action of the plan is incorrect. On the other hand, due to actions computed dynamically during gradient descent, it is also challenging to integrate a backtracking algorithm with a gradient-based framework and determine when to use the backtracking algorithm. Finally, it would be interesting to consider seeking a novel approach to initializing parameters instead of initializing parameters randomly, since the initialization directly influences the quality of plans and efficiency of searching.
}

\section*{Acknowledgement}
This research was funded by the National Natural Science Foundation of China (Grant No. 62076263, 61906216), Guangdong Natural Science Funds for Distinguished Young Scholar (Grant No. 2017A030306028), Guangdong Special Branch Plans Young Talent with Scientific and Technological Innovation (Grant No. 2017TQ04X866), Guangdong Basic and Applied Basic Research Foundation (2020A1515010642). Kambhampati’s research is supported in part by the ONR grants N00014-16-1-2892, N00014-18-1-2442, N00014-18-1-2840, the AFOSR grant FA9550-18-1-0067, and the NASA grant NNX17AD06G.

\bibliographystyle{elsarticle-harv}
\bibliography{aij}

\end{document}